\documentclass[a4paper,12pt, PhD]{stb-thesis}
\usepackage[left=3cm,right=2cm,top=2.5cm,bottom=2.5cm]{geometry}
\usepackage{afterpage}

\usepackage[margin=\the\parindent,small,bf,rm]{caption}
\usepackage{graphicx}
\usepackage{pdfpages}
\usepackage{subfig}

\usepackage[english,afrikaans]{babel}
\usepackage{microtype}
\usepackage{setspace}

\usepackage{enumitem}
\usepackage{listings,xfp}
\usepackage{anyfontsize}
\usepackage{bold-extra}
\usepackage{quotchap}

\makeatletter
{\large %
	\xdef\f@size@large{\f@size}
	\xdef\f@baselineskip@large{\f@baselineskip}
	\Large %
	\xdef\f@size@Large{\f@size}
	\xdef\f@baselineskip@Large{\f@baselineskip}
}
\newcommand{\largetoLarge}{%
	\fontsize
	{\fpeval{(\f@size@large+\f@size@Large)/2}}
	{\fpeval{(\f@baselineskip@large+\f@baselineskip@Large)/2}}%
	\selectfont
}
\makeatother
\makeatletter
{\tiny %
	\xdef\f@size@tiny{\f@size}
	\xdef\f@baselineskip@tiny{\f@baselineskip}
	\normalsize %
	\xdef\f@size@normalsize{\f@size}
	\xdef\f@baselineskip@normalsize{\f@baselineskip}
}
\newcommand{\tinytonormal}{%
	\fontsize
	{\fpeval{(\f@size@tiny+\f@size@normalsize)/2}}
	{\fpeval{(\f@baselineskip@tiny+\f@baselineskip@normalsize)/2}}%
	\selectfont
}
\makeatother
\makeatletter
{\tiny %
	\xdef\f@size@tiny{\f@size}
	\xdef\f@baselineskip@tiny{\f@baselineskip}
	\normalsize %
	\xdef\f@size@normalsize{\f@size}
	\xdef\f@baselineskip@normalsize{\f@baselineskip}
}
\newcommand{\othertinytonormal}{%
	\fontsize
	{\fpeval{(\f@size@tiny+\f@size@normalsize)*0.4}}
	{\fpeval{(\f@baselineskip@tiny+\f@baselineskip@normalsize)*0.4}}%
	\selectfont
}
\makeatother

\makeatletter
{\normalsize %
	\xdef\f@size@normalsize {\f@size}
	\xdef\f@baselineskip@normalsize {\f@baselineskip}
	\large %
	\xdef\f@size@large{\f@size}
	\xdef\f@baselineskip@large{\f@baselineskip}
}
\newcommand{\normaloarge}{%
	\fontsize
	{\fpeval{(\f@size@normalsize +\f@size@large)/2}}
	{\fpeval{(\f@baselineskip@normalsize +\f@baselineskip@large)/2}}%
	\selectfont
}
\makeatother

\usepackage{textcase}
\usepackage{mfirstuc}
\usepackage{titlecaps}
\usepackage{stringstrings}
\usepackage[raggedright,rm,bf]{titlesec}
\usepackage{titlecaps}
\titlelabel{\thetitle.\quad}
\titleformat{\chapter}[display]{\LARGE\rmfamily\raggedleft}{\scshape \chaptertitlename\ \thechapter}{3pt}{\titlerule\vspace{1.5ex} \scshape \Huge \centering}[\vspace{1.5ex}\titlerule]
\titlespacing*{\chapter}{0pt}{30pt}{40pt}  %

\titleformat{\section}{\rmfamily\scshape\Large}{\thesection}{1em}{}
\titleformat{\subsection}{\rmfamily\scshape\large}{\thesubsection}{1em}{}
\titleformat{\subsubsection}{\rmfamily\scshape\normalsize}{\thesubsubsection}{1em}{}

\newcommand{\ShortInTextTitle}[1]{\textbf{\scshape #1}}

\usepackage{xcolor,titletoc}
\makeatletter
\let\originall@chapter\l@chapter
\def\l@chapter#1#2{\originall@chapter{{\rmfamily #1}}{#2}}
\makeatother
\let \savenumberline \numberline
\def \numberline#1{\savenumberline{#1.}}

\usepackage{acro}
\usepackage{multicol}
\usepackage{ragged2e}
\newlength\myitemwidth
\newlist{myacronymlist}{description}{1}
\setlist[myacronymlist]{
	labelindent = 0pt ,
	labelsep    = 0pt ,
	leftmargin  = \myitemwidth ,
	labelwidth  = \myitemwidth ,
	itemindent  = 0pt ,
	format      = \RaggedRight 
}

\RenewAcroTemplate{long-short}{%
	\acroiffirstTF{%
		\acrowrite{long}%
		\acspace(%
		\acroifT{foreign}{\acrowrite{foreign}, }%
		\acrowrite{short}%
		\acroifT{alt}{}%
		\acrogroupcite
		)%
	}%
	{\acrowrite{short}}%
}
 \NewAcroTemplate[list]{styleabbrev}{%
 	\normalsize
 	\AcroNeedPackage{array}%
 	\acronymsmapF{
 		\AcroAddRow{
 			\vspace{1.0ex}\par
 			\acrowrite{short} & \acrowrite{alt}
 			\acropagefill
 			\acropages
 			{\acrotranslate{page}\nobreakspace}
 			{\acrotranslate{pages}\nobreakspace}%
 			\tabularnewline
 		  }%
 	  }
    {\AcroRerun}%
    \acropreamble
     \par\noindent
    {\onehalfspacing\begin{tabular}{>{\bfseries}lp{.7\linewidth}}
     	 \AcronymTable
      \end{tabular}}
  }

\acsetup{
	list/template  = styleabbrev,
}

\usepackage[toc,section=section,nonumberlist]{glossaries-extra}
\usepackage{glossary-superragged}
\usepackage[intoc, english]{nomencl}
\usepackage{tabularx, booktabs}
\usepackage{calc}

\newglossarystyle{mystyle}{
	\setglossarystyle{treegroup}%
	
}
\setglossarystyle{mystyle}

\renewcommand{\nomgroup}[1]{%
	\ifthenelse{\equal{#1}{V}}{\item[{\rmfamily\scshape\Large Variables}]}{%
		\ifthenelse{\equal{#1}{C}}{\item[{\rmfamily\scshape\Large Constants}]}{
			\ifthenelse{\equal{#1}{F}}{\item[{\rmfamily\scshape\Large Functions}]}{}}}%
}

\makenomenclature

\makeatletter
\renewcommand\tableofcontents{%
	\chapter*{\contentsname}%
	\@mkboth{\scshape \contentsname}%
	{\scshape \contentsname}%
	\@starttoc{toc}%
}
\renewcommand\listoftables{%
	\chapter*{\listtablename}%
	\@mkboth{\scshape\listtablename}%
	{\scshape\listtablename}%
	\@starttoc{lot}%
}
\renewcommand\listoffigures{%
	\chapter*{\listfigurename}%
	\@mkboth{\scshape\listfigurename}%
	{\scshape\listfigurename}%
	\@starttoc{lof}%
}
\makeatother

\usepackage{amssymb}
\usepackage{cancel}

\usepackage{placeins}
\usepackage{array}
\usepackage{colortbl}
\usepackage{longtable}
\usepackage{ltcaption}
\usepackage{multirow}

\newcommand{\PreserveBackslash}[1]{\let\temp=\\#1\let\\=\temp}
\newcolumntype{C}{>{\centering\arraybackslash}X}
\newcolumntype{L}{>{\raggedright\arraybackslash}X}
\newcolumntype{A}[1]{>{\PreserveBackslash\raggedright}p{#1}}

\usepackage[T1]{fontenc}
\usepackage{lmodern} %
\rmfamily %
\DeclareFontShape{T1}{lmr}{b}{sc}{<->ssub*cmr/bx/sc}{}
\DeclareFontShape{T1}{lmr}{bx}{sc}{<->ssub*cmr/bx/sc}{}
\setlist[description]{ 
	labelwidth=0.6\textwidth,%
	leftmargin=\labelwidth, 
	align=right,
	font=\normalfont\scshape\bfseries
}

\usepackage{makecell}
\usepackage{mdframed}

\usepackage{environ}
\usepackage{tikz}
\usetikzlibrary{fit,backgrounds,calc}
\usepackage{algorithm}  %
\NewEnviron{defbox}[3]
{
	\vspace{15pt}\\
	\begin{tabular}{@{}lp{#3ex}@{}}
		\RaggedLeft{\textbf{\textsc{#1}}:} & #2
	\end{tabular}
	\vspace{15pt}\\
}

\usepackage[linktocpage=true]{hyperref}
\hypersetup{
	colorlinks=true,
	linktoc=all,
	linkcolor=black, 
	filecolor=black, 
	citecolor = gray!80!black,       
	urlcolor=gray!80!black,
}
\usepackage{blindtext}

\usepackage{algpseudocode}  %
 
\usepackage{etoolbox}
\makeatletter
\patchcmd{\ttlh@hang}{\parindent\z@}{\parindent\z@\leavevmode}{}{}
\patchcmd{\ttlh@hang}{\noindent}{}{}{}
\makeatother

\usepackage[round]{natbib}
\usepackage{bibentry}
\nobibliography*
\usepackage{etoolbox}
\robustify{\bibentry}

\usepackage{fancyhdr}
\usepackage{extramarks}
\renewcommand{\chaptermark}[1]{\markboth{\normalsize \rmfamily \thechapter.\ \scshape #1}{\normalsize \rmfamily \thechapter.\ \scshape #1}}
\renewcommand{\sectionmark}[1]{\markright{\normalsize \rmfamily \thesection.\ \scshape #1}}
\fancypagestyle{plain}{\fancyhead{}
	
}

\newcommand{\mysubsubsection}[2]{
	\let\orisubsubsectionmark\subsubsectionmark
	\renewcommand\subsubsectionmark[1]{}%
	\subsubsection[#1]{#2 \orisubsubsectionmark{#2}}
	\orisubsubsectionmark{#2}
	\let\subsubsectionmark\orisubsubsectionmark
}
\newcommand{\mysubsection}[2]{
	\let\orisubsectionmark\subsectionmark
	\renewcommand\subsectionmark[1]{}%
	\subsection[#1]{#2 \orisubsectionmark{#2}}
	\orisubsectionmark{#2}
	\let\subsectionmark\orisubsectionmark
}
\newcommand{\mysection}[2]{
	\let\orisectionmark\sectionmark
	\renewcommand\sectionmark[1]{}%
	\section[#1]{#2 \orisectionmark{#2}}
	\orisectionmark{#2}
	\let\sectionmark\orisectionmark
}
\newcommand{\mychapter}[3]{
	\ifx&#3&%
	\else
			\addtocontents{toc}{\vspace{15pt}{\centering \textbf{\scshape #3}\\ \vspace{-5pt}}}
	\fi
	\chapter[#1]{#2 \chaptermark{#2}}\vspace{-15pt}%
	\vspace{10pt}
}

\usepackage[prependcaption,textsize=tiny]{todonotes}
\reversemarginpar
\definecolor{othercolor}{HTML}{CC0000}
\definecolor{oraclecolor}{HTML}{8B8589}
\definecolor{rulecolor}{HTML}{DCDCDC}
\definecolor{ashgrey}{rgb}{0.7, 0.75, 0.71}

\usepackage{pifont}
\usepackage{xcolor}
\definecolor{mycolor}{HTML}{008000}%
\definecolor{indiagreen}{HTML}{138808}%
\definecolor{papaya}{HTML}{EE892F}%
\definecolor{mygreen}{HTML}{008000}%
\definecolor{mypurple}{HTML}{9966CC}%
\definecolor{cambrigeblue}{HTML}{A3C1AD}%
\definecolor{darkgreen}{HTML}{014421}
\definecolor{lightgreen}{HTML}{8FBC8F}

\NewEnviron{publicationinfo}[1]
{
	\begin{tikzpicture}
		\node[inner sep=0pt,
		draw=black!50!white,
		line width=1.5pt,
		fill=black!5!white,
		] (box) {\parbox[t]{\textwidth}
			{
				\begin{center}
				\begin{minipage}{0.9\textwidth}
						{\BODY}
					\par\smallskip
				\end{minipage}\hfill
			\end{center}
			}%
		};
		\par
	\end{tikzpicture}
}

\NewEnviron{researchquestions}[1]
{
	\noindent\begin{tikzpicture}
		\node[inner sep=0pt,
		draw=black!50!white,
		line width=1.0pt,
		fill=black!5!white,
		] (box) {\parbox[t]{\textwidth}
			{
				\begin{center}
					\begin{minipage}{0.9\textwidth}
						{\BODY}
						\par\smallskip
					\end{minipage}\hfill
				\end{center}
			}%
		};
		\par
	\end{tikzpicture}
}

\NewEnviron{contributionninfo}[1]
{
	\begin{tikzpicture}
		\node[inner sep=0pt,
		draw=darkgreen!50!white,
		line width=2.5pt,
		fill=darkgreen!20!white,
		] (box) {\parbox[t]{\textwidth}
			{
				\begin{center}
					\begin{minipage}{0.9\textwidth}
						{\BODY}
						\par\smallskip
					\end{minipage}\hfill
				\end{center}
			}%
		};
		\par
	\end{tikzpicture}
}

\usepackage{tipa}
\usepackage{combelow}

\newcommand{\divider}[1]{
	\addtocontents{toc}{\vspace{-5pt}{\large\begin{center}{\scshape#1}\end{center}}\vspace{-20pt}}
	\newpage
	\vspace*{\fill}
	\titlerule\vspace{5ex}
	\begin{spacing}{2.5}
		\begin{center}
			{\Huge\scshape #1}
		\end{center}
	\end{spacing}
	\thispagestyle{empty}
	\vspace{1ex}
	\titlerule
	\vspace*{\fill}
	\addtocounter{page}{-1}
}

\usepackage{cleveref}
\crefname{figure}{Figures}{Figures}
\Crefname{figure}{Figures}{Figures}

\definecolor{leannecolor}{HTML}{87A96B}

\newcommand{\model}{{\small \scshape MattNet}}
\newcommand{\smallmodel}{{\scriptsize \scshape MattNet}}
\newcommand{\qbert}{{\small QbERT}}
\newcommand{\locAtt}{{\small \scshape LocalisationAttentionNet}}
\newcommand{\locAttShort}{{\small \scshape Loc\-AttNet}}

\newcommand{\yoruba}{Yor\`{u}b\'{a}}

\newcommand{\image}[1]{\textit{#1}}
\newcommand{\word}[1]{``#1''}
\newcommand{\textLabel}[1]{{\scshape #1}}

\newcommand{\supportSet}{$\mathcal{S}$}

\newcommand{\matchingSet}{$M$}

\newcommand{\eg}{e.g.\ }
\newcommand{\Eg}{For example, }
\newcommand{\ie}{i.e.\ }
\newcommand{\Ie}{I.e.\ }

\newcommand{\me}{familiar--\underline{novel}}
\newcommand{\Me}{Familiar--\underline{novel}}
\newcommand{\meOther}{familiar--\underline{novel}$^*$}
\newcommand{\MeOther}{Familiar--\underline{novel}$^*$}
\newcommand{\familiar}{familiar--\underline{familiar}}
\newcommand{\Familiar}{Familiar--\underline{familiar}}
\newcommand{\familiarWithNovel}{\underline{familiar}--novel}
\newcommand{\FamiliarWithNovel}{\underline{Familiar}--novel}
\newcommand{\novel}{novel--\underline{novel}}
\newcommand{\Novel}{Novel--\underline{novel}}

\newcommand{\mscoco}{MS\,COCO}
\newcommand{\hubert}{HuBERT}
\newcommand{\VGG}{VGG~16}

\newcommand{\leanneNameNoSpace}{Leanne Nortje}
\newcommand{\leanneName}{Leanne Nortje\ }
\newcommand{\danName}{Dan Onea\cb{t}\u{a}\ }
\newcommand{\danNameNoSpace}{Dan Onea\cb{t}\u{a}}
\newcommand{\yevgenName}{Yevgen Matusevych\ }
\newcommand{\yevgenNameNoSpace}{Yevgen Matusevych}
\newcommand{\hermanName}{Herman Kamper\ }
\newcommand{\hermanNameNoSpace}{Herman Kamper}
\newcommand{\benjiName}{Benjamin van Niekerk\ }
\newcommand{\benjiNameNoSpace}{Benjamin van Niekerk}

\newcommand{\towardsVPKL}{1}
\newcommand{\VPKL}{5}
\newcommand{\fewshotInterspeech}{2}
\newcommand{\fewshot}{3}
\newcommand{\ME}{4}
\newcommand{\MultilingualME}{6}

\newcommand{\RQtowardsVPKL}{{\footnotesize{\scshape Research~Paper~\towardsVPKL}}}
\newcommand{\RQVPKL}{{\footnotesize{\scshape Research~Paper~\VPKL}}}
\newcommand{\RQfewshotInterspeech}{{\footnotesize{\scshape Research~Paper~\fewshotInterspeech}}}
\newcommand{\RQfewshot}{{\footnotesize{\scshape Research~Paper~\fewshot}}}
\newcommand{\RQME}{{\footnotesize{\scshape Research~Paper~\ME}}}
\newcommand{\RQMultilingualME}{{\footnotesize{\scshape Research~Paper~\MultilingualME}}}

\newcommand{\researchquestionone}{{\centering{\textit{Can we get a \ac{VGS} model capable of detecting and localising a keyword depicted by an image within speech from low-resource languages?}}}}
\newcommand{\researchqone}{\noindent{\scshape Research Question 1}: \researchquestionone}
\newcommand{\researchqonecentered}{\noindent{\textbf {\scshape Research Question 1}}: 
	\vspace{-\topsep}
	\begin{center}\researchquestionone\end{center}
}
\newcommand{\rqone}{{\footnotesize{\scshape Research~Question~1}}}

\newcommand{\researchquestiontwo}{{\centering{\textit{Can we get a \ac{VGS} model to learn words from a low-resource language using only a few word-image pairs?}}}}
\newcommand{\researchqtwo}{\noindent{\scshape Research Question 2}: \researchquestiontwo}
\newcommand{\researchqtwocentered}{\noindent{\textbf {\scshape Research Question 2}}: 
	\vspace{-\topsep}
	\begin{center}\researchquestiontwo\end{center}}
\newcommand{\rqtwo}{{\footnotesize{\scshape Research~Question~2}}}

\newcommand{\researchquestionthree}{{\centering{\textit{Does a \ac{VGS} word learning model replicate the mutual exclusivity bias observed in children?}}}}
\newcommand{\researchqthree}{\noindent{\scshape Research Question 3}: \researchquestionthree}
\newcommand{\researchqthreecentered}{\noindent{\textbf{\scshape Research Question 3}}: 
	\vspace{-\topsep}
	\begin{center}\researchquestionthree\end{center}
}
\newcommand{\rqthree}{{\footnotesize{\scshape Research~Question~3}}}

\newcommand{\researchquestionfour}{{\centering{\textit{Does multilingualism affect the mutual exclusivity bias exhibited by a \ac{VGS} word learning model similarly to the effect seen in children?}}}}
\newcommand{\researchqfour}{\noindent{\scshape Research Question 4}: \researchquestionfour}
\newcommand{\researchqfourcentered}{\noindent{\textbf{\scshape Research Question 4}}: 
	\vspace{-\topsep}
	\begin{center}\researchquestionfour\end{center}
}
\newcommand{\rqfour}{{\footnotesize{\scshape Research~Question~4}}}

\newcounter{contributionCount}
\newcommand{\researchContributionOne}{\ShortInTextTitle{Research Contribution 1:} A new keyword localisation task called \acf{VPKL} in which a query keyword depicted by an image should be detected and localised in a speech segment.}

\newcommand{\researchContributionTwo}{\ShortInTextTitle{Research Contribution 2:} Two new \ac{VGS} models capable of doing \ac{VPKL}. 
The first model has a multimodal attention mechanism that includes context vectors.
The second model, which uses few-shot learning and a simpler attention mechanism that excludes the context vectors, outperforms the first model and a visually grounded \ac{BOW} model that localises textual keywords on an English \ac{VPKL} task.}

\newcommand{\researchContributionThree}{\ShortInTextTitle{Research Contribution 3:} We are the first to apply a low-resource language, \yoruba, to the \ac{VPKL} task. 
Although the few-shot pair mining method leaves much room for improvement, the ground truth model shows great promise.}

\newcommand{\researchContributionFour}{\ShortInTextTitle{Research Contribution 4:} A new visually grounded few-shot word acquisition model called \model. 
This model uses a new pair mining method and outperforms an existing few-shot model when using fewer examples per class. }

\newcommand{\researchContributionFive}{\ShortInTextTitle{Research Contribution 5:} We are the first to apply a low-resource language, \yoruba, to a visually grounded few-shot word acquisition model.
By leveraging speech-image data from a well-resourced language, English, we find that the \yoruba\ model outperforms a purely English variant.}

\newcommand{\researchContributionSix}{\ShortInTextTitle{Research Contribution 6:} We use the \model\ model in the few-shot study and train it on spoken words and natural images from familiar classes, to establish whether it exhibits a \ac{ME} bias when prompted with novel spoken words and images. 
None of the previous computational \ac{ME} studies has considered \ac{VGS} models.
Besides finding that the model exhibits a \ac{ME} bias, we find that initialising the audio and vision branches with prior knowledge and using the InfoNCE loss results in the strongest \ac{ME} bias.}

\newcommand{\researchContributionSeven}{\ShortInTextTitle{Research Contribution 7:} We set up an English \ac{ME} dataset to train \ac{VGS} models and test whether these models exhibit the \ac{ME} bias.
	The training set consists of spoken English words and natural images for a set of familiar classes.
	The test set consists of spoken English words paired with images of isolated natural objects for the familiar classes and a set of novel classes.
}

\newcommand{\researchContributionEight}{\ShortInTextTitle{Research Contribution 8:} We are the first to investigate whether multilingualism affects the \ac{ME} bias in a \ac{VGS} model similarly to how multilingualism affects the \ac{ME} bias in children.
The monolingual English \model\ model shows a weaker \ac{ME} bias than the bilingual and trilingual \model\ models, thus proving that the models show the opposite \ac{ME} trend observed in children.
}

\newcommand{\researchContributionNine}{\ShortInTextTitle{Research Contribution 9:} We extend the English \ac{ME} dataset's training set to include Dutch and French spoken words.
	The resulting training set consists of natural images paired with spoken English, Dutch and French words for the familiar classes.
}

\newcommand{\researchpaper}[2]{\ShortInTextTitle{#1}: 
	\noindent\begin{center}
		\begin{minipage}{0.8\linewidth}
			\begin{center}
				#2
			\end{center}
		\end{minipage}
\end{center}
}

\newcommand{\researchpaperbox}[2]{\ShortInTextTitle{#1}: 
		\vspace{5pt}
		\begin{center}
			
			\begin{tikzpicture}
					\node[inner sep=0pt,
					draw=black!50!white,
					line width=1.0pt,
					fill=black!5!white,
					] (box) {\parbox[t]{\textwidth}
							{
									\begin{center}
											\begin{minipage}{0.8\textwidth}
													{
															\begin{center}
																	\vspace{10pt}
																	
																	#2
																\end{center}
														}
													\par\smallskip
												\end{minipage}\hfill
										\end{center}
								}%
						};
					\par
				\end{tikzpicture}
		
			\end{center}
}

\title{{\scshape Visually Grounded Speech Models for Low-resource Languages and Cognitive Modelling}}
\author{\leanneName}{\leanneName}
\faculty{Faculty of Engineering}
\degree{PhD (Electronic Engineering)}{Doctor of Philosophy (Electronic Engineering)}

\supervisor[c]{Prof.\ \hermanName}

\date{text}

\makenoidxglossaries
\DeclareAcronym{mattnet}{
	short = {\scshape MattNet} ,
	long  = \textbf{m}ultimodal \textbf{att}ention \textbf{net}work,
	alt = \textbf{M}ultimodal \textbf{Att}ention \textbf{Net}work
}
\DeclareAcronym{AI}{
	short = AI,
	long  = artificial intelligence,
	alt = Artificial Intelligence
}
\DeclareAcronym{LLM}{
	short = LLM,
	long  = large language model,
	alt = Large Language Model
}
\DeclareAcronym{DL}{
	short = DL,
	long  = deep learning,
	alt = Deep Learning
}
\DeclareAcronym{AE}{
	short = AE ,
	long  = autoencoder,
	alt = Autoencoder
}
\DeclareAcronym{CAE}{
	short = CAE,
	long  = correspondence autoencoder,
	alt = Correspondence Autoencoder
}
\DeclareAcronym{CPC}{
	short = CPC,
	long  = contrastive predictive coding,
	alt = Contrastive Predictive Coding
}
\DeclareAcronym{ME}{
	short = ME,
	long  = mutual exclusivity,
	alt = Mutual Exclusivity
}
\DeclareAcronym{VGS}{
	short = VGS,
	long  = visually grounded speech models,
	alt = Visually Grounded Speech Models
}
\DeclareAcronym{CELL}{
	short = CELL,
	long  = \textbf{c}ross-channel \textbf{e}arly \textbf{l}exical \textbf{l}earning,
	alt = Cross-channel Early Lexical Learning
}
\DeclareAcronym{VQ}{
	short = VQ,
	long  = vector quantised,
	alt = Vector Quantised
}
\DeclareAcronym{CLIP}{
	short = CLIP,
	long = Contrastive Language-Image Pre-training,
	alt = Contrastive Language-Image Pre-training
}
\DeclareAcronym{CNN}{
	short = CNN,
	long = Convolutional Neural Network,
	alt = convolutional neural network
}
\DeclareAcronym{VPKL}{
	short = VPKL,
	long = visually prompted keyword localisation, 
	alt = Visually Prompted Keyword Localisation
}
\DeclareAcronym{BOW}{
	short = BOW,
	long = Bag-of-Words,
	alt = bag-of-words
}
\DeclareAcronym{ASR}{
	short = ASR,
	long = automatic speech recognition,
	alt = Automatic Speech Recognition
}
\DeclareAcronym{YFACC}{
	short=YFACC,
	long=\yoruba\ Flickr Audio Caption Corpus,
	alt=\yoruba\ Flickr Audio Caption Corpus
}
\DeclareAcronym{CGN}{
	short=CGN,
	long=Gesproken Nederlands,
	alt=Gesproken Nederlands
}

\newglossaryentry{low_resource_language}
{
	name=Low-Resource Language,
	text=low-resource language,
	description={is a language with little transcribed speech resources necessary to develop technology for it},
	plural=low-resource languages
}

\newglossaryentry{few-shot_learning}
{
	name=Few-Shot Learning,
	text=few-shot learning,
	description={is a method in which a new class is learned from only a few labelled examples~\citep{li_fei-fei_bayesian_2003, fei-fei_one-shot_2006, lake_one_2011, lake_one-shot_2014, koch_siamese_2015, vinyals_matching_2016, shyam_attentive_2017, snell_prototypical_2017}}
}

\newglossaryentry{multimodal_few-shot_learning}
{
	name=Multimodal Few-Shot Learning,
	text=multimodal few-shot learning,
	description={is a method in which a new class is learned from a few multimodal example pairs. Here, each example pair consists of two samples depicting the same class but the samples are from different modalities~\cite{nortje_direct_2020}}
}

\newglossaryentry{mutual_evclusivity}
{
	name= Mutual Exclusivity,
	text=mutual exclusivity (ME),
	description={is a bias that states once an object has a name, it does not need another. As a result, a novel word should belong to a novel image~\citep{markman_childrens_1988}.}
}

\newglossaryentry{keyword_spotting}
{
	name= Keyword Spotting,
	text=keyword spotting,
	description={is the task ranking spoken utterances in a set by how likely it is that a query keyword occurs in each utterance~\citep{wilpon_automatic_1990}.}
}

\newglossaryentry{keyword_detection}
{
	name= Keyword Detection,
	text=keyword detection,
	description={is the task of detecting whether a query keyword occurs in a spoken utterance~\citep{wilpon_automatic_1990}.}
}

\newglossaryentry{keyword_localisation}
{
	name= Keyword Localisation,
	text=keyword localisation,
	description={is the task of locating where in a spoken utterance a given query keyword occurs if the query is detected in the utterance~\citep{olaleye_attention-based_2021}.}
}

\newglossaryentry{vpkd}
{
	name= Visually Prompted Keyword Detection,
	text=visually prompted keyword detection (VPKD),
	description={is the task of detecting whether a keyword depicted by an image, occurs in a spoken utterance.}
}

\newglossaryentry{vpkl}
{
	name= Visually Prompted Keyword Localisation,
	text=visually prompted keyword localisation (VPKL),
	description={is the task of localising where in a spoken utterance a keyword depicted by an image occurs if the keyword was dected in the utterance (VPKD).}
}
\newglossaryentry{multilingual_transfer}
{
	name= Multilingual Transfer,
	text= multilingual transfer,
	description={is a method that leverages knowledge gained from one or more languages to develop systems for an unseen language~\citep{kamper_improved_2021, hu_multilingual_2020, jacobs_acoustic_2021, kamper_multilingual_2020, ma_acoustic_2021, wang_extending_2020, hermann_multilingual_2021}.}
}

\newglossaryentry{vgs}
{
	name= Visually Grounded Speech Models,
	text= \ac{VGS},
	description={are a specific form of multimodal modelling in which a model is trained simultaneously on speech and image inputs~\citep{harwath_deep_2015, kamper_visually_2017, chrupala_representations_2017}.}
}

\begin{document}

\graphicspath{{frontmatter/fig/}}

\TitlePage

\clearpage
\vspace*{\fill}
\thispagestyle{empty}

\begin{center}
	\begin{minipage}{0.6\linewidth}
		\begin{center}
			\textit{To Pieter, \\for your love and \\unwavering support.}
		\end{center}
	\end{minipage}
\end{center}

\vspace*{\fill} 
\clearpage

\pagenumbering{roman}
\chapter*{Acknowledgements}
\makeatletter\@mkboth{}{Acknowledgements}\makeatother

I would like to thank my official supervisor, Prof.\ \hermanNameNoSpace, for his mentorship and guidance over the past six years, I learnt a lot from you and have the deepest respect for your expertise. 
Your knowledge and insights have been instrumental in shaping my research, and I feel fortunate to have had you as my supervisor.

To my other but unofficial supervisor Dr \danNameNoSpace, without your help especially with analysis and design, there would not have been a PhD dissertation today.
Your analysis figures are out of this world!
There aren't words to convey my gratitude for all your support.

A special thanks to Dr \yevgenNameNoSpace, for helping us with publishing in the computational cognitive field and doing the significance analysis. 
Your insights and your willingness to be involved in the cognitive papers is much appreciated. 

During my PhD, I was fortunate to have received a Google DeepMind Scholarship and a CSIR grant for one year. 
I would also like to thank the South African Department of Science and Innovation for awarding me with a South African Women in Science award in the category DSI-Ndoni Mcunu Fellowships to fund some of the PhD costs.

A shoutout to Kayode Olaleye for developing the \acf{YFACC} dataset that proved to be invaluable for this dissertation. 
\benjiNameNoSpace\ for all the brainstorming sessions when I got stuck on a problem and that we could use your \qbert\ brain child. 

I would also like to thank my examiners, Prof. David Harwath, Prof. Richard Klein and Dr Trienko Grobler, for their invaluable feedback, insightful comments, and constructive criticism. 
I am grateful for the time and effort they invested in reviewing my dissertation.

To Pieter, thank you for your patience, love and support this last two years. 
Without you, this PhD journey would have been cut short!

I would like to thank the heavenly Father for his grace and love throughout this journey.

\DeclarationDate{15 May 2024}
\DeclarationPage

\chapter*{Abstract}
\addcontentsline{toc}{chapter}{Abstract}
\makeatletter\@mkboth{}{\scshape Abstract}\makeatother

\Ac{VGS} learn from unlabelled speech paired with images.
Such models can be valuable to develop speech applications for low-resource languages lacking transcribed data, and understanding how humans acquire language since children learn speech from multimodal cues. 
This dissertation makes contributions to both of these areas.

In the first part of this dissertation, we consider two research questions about using VGS models in low-resource language applications.	
The first research question asks: can we get a \ac{VGS} model that can detect and localise a keyword depicted by an image within speech? 
For this, we propose a new task called \ac{VPKL}. 
Here, an image depicting a keyword query should be detected in spoken utterances. 
A detected query should be localised within the utterance.
To do \ac{VPKL}, we modify a common \ac{VGS} modelling approach using an acoustic and a vision network connected with a multimodal attention mechanism.
On an artificial low-resource language, English, we find that using an ideal tagger to get training pairs outperforms a previous visual \ac{BOW} model locating written keywords in spoken utterances. 
An actual visual tagger results in lower scores than the written keyword baseline.
To do \ac{VPKL} for a real low-resource language, we consider few-shot learning before returning to this problem.

In the second research question, we ask if we can get a \ac{VGS} model to learn words using only a few word-image pairs.
We use an architecture similar to the \ac{VPKL} model's and combine it with a few-shot learning approach that can learn new classes from fewer natural word-image pairs. 
Using the few given word-image example pairs, new unsupervised word-image training pairs are mined from large unlabelled speech and image sets.
Our approach outperforms an existing \ac{VGS} few-shot model when the number of examples per class is small. 
As a result, we apply this approach to an actual low-resource language – \yoruba. 
The \yoruba\ few-shot model outperforms its English variant.

From the few-shot progress we make, we return to the \ac{VPKL} approach and propose a simpler model similar to our previous \ac{VPKL} model.
Here we assume we have access to a dataset consisting of spoken utterances paired with descriptive images. 
To mine speech-image training pairs for a keyword, we use a few spoken word examples of the keyword and compare them to the utterances in the dataset's speech-image pairs.
We found that this approach outperforms our previous approach on an English \ac{VPKL} task and the visual \ac{BOW} model that detects textual keywords in speech. 
As a result, we apply this approach to \yoruba. 
Since the speech system in the pair mining scheme uses a model trained on English, the precision of the few-shot \yoruba\ localisation model is low. 
However, the ground truth \yoruba\ model outperforms the textual keyword localisation model applied to \yoruba\ by large margins.

In the second part of this dissertation, we ask two more research questions regarding the use of \ac{VGS} models in computational cognitive studies.
Our third research question considers whether a \ac{VGS} model exhibits the \ac{ME} bias which is a word learning constraint used by children.
This bias states that a novel word belongs to an unknown object instead of a familiar one. 
To do this, we use our few-shot object and word learning model and generate a speech-image dataset containing spoken English word and image examples for a set of familiar and novel classes. 
The model is trained on the word-image pairs for the familiar classes.
The model is then prompted with novel English spoken words and asked whether the words belong to unknown or familiar objects. 
All variants of the model exhibit the \ac{ME} bias.
A model that uses both self-supervised audio and vision initialisations has the strongest \ac{ME} bias. 
This makes sense from a cognitive perspective since children are exposed to spoken language and visual stimuli in their surroundings when they begin using the \ac{ME} bias.

Various cognitive \ac{ME} studies have considered the effect that factors such as multilingualism have on the \ac{ME} bias.
Since this effect has not yet been studied computationally, our fourth research question asks how multilingualism affects the \ac{ME} bias exhibited by our \ac{VGS} model.
We extend the English \ac{ME} dataset’s training set to contain spoken Dutch and French words for the familiar classes. 
We train a trilingual English-Dutch-French model and two bilingual models: an English-Dutch model and an English-French model. 
These multilingual models are compared to the monolingual English model of the previous research question. 
We find that the monolingual model has a weaker \ac{ME} bias than multilingual models. 
This trend is opposite to the trends seen in children: monolingual children have a stronger \ac{ME} bias than multilingual children.
This study is preliminary and requires further investigation. 

In summary, we find that \ac{VGS} models can be used to develop low-resource applications by using only a small set of ground truth examples. 
We also found that \ac{VGS} models can be used to computationally study the \ac{ME} bias observed in children. 
Further investigation is required into the effect of multilingualism on the bias in \ac{VGS} models and comparing it to the effect in children. 
We believe this dissertation has given enough proof of how valuable VGS models can be and will encourage research in this field to build inclusive speech technology and contribute to understanding human language learning. 
\selectlanguage{english}

\acuseall
\clearpage
\renewcommand{\contentsname}{Table of Contents}
\addcontentsline{toc}{chapter}{Table of contents}
\tableofcontents
\newpage
\addcontentsline{toc}{chapter}{List of figures}
\listoffigures
\addcontentsline{toc}{chapter}{List of tables}
\listoftables
\clearpage

\chapter*{Nomenclature\markboth{}{Nomenclature}}
\addcontentsline{toc}{chapter}{Nomenclature}

\section*{Acronyms and Abbreviations}
\addcontentsline{toc}{section}{Acronyms and abbreviations}

\printacronyms[display=used]

\glsaddall
\newpage
\renewcommand\glossaryname{Definitions}
\printnoidxglossaries

\newpage

\pagenumbering{arabic}
\acresetall

\mychapter{Introduction}{Introduction}{}
\label{chap:introduction}

Multimodal models learn from data streams coming from various different modalities.
\Ac{VGS} are a form of multimodal modelling in which a model is trained simultaneously on speech and image inputs~\citep{harwath_deep_2015, kamper_visually_2017, chrupala_representations_2017}.
These models have recently gained interest since vision can be used as a weak form of supervision to train speech models instead of textual transcriptions~\citep{chrupala_visually_2022}.
Visual grounding is just one method of obtaining speech models without using textual resources.
Other methods include self-supervised~\citep{oord_representation_2019} and unsupervised learning~\citep{kamper_segmental_2017}.

Speech processing models have shown remarkable progress in recent years~\citep{makridakis_large_2023}. 
However, the performance of these models is greatly dependent on the size of the training dataset~\citep{besacier_automatic_2014}, which presents two obstacles.
Firstly, large datasets are scarce since collecting and transcribing data is expensive and time-consuming~\citep{zhao_survey_2023}. 
Secondly, for some languages, there are little or no data resources for the language~\citep{de_sa_category_1998}. 
In extreme cases, there is no written form for a language~\citep{scharenborg_linguistic_2018, lupke_research_2010, bird_sparse_2021, scharenborg_speech_2020, wang_generating_2021}. 
As a result, speech technologies rarely support these \glspl{low_resource_language}.

Arguably, images with spoken descriptions or videos -- speech paired with a continuous sequence of images -- would be easier to collect for low-resource languages than transcribing large amounts of speech~\citep{olaleye_visually_2023}.
Previous work has shown that \ac{VGS} models can be used to detect keywords within spoken segments~\citep{olaleye_keyword_2022}, learn syllable~\citep{peng_syllable_2023} and word~\citep{peng_word_2022} boundaries, and even learn semantic connections that unimodal speech models cannot~\citep{kamper_semantic_2019-1}.
As a result, our first goal is to consider \ac{VGS} models to obtain systems for low-resource languages where datasets are limited.

Our second goal is to use \ac{VGS} modelling in a cognitive study.
The idea of learning from speech and complementary visual signals is inspired by how children learn. 
Children acquire new words from just a few examples without the use of transcriptions~\citep{biederman_recognition-by-components:_1987, miller_how_1987, gomez_infant_2000, lake_one-shot_2014, rasanen_joint_2015}.
\Eg \cite{borovsky_once_2012} found that children can acquire a word for a visual object after seeing it only once.
Additionally, children must infer which unknown spoken words refer to which objects they encounter in their environment.
To do this, we investigate whether a \ac{VGS} word learning model uses the same word learning constraint as children.

\mysection{Motivation}{Motivation}

\Ac{VGS} modelling aims to learn from co-occurring speech and vision signals without textual transcriptions~\citep{chrupala_visually_2022}, making it ideal for many settings where using text is not possible. 
Two useful scenarios for using \ac{VGS} models, which we also consider in this dissertation, are cognitive modelling and low-resource technology development. 	

\mysubsection{Low-resource technology}{Low-resource Technology}

Low-resource languages have limited or no textual resources. 
If we can obtain \ac{VGS} models for low-- or zero-resource languages, we can attempt to build speech technologies like \ac{ASR} systems for these languages.
As a result, technologies like Amazon's Alexa and Google Home can support a more comprehensive range of languages for their customers without having to collect expensive resources.

In Chapter~\ref{chap:localisation}, we consider using \ac{VGS} models that take in entire speech segments and natural images to do \gls{keyword_localisation} for an artificial low-resource language, English, and an actual low-resource language, \yoruba.
With keyword localisation, a given query keyword should be detected and located within a spoken utterance~\citep{olaleye_attention-based_2021}.
We introduce a new \ac{VPKL} task, replacing the query keyword with an image depicting the keyword, a so-called image query.
With this, we can search through speech collections for spoken utterances containing instances of an image query. 
Such systems can be used in humanitarian relief programmes orchestrated by the United Nations for African regions with minimal speech resources~\citep{van_der_westhuizen_feature_2022}.

To obtain low-resource \ac{VGS} models like the \ac{VPKL} models, we consider methods that make use of a small set of paired speech and image data.
Few-shot learning is a method that emulates the speed and efficiency with which children learn new classes from only a few word or object examples.
Concretely, \gls{few-shot_learning} is a method in which a new class is learned from only a few labelled examples~\citep{li_fei-fei_bayesian_2003, fei-fei_one-shot_2006, lake_one_2011, lake_one-shot_2014, koch_siamese_2015, vinyals_matching_2016, shyam_attentive_2017, snell_prototypical_2017}. 
In Chapter~\ref{chap:localisation}, we collect a small set containing a few spoken word examples per keyword class.
We use this set to sample spoken utterances for training that possibly contain the few-shot classes we want to learn. 

The few-shot learning method can be extended to multimodal few-shot learning, which entails learning new classes from paired data.
Each pair consists of two or more samples, each from a different modality~\citep{nortje_direct_2020}.
In the case of multimodal speech-image few-shot learning, new classes are learned from spoken words paired with images representing each class.
In Chapter~\ref{chap:few-shot}, we use a small ground truth set of isolated spoken words paired with images -- which can be easily collected -- to learn words in an artificial low-resource language, English, and an actual low-resource language, \yoruba.
We use this small set, a large unlabelled unimodal image set and a large unlabelled unimodal speech dataset in the low-resource language.
This model acts as a first step in learning words from a low-resource language by collecting a small ground truth multimodal set, using large existing image datasets and collecting unlabelled speech segments from various sources.
This technique of using a few ground truth speech segments and images with no text could be expanded to low-resource models with various downstream applications.

\mysubsection{Computational cognitive modelling}{Computational Cognitive Modelling}

The \ac{VGS} models we use in Chapter~\ref{chap:localisation} and Chapter~\ref{chap:few-shot} could also be used to test cognitive hypotheses surrounding infant language acquisition.
Such models could help to gain insights into learning disabilities and ways to improve treatments for learning problems.
More specifically, in Chapter~\ref{chap:me}, we use the \ac{VGS} model of Chapter~\ref{chap:few-shot} to test the findings of \citet{byers-heinlein_monolingual_2009}.
They study the observation seen in young children known as \gls{mutual_evclusivity}: a tendency to associate a novel noun with a novel object instead of a familiar one.  
After showing that our model shows similar ME tendencies to children, we attempt to test whether language experience such as multilingualism has the same influence on the bias as seen in  children~\citep{byers-heinlein_monolingual_2009}: monolingual children of 17 months of age make use of the \ac{ME} bias to a great extent, whereas bilingual children use it marginally and trilingual infants did not make use of it at all.

\mysection{Visually grounded speech modelling overview}{Visually Grounded Speech Modelling Overview}
\label{sec:vgs}

Before we get to our research questions regarding the use of \acf{VGS} for low-resource language technology and cognitive studies, we give an overview of the approaches commonly used in \ac{VGS} modelling that influence our design choices.
\Ac{VGS} modelling gained interest with the study of \citet{roy_learning_1999}, who proposed \ac{CELL}. 
Inspired by how children learn words from multiple sensory inputs like visual interaction, \ac{CELL} aims to learn words and their meanings. 
Specifically, they aim to learn shape and colour classes from raw speech and image signals.
\citet{chrupala_visually_2022} gives an extensive overview of the development of \ac{VGS} modelling from this study to where we are today.

\mysubsection{The general VGS model architecture}{The General VGS Model Architecture}
\label{subsec:general_structure}

\begin{figure}[tb]
	\centering
	\includegraphics[width=\linewidth]{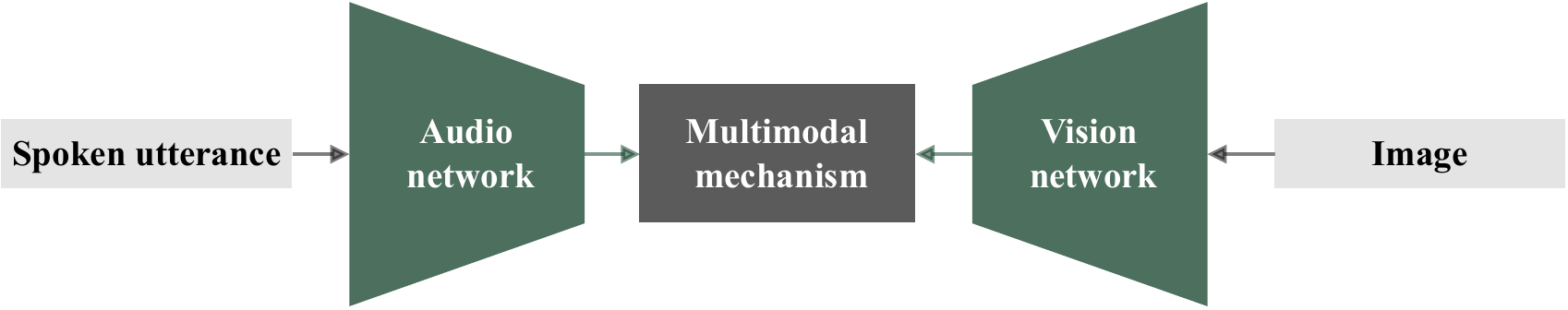}
	\caption{
		Most \ac{VGS} studies use the general model structure consisting of an audio and a vision network connected with a multimodal mechanism.
	}
	\label{fig:general_architecture}
\end{figure} 

Looking back through the development of \ac{VGS} modelling, it is notable that most models use the same general structure: a vision branch and a speech branch connected using a multimodal structure, \eg a multimodal triplet objective function. 
To illustrate the use of this general structure shown in Figure~\ref{fig:general_architecture}, we give a brief overview of a few example studies.
\citet{ngiam_multimodal_2011} use an autoencoder-like network for both the vision and audio branches with a multimodal objective connecting the bottleneck representations of the speech and vision encoders.
Similarly, \citet{gong_contrastive_2023} extended the unimodal \ac{AE} using the masked learning objective of \cite{he_masked_2022} to the multimodal setting to obtain a \ac{VGS} model.
\citet{scholten_learning_2021}  use cosine similarity between an image embedding and an audio embedding to retrieve images containing the visual referents of a word.
\citet{petridis_end--end_2018} also use a vision branch to extract features from an image and an audio branch to extract features from audio waveforms.
They specifically use images containing mouth regions as a supplementary signal for word recognition on speech.
\citet{synnaeve_learning_2014} also aimed to do word recognition using this general structure, but their goal was to learn semantically relevant words.

The FaST-VGS model of \citet{peng_fast-slow_2022} uses separate vision and speech branches consisting of convolutional and transformer layers.
These branches are connected with a complex multimodal attention mechanism to learn which words correspond to which parts of an image.
This attention mechanism was developed from an earlier study that proposed the matchmap~\citep{harwath_jointly_2018}.
A matchmap is an attention mechanism that calculates the dot product between each embedding in a sequence of per frame audio embeddings and each embedding in a sequence of per pixel image embeddings.
Concretely, to get the sequence of pixel embeddings, our image encoder extracts a grid of 2048-dimensional embeddings where each of these embeddings corresponds to a region in the image.
In Chapter~\ref{chap:localisation}, we use this matchmap attention mechanism, whereas the remaining chapters use variations of this mechanism to suit the various applications we consider.
\citet{peng_self-supervised_2022} showed that the attention mechanism used by \citet{peng_fast-slow_2022} in FaST-VGS, together with a masked learning objective, results in a model that can learn the semantics, phonetics and syntax of a language.
Instead, we use a contrastive learning objective and show that this objective and variations of the matchmap attention mechanism result in models that can learn spoken word classes, learn if and where spoken words occur in an utterance, and emulate word learning constraints used by children.

To do image-to-speech or speech-to-image retrieval, \citet{harwath_deep_2015}  use separate vision and audio branches to encode speech and image inputs into fixed-dimensional embeddings.
These embeddings are used in a multimodal hinge loss, which assigns a high similarity score to a speech and image instance if the spoken utterance is a description of the image and a low similarity score otherwise.
Our contrastive objective aims to do something similar: push speech and image representations from the same class closer and push speech and image representations from different classes away from one another. 
In some instances, we also use sequences of embeddings instead of a single embedding to represent a speech input. 

\mysubsection{Taking advantage of state-of-the-art unimodal speech or vision models}{Taking Advantage of State-of-the-Art  Unimodal Speech or Vision Models}
\label{subsec:unimodal_advantage}

Various studies use state-of-the-art unimodal models to find embeddings in the multimodal space for speech and image inputs that represent the inputs’ class.
\citet{harwath_unsupervised_2016} and \citet{harwath_learning_2020} use %
a hinge loss with the general \ac{VGS} architecture shown in Figure~\ref{fig:general_architecture} to get a single fixed-dimensional embedding to represent an image or a speech utterance.
\cite{harwath_learning_2020} add \ac{VQ} layers between the convolutional layers of the audio branch to learn discrete speech units between intermediate layers.
On the image side, both \citet{harwath_unsupervised_2016} and \citet{harwath_learning_2020} took advantage of existing vision models like \VGG~\citep{simonyan_very_2015} and ResNet50~\citep{he_deep_2016}, respectively.

Recently, studies have taken advantage of advances made in self-supervised speech models.
\cite{peng_word_2022} use either wav2vec2.0 or \hubert\ as the audio branch of their \ac{VGS} model called  VG-\hubert.
After training their VG-\hubert\ model on large amounts of images paired with spoken descriptions, they found that the audio branch encodes input speech into valuable acoustic units.
Moreover, \cite{peng_syllable_2023} found that this model can segment input speech into syllables, leading us to use a self-supervised speech model in all our proposed models.

\ac{VGS} models such as \citet{peng_word_2022} and \citeauthor{peng_syllable_2023}'s~(\citeyear{peng_syllable_2023}) are trained to retrieve the specific image corresponding to a spoken description.
However, with these models, it is difficult to classify the word, syllable or sub-word unit corresponding to the segments into which the model naturally divides input speech. 
Since we aim to either learn the spoken and visual representations of a specific word class or to detect and localise the spoken form of a visual object within a spoken utterance, we need to know which word a speech segment represents. 
Therefore, our objective function is class-specific rather than retrieval-based. 
Concretely, for a specific class, we take any spoken word representing the class, an image containing the class or an utterance containing the spoken word of the class as a positive example.
In contrast, the speech-image retrieval objective sees all other speech-image pairs as negatives to an anchor speech-image pair regardless of whether these negative pairs contain the same word and object classes as the anchor.
Nevertheless, these models illustrated the benefits of using a self-supervised speech model. 	

\citet{shih_speechclip_2023} use existing unimodal models for both the vision and audio branches. 
The vision branch uses the image encoder of the \ac{CLIP} model~\citep{radford_learning_2021}.
The speech branch uses the \ac{CLIP} text encoder, which takes in \hubert\ features instead of text representations. 
In this way, they adapt \ac{CLIP} to work on spoken language instead of text: the \ac{CLIP} encoders stay frozen, forcing the \hubert\ encoder to learn how to output discrete pseudo-text units.
\citet{berry_m-speechclip_2023} removed this English text encoder to extend the method to work on multiple languages.
Since our goal is to find \ac{VGS} models to build systems for low-resource languages in Chapter~\ref{chap:localisation} and Chapter~\ref{chap:few-shot}, we, similar to \cite{berry_m-speechclip_2023} do not use textual encoders.
But similarly to \cite{shih_speechclip_2023} and \cite{berry_m-speechclip_2023}, we use existing state-of-the-art unimodal speech and vision models to initialise our model branches.

\mysubsection{Multilingual VGS modelling}{Multilingual VGS Modelling}
\label{subsec:multilingual_VGS}

One of our ultimate goals is to use \ac{VGS} modelling to develop speech applications for low-resource languages.
In the unimodal speech domain, various studies use a method called multilingual transfer in which knowledge gained from models trained on well-resourced languages are used to develop models for low-resource languages~\citep{kamper_improved_2021, hu_multilingual_2020, jacobs_acoustic_2021, kamper_multilingual_2020, ma_acoustic_2021, wang_extending_2020, hermann_multilingual_2021}.
Next, we look at a few examples that use multilingual transfer in \ac{VGS} models that led us to use a well-resourced language, English, to build a model that can learn the visual depictions of spoken words in a low-resource language in Chapter~\ref{chap:few-shot}.

\citet{azuh_towards_2019} train separate English and Hindi audio branches with a vision branch consisting of the \VGG\ network to learn word-like units indirectly through the task of learning word translations.
\citet{havard_models_2019} trained an attention-based \ac{VGS} model on two languages, English and Japanese, to perform cross-lingual speech-to-speech retrieval.
\citet{ohishi_trilingual_2020} use a trilingual semantic model for cross-lingual retrieval.
They train the model to learn a class's visual depictions and its English, Japanese and Hindi spoken word representations.
\citet{ohishi_pair_2020} try to do the same on only English and Japanese by using a dataset containing images and spoken descriptions for each language.
To get such a dataset, they propose a pair expansion algorithm to link disjoint datasets from the different languages: the images in each dataset are compared using image embeddings or object recognition techniques.
If any two images are similar enough, the images and the spoken utterances from different languages paired with these images are used to make new pairs. 

In Chapter~\ref{chap:localisation} and Chapter~\ref{chap:few-shot}, we make use of pair mining schemes.
Chapter~\ref{chap:localisation} uses a similar pair mining scheme as \citet{ohishi_trilingual_2020}, but instead, we use speech comparisons from a query-by-example system to find spoken utterance and image pairs in which the few given ground truth spoken word examples of a class, are predicted to occur.
However, in Chapter~\ref{chap:few-shot}, we use a few ground truth examples to pair up speech to images, and obtain image-image and speech-speech pairs.
Since our ultimate goal is to develop \ac{VGS} models for low-resource languages, we assume two things.
We can collect a reasonably sized speech-image dataset for a low-resourced language from native speakers where no transcriptions are necessary. 
Or we can collect a small speech-image set that contains only a few examples per class, as well as large unlabelled speech datasets which we can use together with large image datasets  that are well-resourced and readily available.	

\mysubsection{Using VGS models for low-resource languages}{Using VGS Models For Low-Resource Languages}
\label{subsec:low_resource_VGS}

From the above discussions, it is evident that recent work showed that \ac{VGS} models can be used to learn words and semantics without any textual supervision.
In actual low-resource settings, we often have access to a limited amount of transcribed speech. 
\citet{pasad_contributions_2019} study how visual grounding can be utilised when there are varying amounts of textual supervision available. 
They found visual grounding useful for semantic retrieval even with textual supervision.
Although we also aim to develop \ac{VGS} models for low-resource languages, we do not use any text.
Instead, we assume that it is possible to find native speakers to give spoken descriptions for a few images or to give a few spoken word examples and images for a word class.

\citet{olaleye_yfacc_2023} also assume that images with spoken descriptions can more easily be collected than speech with textual transcriptions. 
They use such a speech-image dataset that they collect for a low-resource language, \yoruba, to build a visually grounded \ac{BOW} model.
Their goal is to detect and localise written English keywords in \yoruba\ speech.
We consider the same task in Chapter~\ref{chap:localisation}, but we use an image depicting a keyword instead of taking a textual keyword from a well-resourced language.
We also use their \yoruba\ dataset to build our low-resource applications in Chapter~\ref{chap:localisation} and Chapter~\ref{chap:few-shot}.

\mysection{Research questions}{Research Questions}
\label{research_questions}

\begin{figure}[tb]
	\centering
	\includegraphics[width=0.9\linewidth]{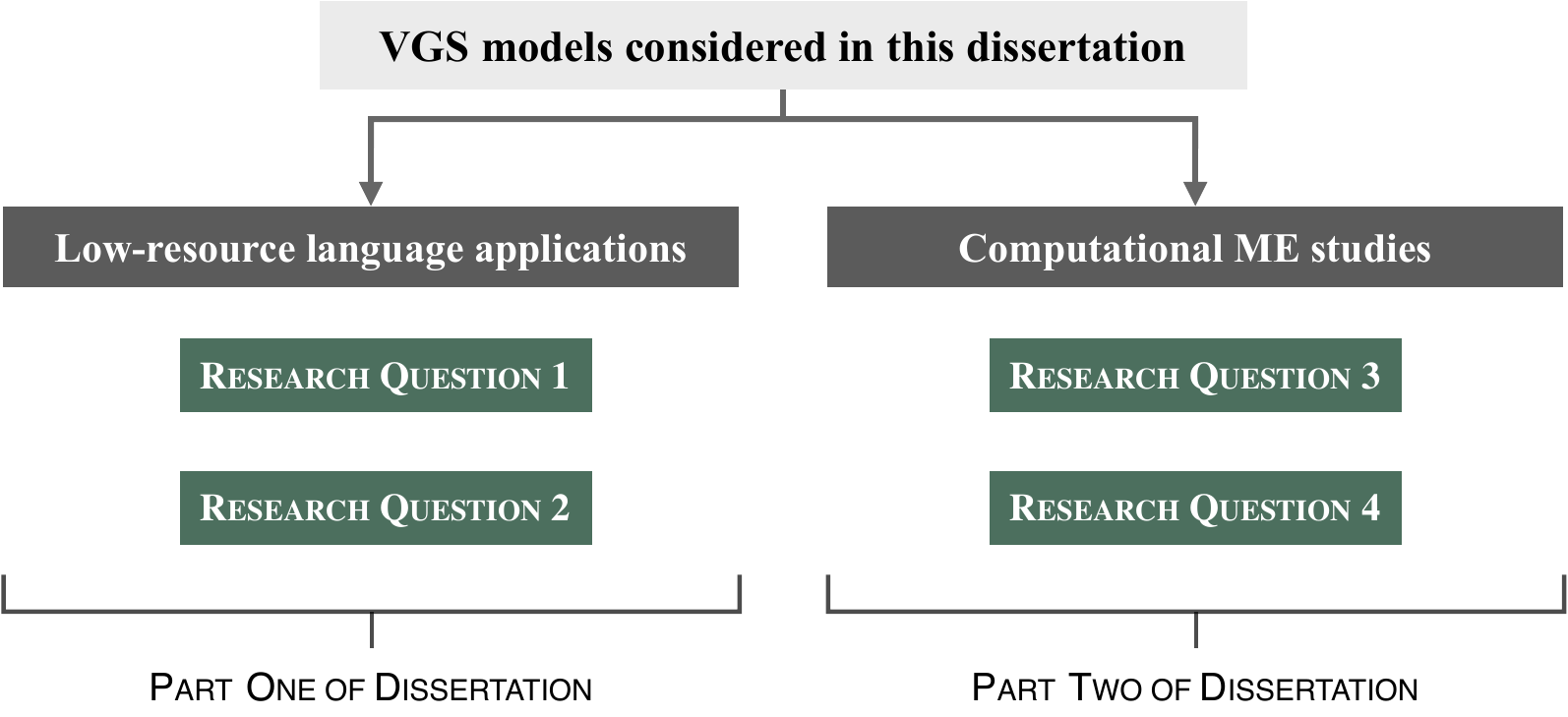}
	\caption{
		The \ac{VGS} models in this dissertation can be organised into two high-level categories: low-resource applications and computational \ac{ME} studies. 
		As a result, \rqone\ and \rqtwo\ are placed under low-resource applications, and \rqthree\ and \rqfour\  are placed under computational \ac{ME} studies. 
	}
	\label{fig:outline}
\end{figure} 

In this dissertation, we consider four research questions regarding the applications of \ac{VGS} models.
The first two questions entail low-resource speech applications of \ac{VGS} models.
The last two questions deal with using \ac{VGS} models to computationally study the \acf{ME} bias employed by children to learn new words.
As a result, this dissertation is divided into two high-level parts, as shown in Figure~\ref{fig:outline}.
Next, we flesh out these questions, where each question is considered in a content chapter.

\mysubsection{Research questions on low-resource applications}{Research Questions on Low-Resource Applications}
\noindent\begin{researchquestions}
	
	{\textbf \researchqone}
	
\end{researchquestions}

\Gls{keyword_localisation} entails detecting if a query keyword occurs in a spoken utterance and, if the keyword occurs, locating where in the utterance the keyword occurs~\citep{olaleye_attention-based_2021}.
\citet{olaleye_yfacc_2023} considered keyword localisation for a low-resource language, \yoruba, using a \ac{VGS} model developed by \citet{olaleye_attention-based_2021} and \citet{olaleye_keyword_2022}.
More specifically, they attempt to localise an English textual keyword in a \yoruba\ spoken utterance.

We modify this keyword localisation task by introducing a new visual keyword localisation task called \gls{vpkl}\acuse{VPKL}.
A model is given an image depicting a keyword, a so-called image query, instead of a textual keyword.
The model should then detect whether the image query occurs in a spoken utterance and if it occurs where the query occurs in the utterance.
A visual query is more flexible than a textual or a spoken keyword query since it can allow the search for semantically related words. 
A \ac{VPKL} model, compared to a \ac{BOW} approach, can search for words outside of the training vocabulary.
We propose a \ac{VGS} model capable of comparing images to spoken utterances to answer whether we can find a model capable of detecting and localising image queries in spoken utterances.

\vspace{2ex}
\noindent\begin{researchquestions}
	
	{\textbf \researchqtwo}
	
\end{researchquestions}

Children can quickly and with great accuracy learn new concepts from only a few word or object examples~\citep{biederman_recognition-by-components:_1987, miller_how_1987,gomez_infant_2000,rasanen_joint_2015}.
Few-shot learning is a machine learning method that emulates the efficiency with which children learn.
More specifically, with \gls{few-shot_learning}, a model learns a new class from only a few labelled examples~\citep{li_fei-fei_bayesian_2003}.
In \gls{multimodal_few-shot_learning}, a new class is learned from a few multimodal example pairs, each consisting of two samples from different modalities but the same class~\citep{eloff_multimodal_2019}.
Any two modalities can be used for multimodal few-shot learning, but by using speech together with images, we can obtain a system that learns spoken word classes without using textual transcriptions.
\Ie speech-image few-shot learning removes the obstacle of obtaining textual transcriptions that are impossible or hard to come by for low-resource languages.
With this, we attempt to answer whether we can learn spoken words from a low-resource language by using images as a weak form of supervision instead of transcriptions.

\mysubsection{Research questions on computational studying the mutual exclusivity bias}{Research Questions on Computationally Studying the Mutual Exclusivity Bias}
\noindent\begin{researchquestions}
	
	{\textbf \researchqthree}
	
\end{researchquestions}

The \ac{ME} bias is a constraint children use to map a novel word they hear to an object for which they do not know a name  rather than to a familiar object~\citep{markman_childrens_1988}.
After the rise of neural architectures in the last few years, we saw renewed interest in the \ac{ME} bias from the computational modelling perspective.
Several studies have examined whether the \ac{ME} bias emerges in machine learning models and under which conditions~\citep{gulordava_deep_2020, gandhi_mutual_2020, vong_cross-situational_2022, ohmer_mutual_2022}.
All these models receive word and object representations as inputs since the ME strategy applies to learning mappings between words and the objects they refer to. 
The object representations vary in complexity, from symbolic representations of a single object class~\citep[e.g.,][]{gandhi_mutual_2020} to continuous vectors representing a natural image consisting of many objects where one object is from the class under consideration~\citep[e.g.,][]{vong_cross-situational_2022}.  
However, all these computational models take the invariable textual representation of a word as input.

Children, however, do not  learn words from discrete textual representations but from continuous speech; there are large variations in how a word is realised due to factors such as word duration, prosody, intonation, etc. 
Therefore, children face an additional challenge of making sense of spoken continuous signals compared to models trained on written words. 
For this reason, it is crucial to computationally investigate the \ac{ME} bias in a more naturalistic setting resembling the environment in which children learn words and employ the \ac{ME} bias. %
We ask whether a \ac{VGS} model trained on spoken words and natural images of entire scenes exhibits the \ac{ME} bias when attempting to make sense of novel words it has never encountered.

\vspace{2ex}
\noindent\begin{researchquestions}
	
	{\textbf \researchqfour}
	
\end{researchquestions}

The previous research question considered the \ac{ME} bias in a \ac{VGS} model to computationally study it in a setting representative of the cognitive \ac{ME} studies done on children.
There has also been research into how the bias in children is influenced by language experience such as multilingualism. 
\citet{kalashnikova_effects_2015}, \citet{davidson_monolingual_1997}, \citet{davidson_monolingual_2005} and \citet{houston-price_language_2010} found that monolinguals exhibited a stronger \ac{ME} bias than bilinguals.
\citet{byers-heinlein_monolingual_2009} found that monolingual children showed a strong \ac{ME} bias while bilinguals showed minor usage of the \ac{ME} bias, and trilinguals showed no \ac{ME} bias.

Although the \ac{ME} bias has been computationally studied in the monolingual setting, the influence of multilingualism on the bias has yet to be studied using a machine learning model since some of the computational \ac{ME} studies found their models to have a \ac{ME} bias, but the bias was not robust.
However, in the previous research question, we establish that an English \ac{VGS} model exhibits a robust and consistent \ac{ME} bias.
As a result, we study the influence of training the same model in various multilingual settings to see if multilingual \ac{VGS} models replicate the \ac{ME} trends seen in children.

\mysection{Approach}{Approach}

After laying out our research questions in the previous section, we now set out the approach to answer each research question.
We use a new general \ac{VGS} model throughout the dissertation and all the research questions.
This general model we propose consists of a vision branch and an audio branch resembling the general architecture of \ac{VGS} models described in Section~\ref{subsec:general_structure}.
For each variation of this model we consider in this dissertation, the vision branch consists of a state-of-the-art unimodal image model to take advantage of advances made in the field of computer vision (Section~\ref{subsec:unimodal_advantage}), followed by an encoder to either get a single image embedding or a sequence of per pixel embeddings.
For the audio branch, we take advantage of advances made in the speech field (Section~\ref{subsec:unimodal_advantage}) by using a self-supervised \acf{CPC} network followed by an encoder to get a single audio embedding or a sequence of per frame embeddings.
We connect the vision and audio branches with a multimodal attention mechanism that differs based on the application we consider.

In summary, to suit our requirements for each application we consider in this dissertation, we adapt our general \ac{VGS} model by modifying either (1) the vision or audio branches to get a single embedding or a sequence of embeddings for an input or (2) the multimodal attention mechanism. 

\mysubsection{\Acf{VPKL}}{\Aca{VPKL} (\ac{VPKL})}
\label{approach:vpkl}

In the task of \ac{VPKL} we consider in \rqone, an image depicting a query keyword should be detected within a spoken utterance.
If the image query is detected in the spoken utterance, it should be predicted where in the segment it occurs. 
To adapt our general model to the \ac{VPKL} application, we adapt both the audio and vision branches to obtain a sequence of embeddings for the speech and image inputs. 
The multimodal attention mechanism takes the dot product between each pixel embedding and each frame embedding. 
This attention mechanism is called a matchmap~\citep{harwath_jointly_2018}.
From the matchmap scores, we can calculate attention weights across the image pixels and across the audio frames, which in turn can be used to get an image context vector and an audio context vector, similar to \cite{olaleye_keyword_2022}.
We obtain an overall similarity score for a speech utterance and image by calculating the cosine distance between the speech and image context vectors.

For training, we use a visual tagger to tag an image, and therefore also its paired spoken utterance, with possible keywords that occur in the pair and which we want to localise.
Although this method shows promise when using an ideal tagger, its performance dropped notably when applied in an unsupervised manner using an actual visual tagger resembling a low-resource setting. 
Therefore, we move our attention to few-shot learning.
We incorporate advances we make in this few-shot setting to improve our \ac{VPKL} approach.

\mysubsection{Visually grounded few-shot spoken word acquisition for low-resource languages}{Visually Grounded Few-Shot Spoken Word Acquisition for Low-Resource Languages}
\label{approach:few-shot}

We combine two core ideas for our visually grounded few-shot word acquisition approach we consider for \rqtwo.
Firstly, we use a set containing a few spoken words paired with images for each new word class we want to learn. 
From this set, we “mine” new noisy word-image pairs from unlabelled speech and image collections. 
Specifically, each spoken word instance in the set is compared to each speech segment in a large unlabelled speech corpus. 
We use a new query-by-example approach to identify segments in the search utterances that match the word in the set.
For mining additional images, each image in the set is compared to each image in a large unlabelled image corpus using cosine distance between pretrained image embeddings.
The mined words and images matched to a class' word and image instances in the given set are then paired to get training word-image pairs.

The second part of our approach relies on adapting the multimodal attention mechanism of our general \ac{VGS} model to a word-to-image attention mechanism.
This model, called \model, encodes a spoken word to a single embedding and an image to a sequence of pixel embeddings.
The word-to-image attention mechanism calculates the similarity between the word embedding and each pixel embedding to learn how and where the word is depicted within an image.
We calculate the overall similarity score between a spoken word and an image by taking the maximum attention scores.

For the scenario where the number of examples per class in the support set is small, our new model outperforms an existing model on few-shot retrieval, where an English query is given and all the natural images it occurs in, should be retrieved.
Additionally, with varying amounts of examples per class, our model achieves scores higher than 80\% on a few-shot word classification task. 
In this task, a spoken English word should be matched to the natural image depicting the word class from a set containing a natural image for each few-shot class we aim to learn. 
Since we are the first to consider this task on natural images, we set a competitive few-shot classification baseline.
Due to these results, we also perform -- for the first time -- visually grounded few-shot word acquisition on an actual low-resource language, \yoruba.
We find that a visually grounded few-shot \yoruba\ model benefits substantially from taking advantage of a more substantial amount of English speech-image data: a \yoruba\ model that leverages English data outperforms an English model variant.
After we establish that we can find a \ac{VGS} model capable of learning the image and spoken word representations in a low-resource language for a class from only a few ground truth examples (\rqtwo), we go back to the \ac{VPKL} task in \rqone.

\mysubsection{\ac{VPKL} for low-resource languages}{\Ac{VPKL} for Low-Resource Languages}

In the task of \ac{VPKL} an image query depicting a keyword should be detected and then localised within a spoken utterance.  
After the breakthroughs we made in the few-shot study discussed in the previous section, we adapt the attention mechanism of our previous \ac{VPKL} model and use a few-shot pair sampling scheme to get training pairs.
As discussed in Section~\ref{approach:vpkl}, we previously used an attention mechanism in which the dot product between each embedding in a sequence of frame embeddings and each embedding in a sequence of pixel embeddings are calculated. 
After that, we obtained context vectors, which we used in a contrastive loss.
Now, we discard the context vectors and simply calculate the similarity score between an image and utterance directly from the matchmap attention scores.
I.e. the audio branch encodes a speech utterance into a sequence of frame embeddings, and the vision branch encodes an image into a sequence of pixel embeddings. 
Next, we calculate the dot product between each frame embedding and each pixel embedding to get a two-dimensional matrix of attention scores -- the matchmap. 
The overall similarity score for the image and spoken utterance is the maximum attention score in the matrix. 

For the few-shot pair mining scheme we employ here, we use the same query-by-example method as in Section~\ref{approach:few-shot}, but now we assume that we have access to images paired with spoken utterances. 
We collect a speech set containing a few word examples for each class we want to learn. 
Then, using these word examples for each class, we find utterances that are predicted to contain the spoken word for the class using the query-by-example method.
With this, we can automatically tag an image with the keywords predicted to be in the image's paired utterance.

On an English \ac{VPKL} task, our new few-shot model outperforms our previous unsupervised model (Section~\ref{approach:vpkl}) and the model of \cite{olaleye_keyword_2022} that localises textual keywords instead of image queries.
To do \ac{VPKL} on an actual low-resource language, we use the \ac{YFACC} of \citet{olaleye_yfacc_2023}.
Although we find that the few-shot \yoruba\ model performs moderately in comparison to the English one, we also find that a ground truth \yoruba\ model outperforms the \yoruba\ textual keyword localisation \ac{VGS} model of \citet{olaleye_yfacc_2023} by large margins.
As a result, we answer \rqone\ by establishing that we can get a \ac{VGS} model capable of detecting and localising a keyword depicted by an image within speech from a low-resource language if we can improve the training pairs further.

\mysubsection{Investigating the \acf{ME} bias in an English \ac{VGS} model}{Investigating the \aca{ME} (\ac{ME}) Bias in an English \ac{VGS} Model}

The \ac{ME} bias is a constraint used by children stating that they would rather associate a novel word with an unfamiliar object than with a familiar object.
To computationally study the ME bias, our few-shot word learning model, \model, is ideal for testing whether a \ac{VGS} model exhibits the \ac{ME} bias.
\Ie answering \rqthree.
Instead of using a pair mining method, we use ground truth labels to find training pairs for a set of familiar classes.
These training pairs consist of spoken words paired with natural images.

After training, we evaluate a model's \ac{ME} bias using the test set consisting of spoken English words paired with images of isolated objects for the familiar classes and a set of novel classes.
Specifically, we prompt a model with a novel spoken word and ask whether the word corresponds to an image of a familiar or novel object.
We consider different model initialisations and find that all the models we consider exhibit a \ac{ME} bias.
The model using both audio and vision initialisations has the strongest \ac{ME} bias.
We also use various losses to investigate how specific the occurrence of the \ac{ME} bias is to our model design and find that the InfoNCE loss results in the strongest \ac{ME} bias. 	

\mysubsection{Investigating the effect of multilingualism on the \acf{ME} bias in \ac{VGS} models}{Investigating the Effect of Multilingualism on the \aca{ME} (\ac{ME}) Bias in \ac{VGS} Models}

Various cognitive studies found that monolingual children have a stronger \ac{ME} bias than bilingual children.
\citet{byers-heinlein_monolingual_2009} found that monolingual children show a stronger use of the \ac{ME} bias than bilingual and trilingual children, and trilingual children use the bias less than bilingual children.
To investigate whether \ac{VGS} models exhibit the same \ac{ME} trends seen in multilingual children, we use the English \model\ model with the InfoNCE loss, which we found had the strongest \ac{ME} bias in the previous research question. 

We apply this \model\ model to two bilingual and one trilingual setting: an English-Dutch model, an English-French model and a trilingual English-Dutch-French model.
To do this, we expand the English \ac{ME} dataset's training set to contain French and Dutch spoken words for the familiar classes.
We use a single audio branch in each model to encode all the words from the languages under consideration.
We find that the \ac{ME} trend between monolingual, bilingual and trilingual models is opposite to that observed in children. 
The monolingual model has the weakest ME bias and the trilingual model has the strongest ME bias, whereas monolingual children have a strong \ac{ME} bias and trilingual children have no \ac{ME} bias.

\mysection{Contributions}{Contributions}
\newcommand{\spacesize}{18pt}

This dissertation makes the following contributions to \ac{VGS} modelling, and specifically to low-resource language applications and cognitive modelling.

\vspace{\spacesize}
\noindent\researchContributionOne

\vspace{\spacesize}
\noindent\researchContributionTwo

\vspace{\spacesize}
\noindent\researchContributionThree

\vspace{\spacesize}
\noindent\researchContributionFour

\vspace{\spacesize}
\noindent\researchContributionFive

\vspace{\spacesize}
\noindent\researchContributionSix

\vspace{\spacesize}
\noindent\researchContributionSeven

\vspace{\spacesize}
\noindent\researchContributionEight

\vspace{\spacesize}
\noindent\researchContributionNine

\mysection{Publications}{Publications}

\newcommand{\spacesizeTwo}{15pt}

We now provide a list of the research papers presented in this dissertation. 
The paper numbers correspond to the chronological order in which the papers were written, submitted and published.
Code resources are given at the start of each chapter.

\vspace{\spacesizeTwo}
\noindent\researchpaperbox{Research Paper \towardsVPKL}{\bibentry{nortje_towards_2023}}

\vspace{\spacesizeTwo}
\noindent\researchpaperbox{Research Paper \VPKL}{\bibentry{nortje_improved_2024}}

\vspace{\spacesizeTwo}
\noindent\researchpaperbox{Research Paper \fewshotInterspeech}{\bibentry{nortje_visually_2023}}

\vspace{\spacesizeTwo}
\noindent\researchpaperbox{Research Paper \fewshot}{\bibentry{nortje_visually_2024}}

\vspace{\spacesizeTwo}
\noindent\researchpaperbox{Research Paper \ME}{\bibentry{nortje_visually_2024-1}}

\vspace{\spacesizeTwo}
\noindent\researchpaperbox{Research Paper \MultilingualME}{\bibentry{nortje_using_2024}}

\mysection{Dissertation outline}{Dissertation Outline}
\renewcommand{\spacesize}{5ex}

In this section, we give an outline of the dissertation as summarised in Figure~\ref{fig:outline}, and a brief description of what can be expected from each chapter.
Each chapter contains one or two publications.
The general structure of each chapter is more or less as follows: an introduction, a related work section, a publication, a dataset discussion, further analysis and a chapter summary.
Chapter~\ref{chap:localisation} contains two publications and no dataset discussion or further analysis section, Chapter~\ref{chap:few-shot} does not include a dedicated dataset discussion and Chapter~\ref{chap:multilingual_me} does not have a dedicated related work section.
We do not have a chapter dedicated to discussing relevant background research, but instead, we have the dedicated related work section in each chapter (except Chapter~\ref{chap:multilingual_me}) to discuss relevant research for the chapter's topic. 

\begin{figure}[tb]
	\centering
	\includegraphics[width=0.9\linewidth]{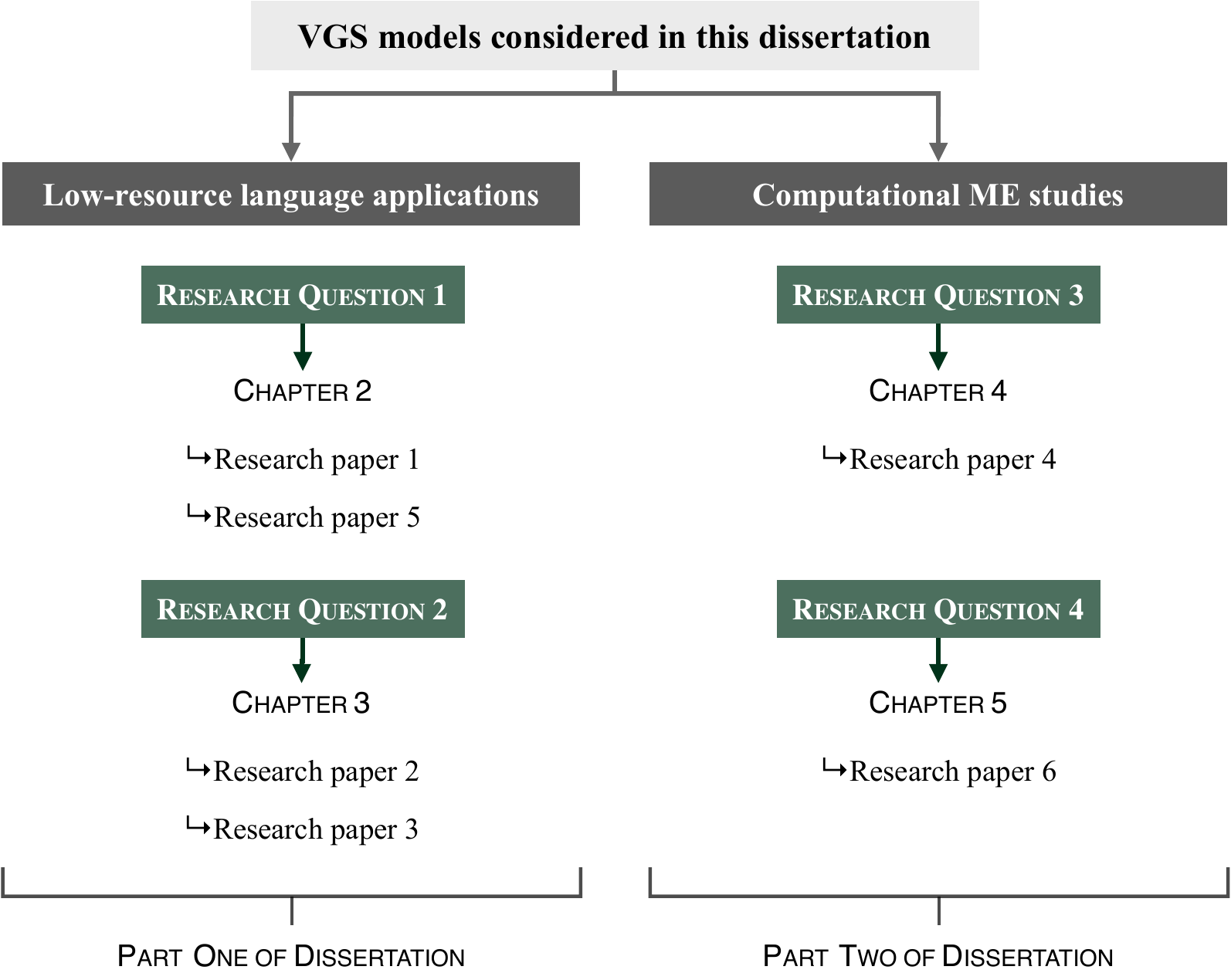}
	\caption{
		This figure contains a detailed dissertation outline. 
		The dissertation consists of two parts: low-resource applications and computational cognitive \ac{ME} studies. 
		This figure shows in which categories the research questions are placed.
		It also shows in which chapters and papers we attempt to answer each research question. 
	}
	\label{fig:outline_detailed}
\end{figure} 

\vspace{\spacesize}
\noindent\ShortInTextTitle{\large Chapter~\ref{chap:localisation}: \nameref{chap:localisation}}\newline
In this chapter, we aim to answer \researchqone\ 
We discuss keyword localisation models and how previous studies approached this using \ac{VGS} models.
We propose a new task called \acf{VPKL} and a new \ac{VGS} model to do this task in the first publication (\RQtowardsVPKL).
However, after applying our approach to an unsupervised setting that closely resembles a low-resource language setting, we find that our approach performed worse than an existing unsupervised \ac{VGS} \ac{BOW} model~\citep{olaleye_attention-based_2021} that locates textual keywords.
As a result, we move our attention to the few-shot study in the next chapter and use the findings to propose a new \ac{VGS} model to do \ac{VPKL} in the second paper of this chapter (\RQVPKL). 
After this approach outperformed the existing \ac{VGS} \ac{BOW} model on English, we apply it to an actual low-resource language, \yoruba.
The performance of the low-resource model using the few-shot mined pairs drops significantly from the ground truth model.
However, the performance of the ground truth model shows that an improvement in the mined pairs could lead to a useful low-resource keyword localisation model.

\vspace{\spacesize}
\noindent\ShortInTextTitle{\large Chapter~\ref{chap:few-shot}: \nameref{chap:few-shot}}\newline
Chapter~\ref{chap:few-shot} considers \researchqtwo\
This chapter discusses few-shot learning in a natural setting consisting of spoken words and natural images of scenes.
We propose a new \ac{VGS} model that outperforms an existing state-of-the-art model when using fewer examples per class.
This work is presented in \RQfewshotInterspeech, but we do not present this research paper in this chapter since we extend it to a journal article, \RQfewshot, which we present in this chapter.
In this extension, we also apply this approach to an actual low-resource language, \yoruba, using the \ac{YFACC} dataset collected by \citet{olaleye_visually_2023}.
We found that taking advantage of a well-resourced language like English results in a low-resource model that is accurate in learning words.

\vspace{\spacesize}
\noindent\ShortInTextTitle{\large Chapter~\ref{chap:me}: \nameref{chap:me}}\newline
In this chapter, we look at \researchqthree\
We discuss the \acf{ME} studies done in both the cognitive and the computational fields. 
This chapter addresses the research gap in which a natural setting consisting of spoken words and natural images has never been used to computationally investigate the \ac{ME} bias.
In \RQME, we establish that the model architecture used in the few-shot chapter, \model, exhibits the \ac{ME} bias.
More specifically, we first train the model on ground truth spoken words and images for familiar classes.
After that, we use samples for novel classes to test whether this model exhibits the \ac{ME} bias by showing it a novel and a familiar image and asking the model which image belongs to a novel spoken word.
We found that the model using unimodal vision and audio initialisations that closely resemble children's knowledge when they start using the \ac{ME} bias, strengthens the \ac{ME} bias.
Finally, we find that the InfoNCE loss results in the strongest \ac{ME} bias.

\vspace{\spacesize}
\noindent\ShortInTextTitle{\large Chapter~\ref{chap:multilingual_me}: \nameref{chap:multilingual_me}} \newline
The effect of multilingualism on the \ac{ME} bias has been investigated in cognitive studies.
However, it has never been investigated computationally.
As a result, we ask \researchqfour\
After showing in the previous chapter that our \model\ InfoNCE model exhibits the \ac{ME} bias, we apply this model to various multilingual settings in \RQMultilingualME.
With this, we aim to test whether the multilingual variants show the same \ac{ME} trends as observed in multilingual children.
The \model\ model shows the opposite monolingual vs multilingual \ac{ME} trends than children since the monolingual model shows a weaker \ac{ME} bias than the multilingual models.

\vspace{\spacesize}
\noindent\ShortInTextTitle{\large Chapter~\ref{chap:conclusion}: \nameref{chap:conclusion}}\newline
This chapter summarises the main conclusions from each chapter.
We also highlight the conclusions that answer our research questions (Section~\ref{research_questions}).
Lastly, we discuss possible avenues for future work and the limitations of our work.

\acresetall
\divider{Low-Resource Studies}
\mychapter{Visually prompted keyword localisation}{Visually Prompted Keyword Localisation}{Research Question 1}
\label{chap:localisation}

\noindent\begin{publicationinfo}{}
	\begin{center}
		{\large\ShortInTextTitle{Chapter~\ref{chap:localisation} Specifics}}
	\end{center}
	
	\vspace{4pt}
	
\researchqonecentered

\vspace{4pt}
	
\researchpaper{Research Paper \towardsVPKL}{\bibentry{nortje_towards_2023}.}

\vspace{4pt}

\researchpaper{Research Paper \VPKL}{\bibentry{nortje_improved_2024}.}

\vspace{4pt}

\ShortInTextTitle{Project Website with Code Resources}: \href{https://sites.google.com/view/vgskeywordlocalision?usp=sharing}{[Link]}
\end{publicationinfo}

Low-resource languages do not have the data resources required to build speech systems. 
One solution is systems that can learn fast and efficiently from unlabelled or little transcribed data.
Recent research uses vision as a weak form of supervision to overcome this scarcity of data hurdle~\citep{chrupala_visually_2022}. 
These \acf{VGS} are trained on images paired with unlabelled spoken captions -- removing the need for text.
Specifically for this chapter, \citet{olaleye_yfacc_2023} have considered keyword localisation for low-resource languages using a \ac{VGS} model developed by \citet{olaleye_attention-based_2021} and \citet{olaleye_keyword_2022}.

\Gls{keyword_localisation}  was developed from \gls{keyword_spotting}, which ranks the spoken utterances in a set by how likely it is that a query keyword occurs in each utterance~\citep{wilpon_automatic_1990}. 
\Gls{keyword_detection} is the task of classifying whether a given query keyword occurs in a single spoken utterance or not.
Keyword localisation is the task of locating where in a spoken utterance a given query keyword occurs after the keyword is detected in the utterance~\citep{olaleye_attention-based_2021}. 
\Ie keyword localisation is a two-step approach: detection and localisation. 
More specifically, given a textual keyword and a spoken utterance, the model should first detect whether the keyword occurs in the utterance. 
If the keyword indeed occurs in the utterance, the model should detect where it occurs in the utterance.

In this chapter and more specifically in \RQtowardsVPKL, we introduce a new visual keyword localisation task called \gls{vpkl}\acuse{VPKL}: a model is given an image depicting a query keyword, a so-called image query.
The model should detect if the image query occurs in a spoken utterance, and if the query occurs in the utterance, the model should localise it in the utterance.
\Eg the model is shown an image of a \image{zebra} and asked whether it occurs in the spoken caption: \word{a zebra grazing in the field}. 
The model should predict that the query occurs in the utterance, whereafter it should also predict where \word{zebra} occurs in the utterance. 

This new \ac{VPKL} task relates to \rqone.
As a reminder, this question asks: \researchquestionone\
To do this, we need a \ac{VGS} model capable of comparing images or regions within images to spoken utterances in a low-resource language.
To develop such a model, we use English as an artificial low-resource language in \RQtowardsVPKL.
We find that our approach in \RQtowardsVPKL\ is not suitable to do \ac{VPKL} in a low-resource setting, and therefore, we propose a new approach in \RQVPKL.
We find that this latest approach works better than any other approach in a setting that resembles a low-resource language. 
As a result, we apply this approach to an actual low-resource language, \yoruba, in \RQVPKL.

\mysection{Related work}{Related Work}

Keyword spotting is a widely studied problem that entails detecting whether a keyword occurs within a set of spoken utterances~\citep{wilpon_automatic_1990}.
An example use case is the system developed by \citet{chelba_retrieval_2008}, where a user enters textual terms into a search engine and the system returns spoken documents containing the search terms. 
\citet{szoke_comparison_2005} and \citet{lee_improved_2012} used a spoken word as the query keyword.
\citet{lee_improved_2012} focussed on matching the acoustics of keywords and the frequency of co-occurring words to do semantic retrieval of spoken utterances.
On the other hand,  \citet{li_towards_2013} considered an unsupervised take on this semantic keyword spotting task.
\citet{lee_spoken_2015} give a thorough overview of traditional speech-based keyword spotting models.

A considerable amount of speech resources are required in the traditional keyword spotting approaches, where the above-mentioned are just a few examples.
As a result, \citet{garcia_keyword_2006} developed a keyword spotting system that requires only about fifteen minutes of transcriptions for training speech data. 
Moreover,  \citet{van_der_westhuizen_feature_2022} considered keyword spotting systems that can work in low-resourced settings with little to no resources available. 
They consider both the cases where there is no transcribed data or a small set of transcribed data.
We, however, consider systems that not only detect keywords but also localise these keywords in speech from a low-resource language.
Additionally, we incorporate images in our systems similarly to \citet{drexler_analysis_2017}, who used images to transcribe speech to train a keyword spotting model in an unsupervised manner. 

\citet{kamper_visually_2017} trained a \ac{VGS} keyword spotting model on images paired with spoken descriptions.
More specifically, they use a multi-label visual classifier to tag the images paired with spoken utterances with textual word labels from a fixed set of keywords.
A visual \ac{BOW} model is trained to map speech to these generated targets.
We also use these visual tagger generated targets in the first paper (\RQtowardsVPKL) we present in this chapter.
After \citet{kamper_visually_2017} found that their system can detect if any of the keywords in the fixed set occur in a spoken utterance, \citet{kamper_visually_2018} extended the model to do cross-lingual keyword spotting.
In this cross-lingual spotting task, a textual German keyword is given, and spoken utterances containing the English translation of the keyword should be retrieved. 
Such systems could enable text queries in a high-resource language to search through and retrieve speech in a low-resource language.
Our low-resource \ac{VGS} model proposed in this chapter detects and localises language independent image queries in \yoruba\ speech.
Therefore, a speaker with any native language can retrieve \yoruba\ utterances using such a system.

Up until here, the studies we discussed consider keyword spotting which is different to the keyword detection that \citet{olaleye_towards_2020} consider. 
Keyword spotting entails the ranking of spoken utterances according to how likely it is that a query keyword occurs in each utterance~\citep{olaleye_visually_2023}. 
However, keyword detection entails detecting whether a query occurs in a single spoken utterance. 
\citet{olaleye_towards_2020} are the first to consider a \ac{VGS} model that not only detects whether a keyword occurs within a spoken utterance but also where the keyword occurs. 
They do not explicitly provide location information during the training of their \ac{VGS} \ac{BOW} keyword localisation model. 
\Ie the model is trained without alignment supervision. 
They specifically consider using the visual tagger targets of \citet{kamper_visually_2017} but found that alternative modelling methods that can better take advantage of the complementary visual signal are necessary.

Due to the findings of \citet{olaleye_towards_2020}, \citet{olaleye_attention-based_2021} used the same generated visual tagger targets to train an attention-based \ac{VGS} model.
The hope is that the attention mechanism will indirectly learn to localise the keyword under consideration.
Although the localisation performance of their \ac{VGS} model is still relatively low compared to supervised models, they found that the attention model contributes a large performance gain over previous \ac{VGS} approaches.
After this, \citet{olaleye_keyword_2022} added an input masking approach to the attention-based \ac{VGS} model. 
They found that this model outperformed the localisation \ac{VGS} models of \citet{olaleye_towards_2020} and \citet{olaleye_attention-based_2021}.
We also use an attention-based \ac{VGS} model to do keyword localisation. 
Unlike the \ac{VGS} keyword localisation studies mentioned above, we incorporate the supplementary visual signal directly in our model so that we can use images instead of textual keyword queries.
Until now, all the \ac{VGS} keyword localisation studies and our approach in \RQtowardsVPKL\ use artificial low-resource languages.
In the second research paper (\RQVPKL) we present in this chapter, we apply our approach to an actual low-resource language. 

\citet{olaleye_yfacc_2023} is the only other \ac{VGS} study considering keyword localisation on an actual low-resource language.
To train a \yoruba\ attention-based \ac{VGS} \ac{BOW} model to do keyword localisation similarly to the English model of \citet{olaleye_keyword_2022}, they collect \yoruba\ audio captions for the Flickr 8k image dataset.
They use the visual tagger targets of \citet{kamper_visually_2017} to tag the images paired with \yoruba\ spoken captions with English textual labels. 
With this model, they do cross-lingual keyword localisation similar to \citet{kamper_visually_2018}.
However, unlike \citet{kamper_visually_2018}, they prompt the model to detect and localise a written English query in \yoruba\ spoken utterances.
We use the same \yoruba\ dataset to train a \ac{VGS} keyword localisation model in \RQVPKL, where a language independent image query can retrieve spoken utterances.
\Ie we use a \yoruba\ \ac{VGS} model to do \acf{VPKL}.
The specifics of this \ac{VPKL} task are given in  \RQtowardsVPKL\ (Section~\ref{localisation_publication}), which we present next.

\mysection{Publication: Research paper \towardsVPKL}{Publication: Research Paper \towardsVPKL}
\label{localisation_publication}

The publication we present in this section (\RQtowardsVPKL), delves into our approach to using a \ac{VGS} model to do \ac{VPKL}: detecting and localising a keyword query depicted by an image in a spoken utterance.
Before we get to the paper, we give a layout of each author's contribution to the paper.

\mysubsection{Contribution declaration}{Contribution Declaration}

\begin{table}[h]
	\caption{
		A layout of the authors’ contributions to the publication: \bibentry{nortje_towards_2023}.
	}
	\label{tab:localisation_contributions}
	\centering
	\renewcommand{\arraystretch}{1.2}
	\begin{tabularx}{\linewidth}{@{}lL@{}}
			\toprule
			Author &  Contributions \\
			\midrule
			\leanneName &  The layout of the research question, the model architecture, the implementation and the generation of numerical results and graphs.\\
			\addlinespace
			\hermanName &   Layout of research question and editorial role.\\ 
			\bottomrule
	\end{tabularx}
\end{table} 

All work in the following publication was proposed and implemented by the first author, \leanneNameNoSpace, except if stated otherwise in Table~\ref{tab:localisation_contributions}.

\mysubsection{Paper}{Paper}

\begin{figure}[!t]
	\centering
	\includegraphics[width=\linewidth]{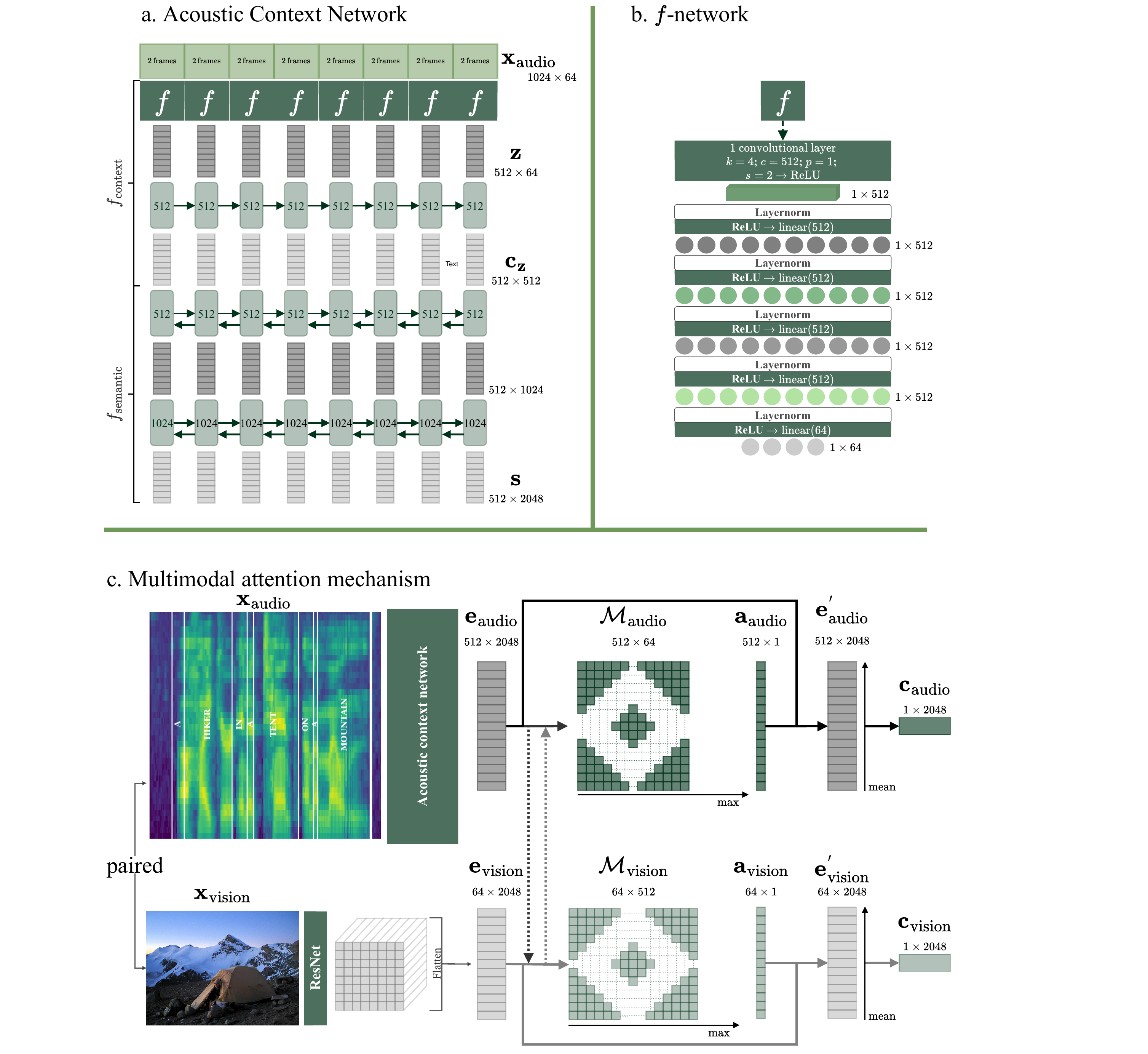}
	\caption{
		(c) \locAtt\ presented in \RQtowardsVPKL, consists of a vision network and an audio (a+b) network. 
		The two branches are connected through a multimodal attention mechanism consisting of a matchmap~\citep{harwath_jointly_2018}. 
		The model outputs a similarity score $S$ for a speech and an image input based on the context vectors obtained from the matchmap.%
	}
	\label{fig:loc_model}
\end{figure} 

We now present \RQtowardsVPKL.
For more information, we give  Figure~\ref{fig:loc_model}, which shows the entire model architecture of the model we call \locAtt\ used in this paper. 
Figure~\ref{fig:loc_model}~(a) and Figure~\ref{fig:loc_model}~(b) show the entire architecture of the audio branch we use. 

\newpage
\includepdf[pages={1-},scale=1.0]{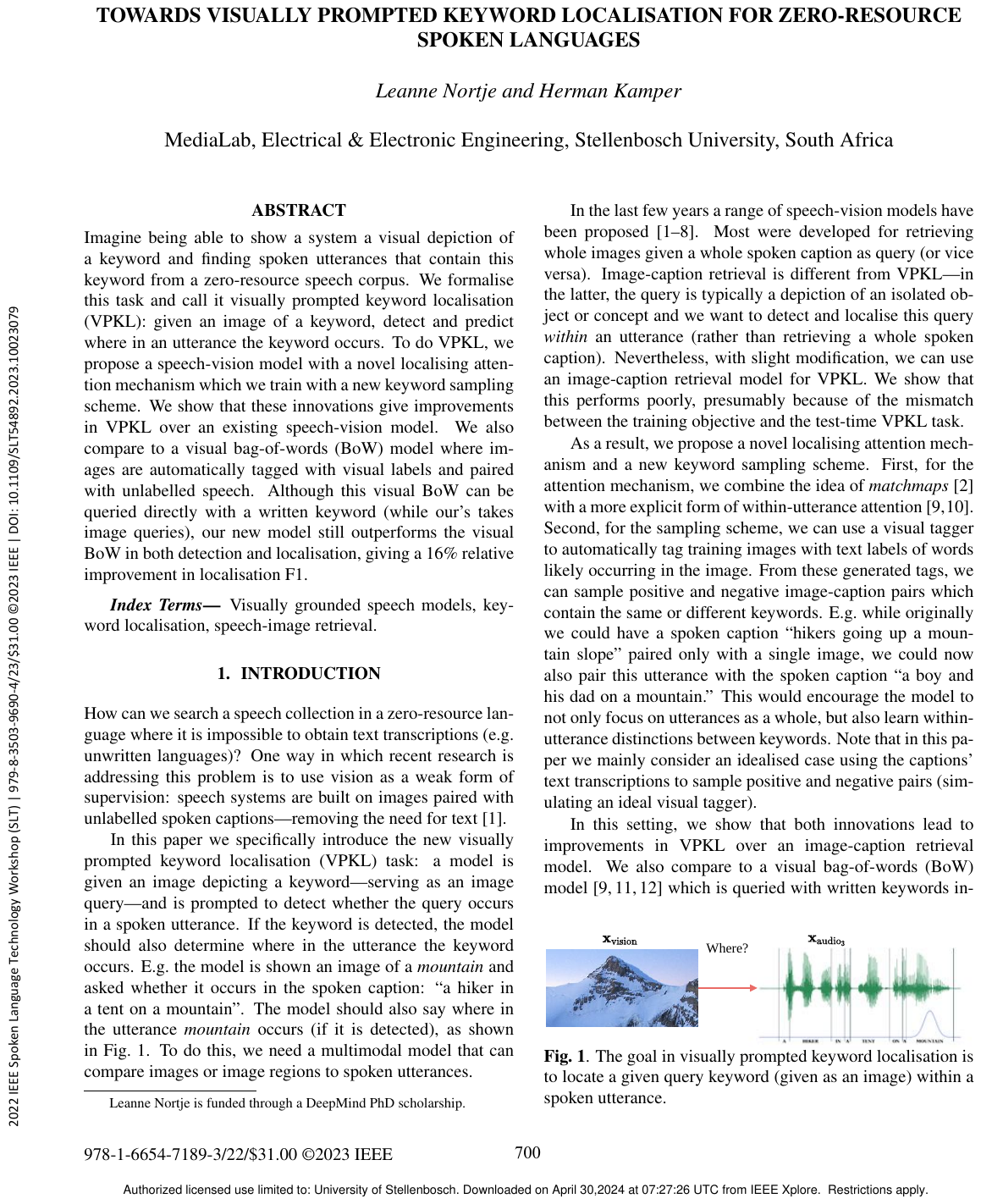} 

\mysection{Link between publications}{Link between Publications}

In the previous publication, we proposed \ac{VPKL}: detecting and localising a keyword query depicted by an image in a spoken utterance.
\ac{VPKL} systems could be valuable to provide solutions for humanitarian relief in regions where low-resource languages are used: a language independent keyword query can be used to retrieve speech in a low-resource language that conveys emergency or disaster situations.
However, the previous publication showed that the proposed system must be improved before applying it to an actual low-resource language.
More specifically, we found that the proposed model outperforms the \ac{VGS} \ac{BOW} of \citet{olaleye_keyword_2022} when using ground truth transcriptions.
But, when we apply the model in an unsupervised manner to resemble a low-resource setting, the model performs poorly.

The next publication attempts to overcome the challenge of getting a \ac{VGS} model capable of doing \ac{VPKL} in a low-resource language. 
First, we develop a new approach on an artificial low-resource language, English. 
After we find that this new approach outperforms our previous \ac{VPKL} approach and a visual \ac{BOW} approach in a setting that resembles a low-resource language, we apply our new approach to an actual low-resource language, \yoruba.

\mysection{Publication: Research paper \VPKL}{Publication: Research Paper \VPKL}
\label{few-shot_localisation_publication}

In the following publication referred to as \RQVPKL, we develop a new \ac{VGS} model to do \ac{VPKL} on an actual low-resource language, \yoruba.
Section~\ref{sec:vpkl_contributions} outlines each author’s contribution to the paper, and Section~\ref{vpkl_paper} presents the paper. 

\mysubsection{Contribution declaration}{Contribution Declaration}
\label{sec:vpkl_contributions}

\begin{table}[h]
	\caption{A layout of the authors’ contributions to the publication: \bibentry{nortje_improved_2024}.}
	\label{tab:few-shot_localisation_contributions}
	\centering
	\renewcommand{\arraystretch}{1.2}
	\begin{tabularx}{\linewidth}{@{}lL@{}}
			\toprule
			Author &  Contributions \\
			\midrule
			\leanneName & The layout of the research question, the model architecture, the pair mining algorithm and the implementation on English and \yoruba. The generation of all the numerical results.\\
			\addlinespace
			\danName &  The visualisation analysis for the localisation of image queries within spoken utterances, and an editorial role. \\
			\addlinespace
			\hermanName & Layout of research question and editorial role.\\ 
			\bottomrule
		\end{tabularx}
	\end{table} 

All work in the following publication was proposed and implemented by the first author, \leanneNameNoSpace, except as stated otherwise in Table~\ref{tab:few-shot_localisation_contributions}.

\mysubsection{Paper}{Paper}
\label{vpkl_paper}

\begin{figure}[tb]
	\centering
	\includegraphics[width=0.7\linewidth]{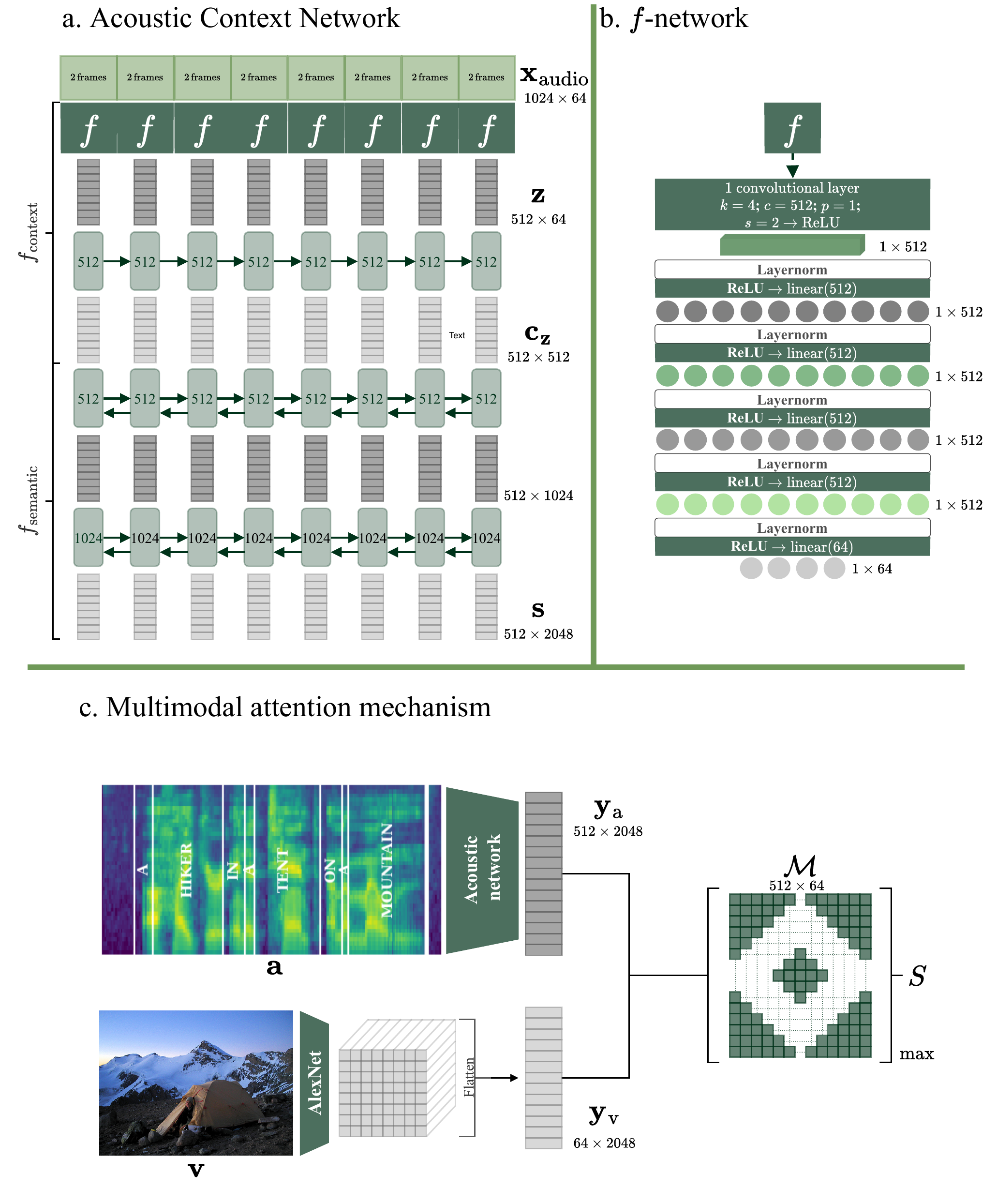}
	\caption{
		(c) \locAttShort\ consists of a vision and an audio (a+b) network. 
		The two branches are connected through a multimodal attention mechanism consisting of a matchmap $\mathcal{M}$~\citep{harwath_jointly_2018} used to calculate a similarity score $S$ for a speech and an image input.%
	}
	\label{fig:simple_loc_model}
\end{figure} 
	
We now present \RQVPKL.
It is important to note that we change the \ac{VGS} \ac{VPKL} model name from \locAtt\ in the previous publication to \locAttShort\ in the following publication. 
This name change is due to the space limitation in the next publication and to compare the model proposed in the previous publication more clearly to the model in the following publication.

We give the entire architecture of the model we propose in Figure~\ref{fig:simple_loc_model} since it provides more information on the architecture of the audio branch.

\newpage
\includepdf[pages={1-},scale=1.0]{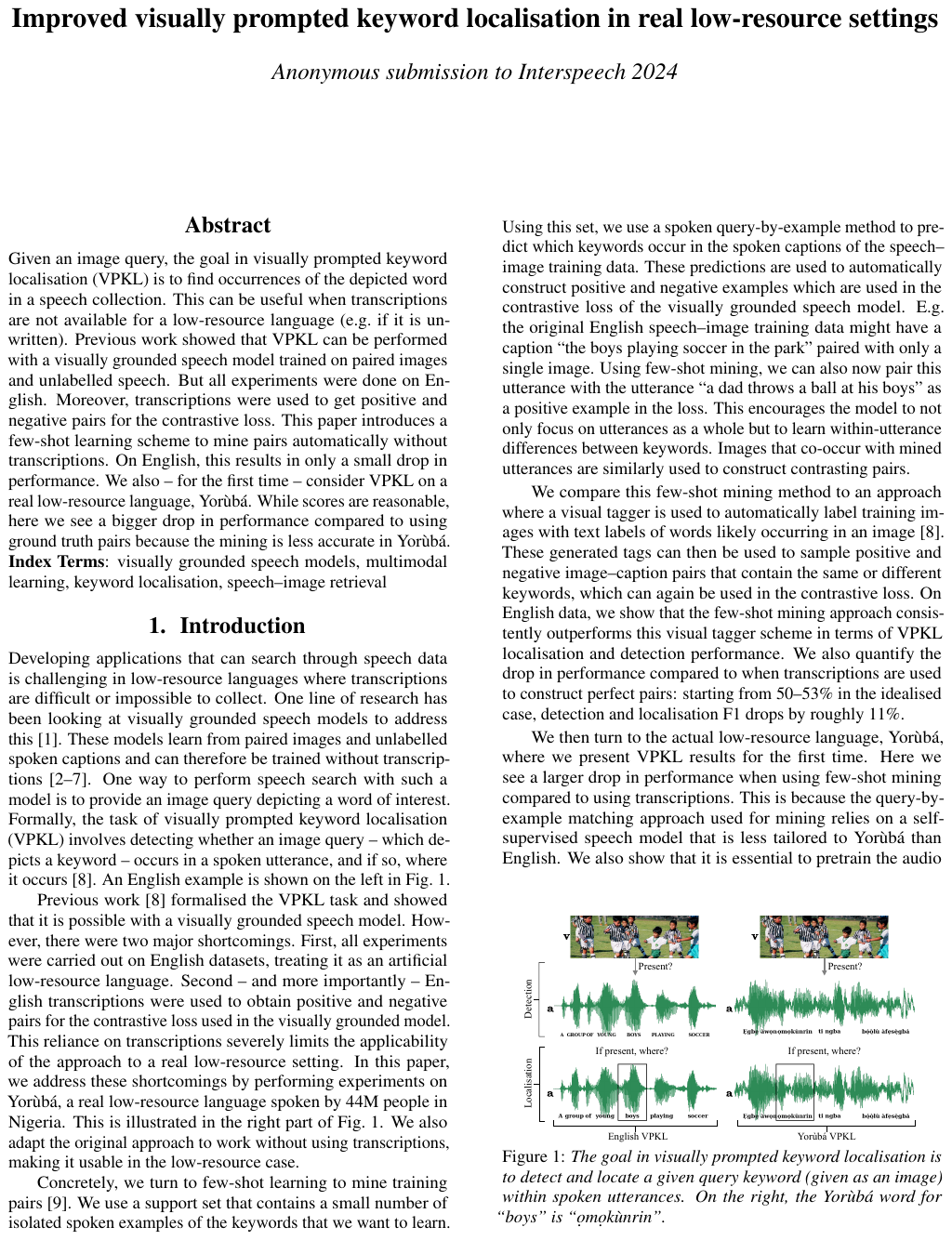} 

\mysection{Chapter summary}{Chapter Summary}

In this chapter, we proposed a new task called \acf{VPKL}, in which a keyword depicted by an image should be detected and localised in spoken utterances.
We proposed a \ac{VGS} model with a new localising attention mechanism to perform this task.
We also used a data sampling scheme in which we mine positive and negative speech-image pairs by using the images' visual tagger keyword targets.
Using an ideal tagger (ground truth transcriptions) to mine pairs, gave improvements over a previous \ac{VGS} model on keyword localisation in English search utterances.
This is notable since we used an image for a keyword query, whereas the previous model used textual queries.
However, when using an actual visual tagger to generate possible keyword labels for a speech-image pair, we found the model’s performance dropped below the existing unsupervised visual \ac{BOW} model that localises textual queries.

Since our ultimate goal was to do \ac{VPKL} in a low-resource setting, we first had to improve the approach on the English \ac{VPKL} task.
As a result, we proposed a second \ac{VGS} model with a simpler multimodal attention mechanism and a few-shot approach to mine positive and negative training pairs automatically.
This pair mining method uses a small set of isolated spoken word examples for the keywords we want to localise.
On English, we show that this second model with the simpler attention mechanism and unsupervised mined pairs, outperforms the first model with the complex attention mechanism and visual tagger pairs and the \ac{VGS} \ac{BOW} model that localises textual keyword queries.
Finally, we applied our second model to \yoruba.
Although the model trained on unsupervised \yoruba\ mined pairs had low precision scores, the ground truth model showed great promise since it outperformed the \yoruba\ \ac{VGS} \ac{BOW} by large margins.

Our proposed models could ideally also compare speech to speech. 
Therefore, spoken keywords from the same language as the search utterances can be used as queries rather than images.
Using this task on our models could be considered in future work.
Another possible avenue for future work is extending the \ac{VPKL} task to do semantic retrieval of spoken utterances. 
Future work should also look into removing an important limitation of our particular approach: we rely on a few-shot support set that contains spoken word examples of the keywords we consider to make our approach applicable to low-resource settings, but it constrains the vocabulary.
The next chapter considers learning a small vocabulary using few-shot speech-image learning and shows how the few-shot techniques we use in \RQVPKL\ (above) were actually developed.

\mychapter{Visually prompted few-shot word acquisition}{Visually Prompted Few-Shot Word Acquisition}{Research Question 2}
\label{chap:few-shot}

\noindent\begin{publicationinfo}{}
	\begin{center}
		{\large\ShortInTextTitle{Chapter~\ref{chap:few-shot} Specifics}}
	\end{center}
	
	\vspace{4pt}
	
	\researchqtwocentered
	
	\vspace{4pt}
	
	\researchpaper{Research Paper \fewshotInterspeech}{\bibentry{nortje_visually_2023}.}
	
	\vspace{4pt}
	
	\researchpaper{Research Paper \fewshot}{\bibentry{nortje_visually_2024}.}
	
	\vspace{4pt}
	
	\ShortInTextTitle{Project Website with Code Resources}: \href{https://sites.google.com/view/few-shotwordacquisition/home}{[Link]}
\end{publicationinfo}

Chapter~\ref{chap:localisation} considers using few-shot learning to perform \acf{VPKL} in a low-resource setting. 
We used few-shot learning because of our findings in this chapter. 
However, Chapter~\ref{chap:localisation} uses a unimodal set consisting of spoken word examples.
In contrast, this chapter uses a multimodal set consisting of spoken words paired with natural images to perform few-shot learning.

\cite{eloff_multimodal_2019} were the first to propose multimodal few-shot learning.
This technique replicates the speed and efficiency with which children learn new words from only a few spoken word and object examples~\citep{biederman_recognition-by-components:_1987, miller_how_1987,gomez_infant_2000,rasanen_joint_2015}.
\Gls{multimodal_few-shot_learning} is a method in which a new class is learned from a few multimodal example pairs, where each example pair consists of two samples from different modalities but from the same class~\citep{eloff_multimodal_2019}.
Although \citet{eloff_multimodal_2019} specifically use the speech and vision modalities, any two modalities can be used for multimodal few-shot learning.
We use the speech and vision modalities to get a textless word learning system that works on low-resource languages.
\Ie we aim to answer \rqtwo: \researchquestiontwo\
\Eg we are given a set containing spoken word examples \word{apple}, \word{mountain} and \word{zebra}, and each word is paired with an image. 
After seeing this set, the model is prompted with a new spoken word query, say \word{zebra}, and asked which image corresponds to the query in another set of images.
This other set of images contains one new image per few-shot class, \eg an image of a \image{mountain}, a \image{zebra} and an \image{apple}.
The model should choose the \image{zebra} image.

In \RQfewshotInterspeech, we combine two core ideas to do few-shot word learning. The first idea uses a set of examples consisting of a few spoken words paired with images for each word class we want to learn.
This set is used to ``mine'' noisy training word-image pairs from unlabelled speech and image collections.
The second idea is to adapt the multimodal attention mechanism of our general \ac{VGS} model to get a word-to-image attention mechanism.
\RQfewshot\ is a journal article that expands \RQfewshotInterspeech\ which is a conference paper.
\RQfewshot\ improves the models proposed in \RQfewshotInterspeech\ and performs more detailed analysis. 
We find that for the scenario where the number of given examples per class is small, the approach we propose outperforms an existing model on few-shot retrieval of images given spoken English queries.
Additionally, our new approach sets a competitive baseline for few-shot word classification where an English spoken word is matched to a natural image containing the visual depiction of the word.
Consequently, \RQfewshot\ performs, for the first time, visually grounded few-shot word acquisition on an actual low-resource language, \yoruba.

\mysection{Related work}{Related Work}

Children are fast and efficient in learning new classes from only a few word or object examples~\citep{biederman_recognition-by-components:_1987, miller_how_1987,gomez_infant_2000,rasanen_joint_2015}.
Few-shot learning is a method used by machine learning networks that emulate children's learning efficiency~\citep{li_fei-fei_bayesian_2003}.
Concretely, \gls{few-shot_learning} entails learning a new class from only a few labelled examples~\citep{li_fei-fei_bayesian_2003, fei-fei_one-shot_2006, lake_one_2011, lake_one-shot_2014, koch_siamese_2015, vinyals_matching_2016, shyam_attentive_2017, snell_prototypical_2017}.
More specifically, a model is given a fixed amount of examples for a class called a \textit{support set} \supportSet\ from which the class should be learned.
Few-shot learning can be done in any modalities with some examples including gesture recognition~\citep{thomason_recognizing_2017,wu_one_2012,wan_one-shot_2013}, video~\citep{stafylakis_zero-shot_2018}, robotics~\citep{walter_one-shot_2012, finn_one-shot_2017}, speech classification~\citep{parnami_few-shot_2020, lake_one-shot_2014} and recognition of objects in images~\citep{li_fei-fei_bayesian_2003, fei-fei_one-shot_2006, lake_one-shot_2013, lake_human-level_2015,koch_siamese_2015, vinyals_matching_2016, snell_prototypical_2017, salakhutdinov_one-shot_2012, ravi_optimization_2017, santoro_meta-learning_2016, mishra_simple_2018, finn_one-shot_2017, tian_rethinking_2020}.

\citet{eloff_multimodal_2019} was the first to propose multimodal few-shot learning.
This method more accurately resembles how children learn.
To recap, \gls{multimodal_few-shot_learning} is a method in which a new class is learned from a few multimodal example pairs. 
Here, each example pair consists of two samples depicting the same class, but the samples are from different modalities.
Any two modalities can be used for multimodal few-shot learning.

Similarly to \citet{eloff_multimodal_2019}, we use the speech and vision modalities in our multimodal few-shot setup.
More specifically, for their approach, they used transcribed background datasets not containing any of the few-shot classes to train separate supervised  
speech and vision networks.
\Ie they use transfer learning to find unimodal models that can generalise to unseen classes~\citep{chechik_large_2009,donahue_decaf_2014,koch_siamese_2015}.
The \textit{multimodal support set} \supportSet\ is used as a pivot between the speech and vision networks.
To match the spoken query to the image it belongs to, the spoken query is compared to each spoken word in the support set using the unimodal speech network. 
The image paired with the support set word most similar to the query is then compared to each image in the matching set \matchingSet\ using the unimodal vision network.
The image most similar to the paired support set image is chosen to belong to the query word.

The setup of \cite{eloff_multimodal_2019} still requires background data with textual transcriptions, and again, with low-resource languages we do not have such transcribed datasets.
\citet{nortje_unsupervised_2020} investigated unsupervised and transfer learning methods to obtain models that do not use textual transcriptions.
Their findings were that these methods did not perform as well as the supervised transfer learning models of \citet{eloff_multimodal_2019}.
As a result, \citet{nortje_direct_2021} considered an approach to directly train on the few-shot support set and leveraging unlabelled datasets. 
This method not only removes the need for transcriptions but also removes the two-step matching approach where mistakes are compounded.

\citet{nortje_direct_2021} attempt to learn a direct mapping between the spoken word and image representations of the few-shot classes from only a few given examples. 
They can then directly compare the spoken query words to possible matching images, eliminating the two-step word-word and image-image matching approach.
To do this, they use the few-shot support set to mine word and image instances predicted to be of a few-shot class.
A transfer learned speech model compares the words in the support set to each spoken word in a large set of unlabelled spoken words. 
The instances most similar to a few-shot class' word examples are predicted to be from this class.
The same is done for the images using a transfer learned vision model.
They found that training a multimodal triplet model outperforms any existing multimodal few-shot approach. 
However, they found that transcribed background data is required to train the networks used for pair mining.

Some of the work in this chapter builds off the master's thesis~\citep{nortje_direct_2020} of \leanneName\ (author of this dissertation).
Specifically, the work of this thesis published in \citet{nortje_direct_2021} and \citet{nortje_unsupervised_2020} are considered precursors to the multimodal few-shot work presented in this chapter.

All the multimodal few-shot studies up until here considered learning digit classes from images containing only the MNIST written digits~\citep{lecun_yann_gradient-based_1998} and spoken words from the TIDigits speech corpus~\citep{leonard_r._gary_tidigits_1993}.
As a result, \citet{miller_tyler_exploring_2022} consider multimodal few-shot learning using natural images. 
They specifically consider multimodal few-shot learning as a way to update existing models with new classes. %
After updating the model with the new few-shot classes, they test how well the models know the few-shot classes while still remembering previously learned classes.
However, they only considered five classes, and after considering various masking procedures, their models showed low precision scores when less than 100 examples per class were given in the support set. 
This number of examples per class is high and resembles a more supervised setting.

Similarly to \cite{miller_tyler_exploring_2022}, we consider few-shot speech-image learning using natural images.
We use their naturalistic few-shot setup in \RQfewshotInterspeech\ and \RQfewshot.
However, we aim to develop strategies that use fewer paired examples per few-shot class to be able to apply our approach to an actual low-resource language: \yoruba.
To accomplish this, we use a similar mining approach as \citet{nortje_direct_2021}, but we leverage advances in self-supervised speech processing and object classification.
We propose a more complex method for the speech pair mining part since we want to find entire spoken utterances containing the spoken support set words and segment out possible words that match the spoken words from the utterances.
For the vision part of the pair mining algorithm, we can leverage vision systems that are trained on labelled image datasets similarly as \citet{eloff_multimodal_2019}, \citet{nortje_unsupervised_2020} and \citet{nortje_direct_2021} since vision is language independent.
Our approach is presented and analysed in \RQfewshot\ in the following section.

\mysection{Publication: Research paper \fewshot}{Publication: Research Paper \fewshot}

The publication presented in this section (\RQfewshot) is an extension of our conference paper, \RQfewshotInterspeech, given in Appendix~\ref{appen:interspeech}. 
In \RQfewshotInterspeech, we propose our \ac{mattnet} to do multimodal few-shot learning using natural images and spoken words but with fewer word-image examples per class.
Ultimately, the purpose is to find fast and efficient learning algorithms that can be used to develop systems for low-resource languages.
As a result, after \RQfewshot\ further improves and analyses our proposed approach, we apply it to an actual low-resource language, \yoruba.
Before we get to \RQfewshot\ in Section~\ref{few-shot_publication}, we give a layout of each author's contribution to the paper in Section~\ref{few-shot_contributions}.

\mysubsection{Contribution declaration}{Contribution Declaration}
\label{few-shot_contributions}
\begin{table}[h]
	\caption{A layout of the authors’ contributions to the publication: \bibentry{nortje_visually_2024}.}
	\label{tab:few-shot_contributions}
	\centering
	\renewcommand{\arraystretch}{1.2}
	\begin{tabularx}{\linewidth}{@{}lL@{}}
			\toprule
			Author &  Contributions \\
			\midrule
			\leanneName & The layout of the research question, the proposal of the model architecture and the pair mining algorithm. The implementation of the approach on English and \yoruba, and the generation of all the numerical results. \\
			\addlinespace
			\benjiName & The proposal and implementation of \qbert. \\
			\addlinespace
			\danName &  The visualisation analysis of all attention maps, as well as assistance on the in-depth analysis. \\  %
			\addlinespace
			\hermanName &  The layout of the research question and an editorial role. \\ 
			\bottomrule
		\end{tabularx}
\end{table} 

The first author, \leanneNameNoSpace, proposed and implemented all work in the following publication, except as stated otherwise in Table~\ref{tab:few-shot_contributions}.	

\mysubsection{Paper}{Paper}
\label{few-shot_publication}

We now present \RQfewshot, which proposes the \acf{mattnet} to do few-shot learning of natural images and spoken words with fewer word-image examples per class.
We also apply this model to an actual low-resource language, \yoruba.

\newpage\includepdf[pages={1-},scale=1.0]{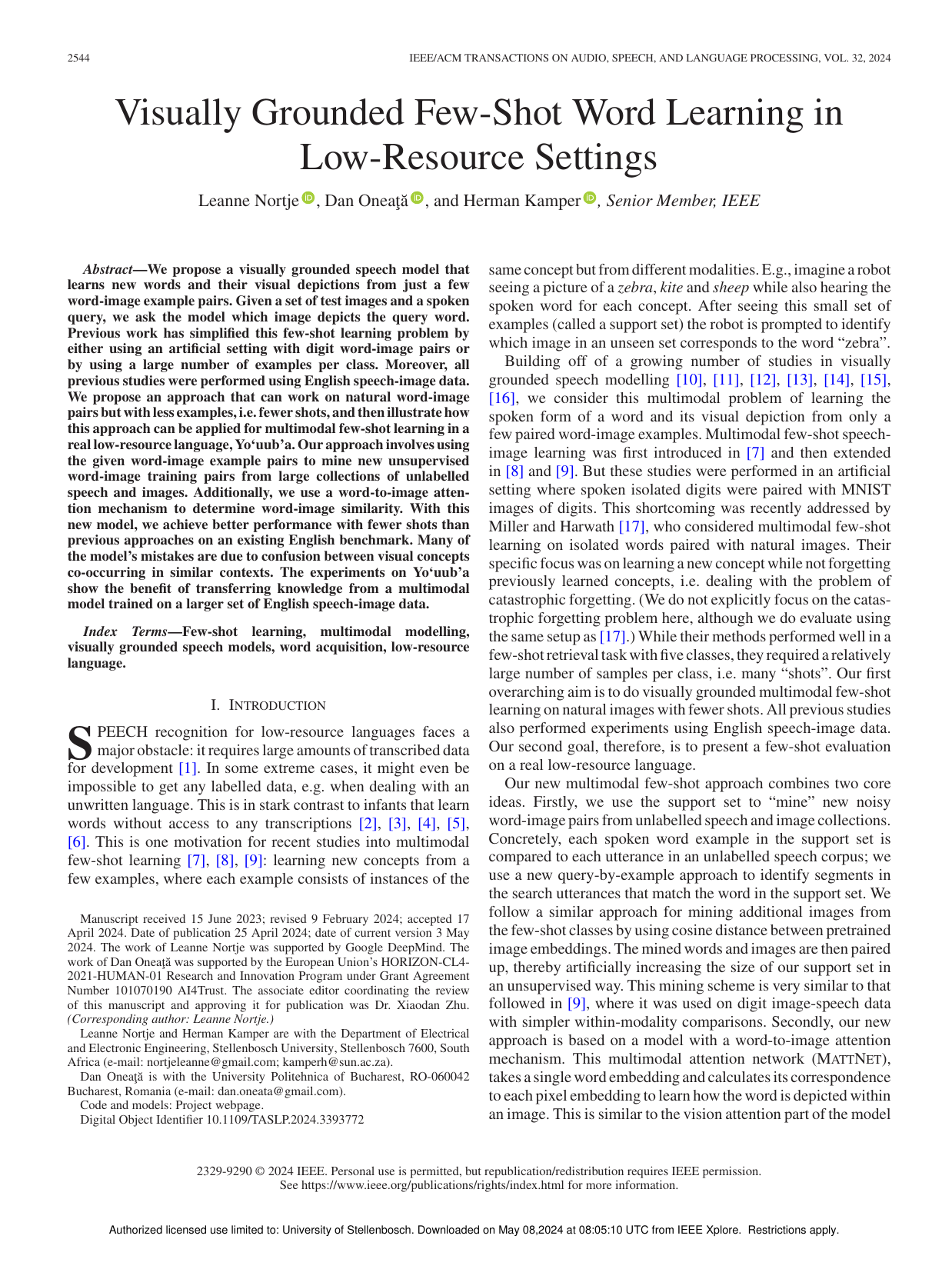} 

\graphicspath{{visually_grounded_few-shot_word_acquisition/figures/}}
\mysection{Further analysis}{Further Analysis}

Although the publication in the previous section covered a thorough analysis of our multimodal few-shot learning approach, we provide even more analysis in this section.
First, we analyse the mined pairs used to train \ac{mattnet} in more depth.
Then, we further analyse the model's resulting attention maps before and after training. 

\mysubsection{Analysing the mined pairs}{Analysing the Mined Pairs}

\begin{table}[!b]
	\caption{The precision in percentages (\%) of the audio and image pairs mined from the $K$ few-shot examples in $\mathcal{S}$.}
	\label{tab:pair_accuracy}
	\centering
	\renewcommand{\arraystretch}{1.2}
	\begin{tabular}{rcc}
		\toprule
		$K$ &  Audio pairs &  Image pairs \\
		\midrule
		5 & 81.1 &  43.9\\
		10 & 83.0 & 47.5\\
		50 & 85.4 & 48.0 \\
		100 & 87.6 & 51.5\\
		\bottomrule
	\end{tabular}
\end{table}

\begin{table}[tb]
	\caption{The $P@N$ few-shot retrieval scores in percentages (\%) for each of the five few-shot classes. $K$ is the number of support-set examples per class. The number of examples in the matching set $\mathcal{M}$ for each class, $N$, is given in brackets.}
	\label{tab:keyword_analysis}
	\centering
	\renewcommand{\arraystretch}{1.2}
	\begin{tabularx}{\linewidth}{@{}lCCCCC@{}}
		\toprule
		& \textit{broccoli}  & \textit{fire hydrant}  & \textit{kite}  & \textit{sheep}  & \textit{zebra}\\
		$K$ & (57) & (62) & (91) & (63) & (90)\\
		\midrule
		5 & 40.4 & 11.3 & 27.5 & 33.3 & 77.8\\
		10 & 49.1 & 9.7 & 30.8 & 33.3 & 85.6\\
		50 & 50.9 & 8.1 & 20.9 & 36.5 & 82.2\\
		100 & 52.6 & 4.8 & 33.0 & 38.1 & 80.0\\
		\bottomrule
	\end{tabularx}
\end{table}

\begin{figure*}[tb]
	\centering
	\includegraphics[width=\linewidth]{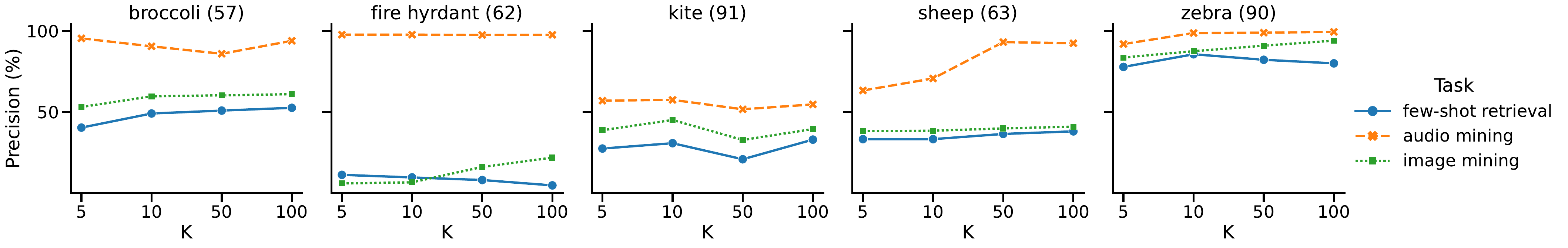}
	\caption{%
		The per class precision in percentages (\%) 
		of the $n=600$ audio and image pairs mined from the $K$ few-shot examples for a class in $\mathcal{S}$.
		The plot also shows the per class retrieval scores in percentages (\%) for \model\ with varying values of $K$.}
	\label{fig:pair_analysis}
\end{figure*}

In Section~III-B of the above-mentioned publication, we describe the process we use to mine training pairs for \model\ from the few ground truth word-image pairs in the support set \supportSet.
Firstly, we use \qbert\ to find matches for each few-shot class' spoken word examples in \supportSet\ from a large collection of unlabelled spoken utterances.
We use the predicted word alignments from \qbert\ to extract possible word matches.
Similarly, we use AlexNet to find image matches for each few-shot class' image examples in \supportSet\ from a large unlabelled image collection.
It is important to note that the word and image pairs are predictions.
\Ie the pairs are not 100\% correct.
Here, we evaluate the influence of the mined pairs on \model 's performance.
We report the precision of both the word and image pairs in Table~\ref{tab:pair_accuracy} (repeated from \RQfewshot), which shows that the precision scores increase as $K$ increases.
Although this is sensible, these aggregated scores do not explain why \model ’s retrieval scores for some classes like fire hydrant, as shown in Table ~\ref{tab:keyword_analysis} (repeated from \RQfewshot), decrease with $K$.
As a result, we consider the per class precision scores for the word pairs in Table~\ref{tab:audio_pair_accuracy} and the image pairs in Table~\ref{tab:image_pair_accuracy}.
Figure~\ref{fig:pair_analysis} summarises the trends we see in the precision scores of the image pairs, the word pairs and the retrieval scores for varying values of $K$.

\begin{table}[tb]
	\caption{The per class precision in percentages (\%) of the $n=600$ audio pairs mined from the $K$ few-shot examples for a class in $\mathcal{S}$.}
	\label{tab:audio_pair_accuracy}
	\centering
	\setlength{\tabcolsep}{0.25em}
	\renewcommand{\arraystretch}{1.2}
	\begin{tabular*}{\linewidth}{r@{\extracolsep{\fill}}ccccc}
		\toprule
		$K$ & \textit{broccoli}  & \textit{fire hydrant}  & \textit{kite}  & \textit{sheep}  & \textit{zebra}\\
		\midrule
		5 & 95.4 & 97.7 & 57.0 & 63.3 & 91.9\\
		10 & 90.5 & 97.7 & 57.5 & 70.7 & 98.7\\
		50 & 85.9 & 97.5 & 51.7 & 93.1 & 98.9\\
		100 & 93.9 & 97.6 & 54.7 & 92.4 & 99.4\\
		\bottomrule
	\end{tabular*}
\end{table}

\begin{table}[tb]
	\caption{The per class precision given in percentages (\%) of the $n=600$ image pairs mined from the $K$ few-shot examples for a class in $\mathcal{S}$.}
	\label{tab:image_pair_accuracy}
	\centering
	\setlength{\tabcolsep}{0.25em}
	\renewcommand{\arraystretch}{1.2}
	\begin{tabular*}{\linewidth}{ l@{\extracolsep{\fill}}ccccc }
		\toprule
		$K$ & \textit{broccoli}  & \textit{fire hydrant}  & \textit{kite}  & \textit{sheep}  & \textit{zebra}\\
		\midrule
		5 & 53.1 & 6.0 & 38.8 & 38.2 & 83.6\\
		10 & 59.7 & 6.7 & 45.1 & 38.5 & 87.5\\
		50 & 60.3 & 16.15 & 32.8 & 39.9 & 90.9\\
		100 & 61.0 & 21.9 & 39.5 & 41.0 & 94.0\\
		\bottomrule
	\end{tabular*}
\end{table}

To explain why the few-shot retrieval scores in Table~\ref{tab:pair_accuracy} for \image{fire hydrant} decrease as $K$ increases, we look at the precision of the word and image \image{fire hydrant} pairs.
However, from Table~\ref{tab:audio_pair_accuracy}, we see that the \word{fire hydrant} word pairs are very accurate and relatively the same across varying values of $K$. 
As a result, we turn to the precision of the \image{fire hydrant} image pairs.
Although these scores are low, the scores still increase as $K$ increases.
These trends for \image{fire hydrant} are visualised in Figure~\ref{fig:pair_analysis}.
We speculate that the precision of the \image{fire hydrant} pairs cannot explain its retrieval scores.
Therefore, the only explanation remaining is the finding of Section~V-B in \RQfewshot.
As K increases for classes with low retrieval scores, where \image{fire hydrant} is one of them, \model\ becomes more dependent on the recurring contextual information.
Contextual information is especially difficult to untangle for objects occupying a small part of images.
As the precision of the image pairs for \image{fire hydrant} increases, more training images contain fire hydrants with street views.
Rather than focusing on the \image{fire hydrant} in an image, \ac{mattnet} focuses on the street views since the \image{fire hydrant} is not prominent enough in the image.

In \RQfewshot, the retention of context information explanation was also given to the fluctuating trend we see in the few-shot retrieval scores of \image{kite} in Table~\ref{tab:pair_accuracy}.
However, the fluctuation trend is also directly reflected in the precision scores of the \textit{kite} image pairs in Table~\ref{tab:image_pair_accuracy}.
Additionally, the precision scores for the \word{kite} word pairs decrease as $K$ increases. 
These trends are visualised in Figure~\ref{fig:pair_analysis}, which shows that the pair precision trends are directly reflected in the retrieval scores.
We speculate that the precision of the mined pairs influences the retrieval scores of \image{kite} and not \image{fire hydrant} since \image{kite} objects occupy a bigger part of images than \textit{fire hydrant} objects.
Nevertheless, the retention of contextual information still has some degree of influence on the \image{kite} retrieval scores, as shown in \RQfewshot. 

Similar explanations can be made for the trends in the other three classes: the few-shot retrieval scores will follow the trend in the precision scores of the image or audio pairs, depending on which has the lowest precision.
\Eg if the image pairs' precision is lower than the audio pairs', the few-shot retrieval scores will follow the trend of the image pairs.
For \textit{broccoli}, the precision of the image pairs increases, and the scores for the word pairs fluctuate.
However, since the overall precision of the image pairs is lower than the audio pairs', the retrieval scores follow the trend in the image pairs' precision. 

The precision of the image and word pairs of \textit{sheep} increase with $K$, but the margins of increase for the word pairs are significantly larger than those of the image pairs. 
This results in the few-shot retrieval scores increasing as $K$ does. 

For \textit{zebra}, the precision scores for both the image and word pairs increase as $K$ increases. 
Surprisingly, the few-shot retrieval scores increase until $K=10$, whereafter the scores decrease with increasing $K$.
Since the \textit{zebra} objects occupy a fair portion of the mined training images and the model accurately localises the \textit{zebra} objects in an image (Figure~5 in \RQfewshot), future work has to investigate this anomaly.

Improving the image pair mining process is crucial for future work.
Not only did this become evident in the above discussion, but from the overall scores in Table~\ref{tab:audio_pair_accuracy}, Table~\ref{tab:image_pair_accuracy} and  Table~\ref{tab:pair_accuracy}, we see that the precision of the image pairs is consistently lower than the precision of the word pairs.
\Ie  the image pairs are possibly the biggest factor limiting the performance of \model.
	\newcommand{\mypath}{visually_grounded_few-shot_word_acquisition/figures/imgs/attention-random/}
	\newcommand{\subsize}{0.9}
	\newcommand{\prefix}{COCO_val2014_}
	\newcommand{\name}{000000092660}
	\newcommand{\link}{-attention-}
	\newcommand{\randomInit}{harwath-random}
	\newcommand{\cpcalexnet}{harwath-independent}
	\newcommand{\background}{harwath-joint-background}
	\newcommand{\trained}{mattnet-joint-few-shot}
	\newcommand{\jpg}{.jpg}
	\begin{figure}
		\centering
		\small
		\caption{The resulting attention maps when prompting \model\ @$K=100$ with a few-shot spoken word and an image that may or may not contain the object corresponding to the word.}
		\label{tab:attention_maps_at_initialisation}
		\begin{tabularx}{\linewidth}{@{}lCCCCC@{}}

			\toprule
			& Original image & Random initialisation & After audio and vision initialisation & After background initialisation & Trained \model \\
			\midrule
			{\small\word{broccoli}} & \includegraphics[width=\subsize\linewidth]{\mypath\prefix\name\jpg} & \includegraphics[width=\subsize\linewidth]{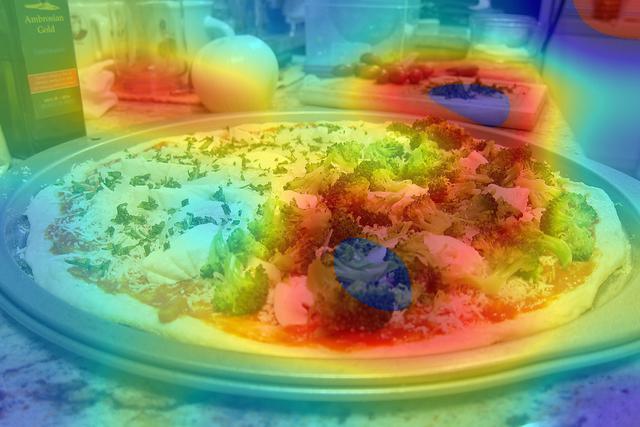} & \includegraphics[width=\subsize\linewidth]{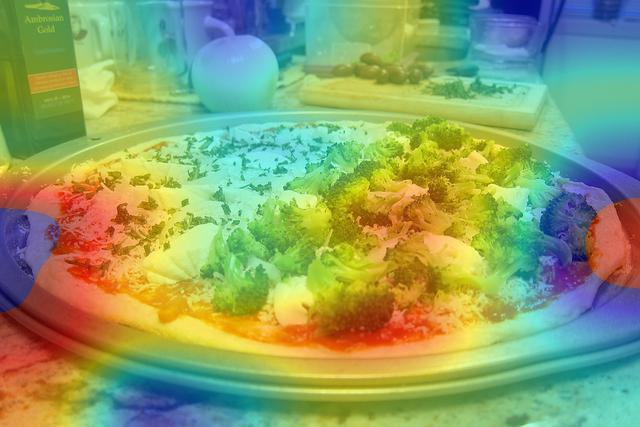} & \includegraphics[width=\subsize\linewidth]{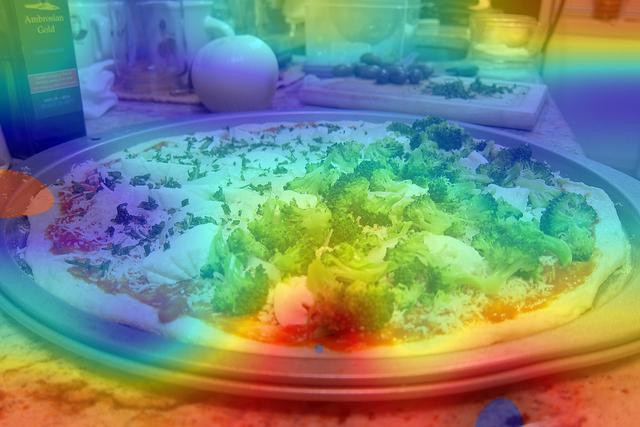}  & \includegraphics[width=\subsize\linewidth]{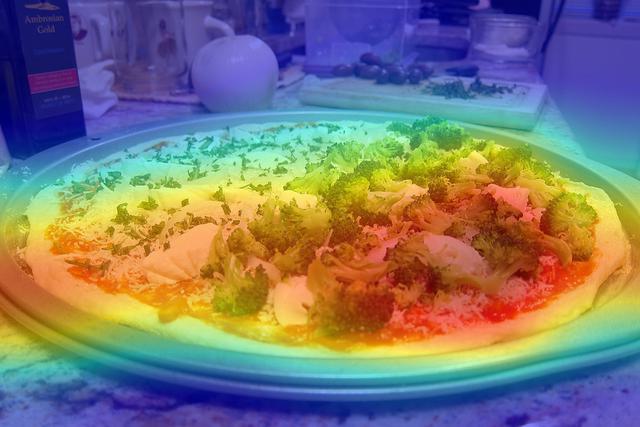}\\
			
			\gdef\name{000000100582}
			{\small\word{broccoli}} & \includegraphics[width=\subsize\linewidth]{\mypath\prefix\name\jpg} & \includegraphics[width=\subsize\linewidth]{\mypath\prefix\name\link\randomInit\jpg} &
			\includegraphics[width=\subsize\linewidth]{\mypath\prefix\name\link\cpcalexnet\jpg} & \includegraphics[width=\subsize\linewidth]{\mypath\prefix\name\link\background\jpg}  & \includegraphics[width=\subsize\linewidth]{\mypath\prefix\name\link\trained\jpg} \\
			
			\gdef\name{000000185409}
			{\small\word{zebra}} & \includegraphics[width=\subsize\linewidth]{\mypath\prefix\name\jpg} & \includegraphics[width=\subsize\linewidth]{\mypath\prefix\name\link\randomInit\jpg} & \includegraphics[width=\subsize\linewidth]{\mypath\prefix\name\link\cpcalexnet\jpg} & \includegraphics[width=\subsize\linewidth]{\mypath\prefix\name\link\background\jpg}  & \includegraphics[width=\subsize\linewidth]{\mypath\prefix\name\link\trained\jpg} \\ 
			
			
			\gdef\name{000000250758}
			{\small\word{zebra}} & \includegraphics[width=\subsize\linewidth]{\mypath\prefix\name\jpg} & \includegraphics[width=\subsize\linewidth]{\mypath\prefix\name\link\randomInit\jpg} & \includegraphics[width=\subsize\linewidth]{\mypath\prefix\name\link\cpcalexnet\jpg} & \includegraphics[width=\subsize\linewidth]{\mypath\prefix\name\link\background\jpg}  & \includegraphics[width=\subsize\linewidth]{\mypath\prefix\name\link\trained\jpg} \\

			\gdef\name{000000314182}
			{\small\word{broccoli}} & \includegraphics[width=\subsize\linewidth]{\mypath\prefix\name\jpg} & \includegraphics[width=\subsize\linewidth]{\mypath\prefix\name\link\randomInit\jpg} & \includegraphics[width=\subsize\linewidth]{\mypath\prefix\name\link\cpcalexnet\jpg} & \includegraphics[width=\subsize\linewidth]{\mypath\prefix\name\link\background\jpg}  & \includegraphics[width=\subsize\linewidth]{\mypath\prefix\name\link\trained\jpg} \\

			\gdef\name{000000501243}
			{\small\word{zebra}} & \includegraphics[width=\subsize\linewidth]{\mypath\prefix\name\jpg} & \includegraphics[width=\subsize\linewidth]{\mypath\prefix\name\link\randomInit\jpg} & \includegraphics[width=\subsize\linewidth]{\mypath\prefix\name\link\cpcalexnet\jpg} & \includegraphics[width=\subsize\linewidth]{\mypath\prefix\name\link\background\jpg}  & \includegraphics[width=\subsize\linewidth]{\mypath\prefix\name\link\trained\jpg} \\

			\gdef\name{000000527528}
			{\small\word{kite}} & \includegraphics[width=\subsize\linewidth]{\mypath\prefix\name\jpg} & \includegraphics[width=\subsize\linewidth]{\mypath\prefix\name\link\randomInit\jpg} & \includegraphics[width=\subsize\linewidth]{\mypath\prefix\name\link\cpcalexnet\jpg} & \includegraphics[width=\subsize\linewidth]{\mypath\prefix\name\link\background\jpg}  & \includegraphics[width=\subsize\linewidth]{\mypath\prefix\name\link\trained\jpg} \\

			\gdef\name{000000546823}
			{ \small\word{fire hydrant}} & \includegraphics[width=\subsize\linewidth]{\mypath\prefix\name\jpg} & \includegraphics[width=\subsize\linewidth]{\mypath\prefix\name\link\randomInit\jpg} & \includegraphics[width=\subsize\linewidth]{\mypath\prefix\name\link\cpcalexnet\jpg} & \includegraphics[width=\subsize\linewidth]{\mypath\prefix\name\link\background\jpg}  & \includegraphics[width=\subsize\linewidth]{\mypath\prefix\name\link\trained\jpg} \\
			
			\gdef\name{000000576654}
			{\small\word{kite}} & \includegraphics[width=\subsize\linewidth]{\mypath\prefix\name\jpg} & \includegraphics[width=\subsize\linewidth]{\mypath\prefix\name\link\randomInit\jpg} & \includegraphics[width=\subsize\linewidth]{\mypath\prefix\name\link\cpcalexnet\jpg} & \includegraphics[width=\subsize\linewidth]{\mypath\prefix\name\link\background\jpg}  & \includegraphics[width=\subsize\linewidth]{\mypath\prefix\name\link\trained\jpg} \\
			
			\gdef\name{000000396205}
			{\small\word{sheep} }& \includegraphics[width=\subsize\linewidth]{\mypath\prefix\name\jpg} & \includegraphics[width=\subsize\linewidth]{\mypath\prefix\name\link\randomInit\jpg} & \includegraphics[width=\subsize\linewidth]{\mypath\prefix\name\link\cpcalexnet\jpg} & \includegraphics[width=\subsize\linewidth]{\mypath\prefix\name\link\background\jpg}  & \includegraphics[width=\subsize\linewidth]{\mypath\prefix\name\link\trained\jpg} \\
			
			\bottomrule
			
		\end{tabularx}
	\end{figure}
	
	\mysubsection{The contribution of pretrained weights}{The Contribution of Pretrained Weights}
	
	Our goal is to find speech-image few-shot models that can be applied to low-resource languages.
	Since previous multimodal few-shot learning studies~\citep{eloff_multimodal_2019, nortje_unsupervised_2020, nortje_direct_2021} found that using transfer learning leads to large performance gains, we use it in our approach.
	We transfer the knowledge gained by a self-supervised speech model that does not require transcribed speech datasets and a supervised object classification model to our \model\ model. 
	We do this by initialising the vision branch with pretrained AlexNet weights~\citep{krizhevsky_imagenet_2017} and the acoustic branch with a self-supervised \ac{CPC} network~\citep{oord_representation_2019} trained on LibriSpeech~\citep{panayotov_librispeech_2015} and multilingual Places~\citep{harwath_vision_2018}.
	After using these unimodal initialisations, we pre-train the network using the speech-image retrieval loss of \citet{harwath_jointly_2018} on unsupervised background speech-image pairs that do not contain instances of the few-shot classes. 
	This background model is the basis on which we fine-tune \model\ on the mined pairs.
	
	We consider the attention maps of a few spoken word and image instances to ensure that the few-shot models learn the few-shot classes during training and that its performance is not just due to the initialisations we use.
	Figure~\ref{tab:attention_maps_at_initialisation} shows the attention maps at various stages of model development: random initialisation (column 3), after \ac{CPC} and AlexNet initialisation (column 4), after initialising the model with the pre-trained background weights (column 5) and after training \model\ at $K=100$ (column 6).
	We also add the original images in column 2 and the spoken word query in column 1 for reference
	
	From the attention maps of the fully trained \model, we see that the maps are focussed on one region in the image that mostly corresponds to the spoken word query.
	On the other hand, the attention maps at random initialisation and after \ac{CPC} and AlexNet initialisation are random.
	The latter makes sense since the layers of the audio branch used to encode input speech into a single embedding are still randomly initialised.
	In some instances, the maps obtained from the pre-trained background model focus on large regions that might include the few-shot class of the spoken query.
	In other instances, the maps focus on all the parts of the image that do not include the few-shot class or the contextual information it associates with the few-shot class.
	Nevertheless, we conclude that pre-training helps \model, but fine-tuning on the mined pairs leads the model to focus on the object corresponding to a spoken few-shot word.
	
	\begin{table}[tb]
		\caption{
			The few-shot retrieval scores in percentages (\%) obtained from \smallmodel\ trained with and without background data. 
			We also give the scores for \smallmodel\ initialised with the pretrained background weights but without using negative background images during fine-tuning on the mined pairs. 	
		}
		\label{tab:background_data_retrieval}
		\centering
		\footnotesize	
		\setlength{\tabcolsep}{0.1em}
		\renewcommand{\arraystretch}{1.2}
		\begin{tabularx}{\linewidth}{@{}lCCCC@{}}
			\toprule
			\multicolumn{1}{c}{Model} & 
			\multicolumn{4}{c}{$K$}\\
			\cline{2-5}
			& 5 & 10 & 50 & 100 \\
			\midrule
			\smallmodel\ & \textbf{40.3$\pm$0.1} & \textbf{44.2$\pm$0.1} & \textbf{41.7$\pm$0.2} & \textbf{43.7$\pm$0.1}\\
			\smallmodel, no background data & 18.1$\pm$1.0 & 23.6$\pm$1.5 & 26.1$\pm$0.2 & 23.1$\pm$0.5\\
			\smallmodel, no background image negatives & 29.9$\pm$0.1 & 31.4$\pm$0.2 & 32.2$\pm$0.2 & 32.1$\pm$0.1\\
			\bottomrule
		\end{tabularx}
	\end{table}
	
	\begin{table}[tb]
		\caption{
			The few-shot classification accuracy (\%) obtained from \smallmodel\ trained with and without background data. 
			We also give the scores for \smallmodel\ initialised with the pretrained background weights but without using negative background images during fine-tuning on the mined pairs. 	
		}
		\label{tab:background_data_classification}
		\centering
		\footnotesize
		\setlength{\tabcolsep}{0.3em}
		\renewcommand{\arraystretch}{1.2}
		\begin{tabularx}{\linewidth}{@{}lCCCC@{}}
			\toprule
			\multicolumn{1}{c}{Model} & \multicolumn{4}{c}{$K$}\\
			\cline{2-5}
			& 5 & 10 & 50 & 100 \\
			\midrule
			\smallmodel\ & 80.1 & 81.1 & 88.5 & 93.2\\
			\smallmodel, no background data & 65.1 & 64.7 & 75.3 & 77.5\\
			\smallmodel, no background image negatives & \textbf{88.0} & \textbf{90.1} & \textbf{94.8} & \textbf{95.3}\\
			\bottomrule
		\end{tabularx}
	\end{table}
	
	Although it is difficult to discern from the qualitative attention maps above, our analysis in Table~\ref{tab:background_data_retrieval} and Table~\ref{tab:background_data_classification}  replicated from \RQfewshot\ concludes that leveraging knowledge from existing unimodal speech and vision models, along with pre-training on large unlabelled datasets, does benefit \model.
	In row 1 of the tables, \model\ is initialised with the pre-trained background model weights and uses background images during fine-tuning on the mined pairs. 
	Row 2 shows the \model\ that only uses the \ac{CPC} and AlexNet initialisations and does not use background pretrained weights or background images during fine-tuning on the mined pairs. 
	The last row shows \model\ initialised with the pretrained background model weights, but no background images are used during fine-tuning on the mined pairs.
	
	Table~\ref{tab:background_data_retrieval} and Table~\ref{tab:background_data_classification} show that initialisation with the pretrained background model is essential.
	The background images used during fine-tuning are necessary to retrieve all the images containing a spoken word query. 
	However, using background images for fine-tuning hurts the model on a classification task in which the spoken word query should be matched to the image depicting the query class in a set containing one image per few-shot class. 
	In short, leveraging background knowledge helps, but fine-tuning on mined pairs enables the model to learn the few-shot classes. 
	
\mysection{Chapter summary}{Chapter Summary}

Our goal was to do multimodal few-shot learning of natural images and spoken words in a low-resource language. 
We proposed a new few-shot pair mining method which we use to train a new multimodal word-to-image attention model: \ac{mattnet}. 
This model outperformed an existing model on a few-shot benchmark where an English spoken word query is used to retrieve all images containing the object corresponding to the query.
Together with this, we set a competitive baseline for natural few-shot word classification on English.

When analysing the few-shot model, we found that it can mostly locate occurrences of a few-shot class’ spoken word in an image.
We also found that many of the model's mistakes are due to the model associating a few-shot class with visual contextual information commonly co-occurring with the class' object.
However, this mainly occurs when a few-shot class occupies a small part of the images and is not the obvious focus in the images.

Lastly, we applied \model\ to an actual low-resource language, \yoruba.
We are the first to do multimodal few-shot learning on a low-resource language.
The multimodal \yoruba\ few-shot model benefited substantially by leveraging a well-resourced language: we initialised the model on a substantial amount of English speech-image data.
On both English and \yoruba\ models, our analysis revealed that the image part of the mining scheme is the major factor limiting performance. 

These multimodal few-shot models are loosely inspired by how children acquire language and make sense of their environment.
As a result, in the next part of the dissertation, we use the model proposed in this chapter to test whether \ac{VGS} models use the same mechanisms as children when learning new words.

\divider{Computational Mutual Exclusivity Studies}
\mychapter{Mutual exclusivity in visually grounded speech models}{Mutual Exclusivity in Visually Grounded Speech Models}{Research Question 3}
\label{chap:me}

\noindent\begin{publicationinfo}{}
	\begin{center}
		{\large\ShortInTextTitle{Chapter~\ref{chap:me} Specifics}}
	\end{center}
	
	\vspace{4pt}
	
	\researchqthreecentered
	
	\vspace{4pt}
	
	\researchpaper{Research Paper \ME}{\bibentry{nortje_visually_2024-1}.}
	
	\vspace{4pt}

	\ShortInTextTitle{Project Website with Code Resources}: \href{https://sites.google.com/view/mutualexclusivityinvgs}{[Link]}
\end{publicationinfo}

The \ac{VGS} models in Chapter~\ref{chap:few-shot} and Chapter~\ref{chap:me}, along with most other \ac{VGS} models, are inspired by how children learn their native language.
Although we construct these networks according to how we envision children learning, we never test whether the models employ the same techniques and biases children use to learn new words and objects.

The \gls{mutual_evclusivity}\acuse{ME} bias is a word learning constraint in which a child would associate a novel word with an object for which the child does not know a name rather than an object for which the child already knows a name.
I.e. if an object has a name, it does not need another; therefore, a new word should be used to name a new object.
This bias results in a one-to-one mapping of objects and words~\citep{markman_childrens_1988}. 
Children's use of this bias has been studied extensively in the developmental sciences~\citep{merriman_mutual_1989, markman_use_2003, mather_learning_2009, lewis_role_2020}. 
With the increasing interest in machine learning in recent years, there has been renewed interest in studying the ME bias using computational approaches.
Various studies have examined whether the ME bias emerges in machine learning models. 
Some studies have investigated under which conditions the bias emerges~\citep{gulordava_deep_2020, gandhi_mutual_2020, vong_cross-situational_2022, ohmer_mutual_2022}.

All these computational studies use models that receive inputs representing words and objects since the ME strategy deals with the mappings between words and the objects they refer to. 
Some studies used symbolic (one-hot) representations of single objects~\citep[e.g.,][]{gandhi_mutual_2020}, while others used continuous vectors that encode natural images~\citep[e.g.,][]{gulordava_deep_2020, vong_cross-situational_2022}.  
Most importantly, all these studies use the textual representations of word inputs. 
A couple of studies encode the text inputs into one-hot vectors~\citep[e.g.,][]{gandhi_mutual_2020}, and others encoded it into continuous word embeddings~\citep[e.g.,][]{gulordava_deep_2020, vong_cross-situational_2022}.
Children, however, learn words from continuous speech. 
A word's written form does not capture the large variation in its spoken form.
A few factors contributing to the variation in speech include word duration, prosody and intonation.
Therefore, children face an additional challenge compared to the computational \ac{ME} models trained on written words.
The \ac{ME} bias has not been computationally studied in \ac{VGS} models that learn from images paired with unlabelled speech ~\citep{harwath_unsupervised_2016, harwath_vision_2018, kamper_semantic_2019, chrupala_visually_2022, peng_fast-slow_2022, peng_syllable_2023, berry_m-speechclip_2023, shih_speechclip_2023}.

In this chapter, we aim to study the \ac{ME} bias computationally in a naturalistic setting using natural images and spoken words.
\rqthree\ states: \researchquestionthree\ 
\RQME\ attempts to answer this by constructing a speech-image setup to evaluate the \ac{ME} bias in \ac{VGS} models.
We use the \ac{VGS} model architecture proposed in Chapter~\ref{chap:few-shot} and train it on speech-image pairs from a set of familiar classes. 
We evaluate the \ac{ME} bias by prompting the model on novel samples it has never encountered.
We evaluate the strength of the \ac{ME} bias when using various initialisations for the audio and vision branches that simulate the prior knowledge a child will have by the time they start using the bias.
After we find that the model with both audio and vision initialisations has the strongest \ac{ME} bias, we analyse the model's representation space and consider various loss functions to establish how specific our findings are to our model design.

\mysection{Related work}{Related Work}

\citet{barrett_lexical_1978} was the first to introduce the hypothesis that children use a \acf[first-style=long-short]{ME} bias to learn new words.
Hereafter, \citet{markman_childrens_1988} coined the term ``mutual exclusivity'' to refer to this bias.
The theory was devised since most word categories are mutually exclusive.
Therefore, they speculated that children initially assume all word categories are mutually exclusive until they encounter evidence to prove different. %
\Eg a child will assume the words \word{husky} and \word{dog} are mutually exclusive until they are provided with information stating that a \word{husky} is a \word{dog}. 

In a broader context, \citet{barrett_lexical_1978} claims that children use the \ac{ME} bias in a contrastive learning approach to learn class hierarchies and the semantic links between words.
\Eg a child will assume the words \word{pet}, \word{husky}, \word{spaniel} and \word{dog} are mutually exclusive upon hearing it for the first time. 
After being given more information, the child learns that \word{husky} belongs to the class \word{dog} and so does \word{spaniel}, but \word{husky} and \word{spaniel} are still mutually exclusive. 
The child will then learn the semantic link between \word{pet} and \word{dog} and then infer that \word{husky} and \word{spaniel} have a semantic link to \word{pet}.

Formally, the \ac{ME} bias states that once an object has a name, it does not need another. 
As a result, a novel word should belong to a novel image~\citep{markman_childrens_1988}.
Several studies showed that children across various ages during early childhood use the \ac{ME} bias~\citep{bion_fast_2013, halberda_development_2003, merriman_mutual_1989}.
It has also been hypothesised that the bias aids children in inferring the meaning of words referring to a feature or part of an object~\citep{markman_constraints_1990}.
\Eg say the child knows the word \word{dog} and what a \image{dog} looks like, as well as the word \word{tree} and what a \image{tree} looks like.
Now, the child is shown new images of a \image{dog} and a \image{tree} and asked, \word{Show me the paw}.
Since the names of the two objects are known, this novel word \word{paw} should refer to a part or feature of one of the familiar objects. 

Other studies looked into the effect of linguistic experience on the bias.
The common conclusion was that with greater linguistic experience, \eg encountering more speech, children will likely have a stronger \ac{ME} bias~\citep{golinkoff_early_1994, mervis_two-year-olds_1994, markman_use_2003, grassmann_childrens_2015, merriman_reasons_1986}.
Some studies went into even more detail by showing that children with larger vocabularies tend to show a stronger \ac{ME} bias~\citep{bion_fast_2013, deak_hunting_2000, grassmann_childrens_2015, law_effects_2015}.

It is clear that the \ac{ME} bias is a word acquisition strategy widely researched in children.
On the other hand, word acquisition is a widely studied phenomenon in machine learning.
Therefore, a comparison between the \ac{ME} bias in children and machine learning methods would be valuable.
We aim to investigate the \ac{ME} bias specifically in \ac{VGS}.
To do this, we replicate the \ac{ME} task children were tasked with in the developmental science studies: a child is shown a familiar object, e.g. \image{banana}, and a novel (unknown) object, e.g. \image{mango}.
The child is then asked: \word{Show me the dax}, where \word{dax} is a novel word.
They found that children had the tendency to choose the novel object \image{mango} instead of the familiar \image{banana}. 

On a similar task using text instead of speech, \cite{gulordava_deep_2020} were the first to test whether a machine learning model using natural images exhibits the bias. 
We also use natural images, but we use speech instead of text for the word representations. 
Another difference between our study and theirs is that we aim to establish whether the \ac{ME} bias naturally occurs within the model, and they aim to investigate how they can enforce the \ac{ME} bias in their model.
More specifically, they use objective functions to mimic the following learning biases: (1) a word class can only be mapped to a single object class, (2) an object class can only be mapped to a single word class, or (3) a combination of (1) and (2) resulting in the traditional \ac{ME} bias' one-to-one mapping between word and object classes.
Specifically, they use a max-margin hinge loss using either (1) negative object samples, (2) negative word samples or (3) both negative word and object samples.
We use a similar contrastive objective to learn a set of familiar classes with negative object and word samples. 
For a given familiar class, we only use negative samples from other familiar classes.
In contrast, \cite{gulordava_deep_2020} use samples from their novel classes as negative samples.

Similarly to \cite{gulordava_deep_2020}, \cite{vong_cross-situational_2022} use natural images paired with textual word labels in their computational \ac{ME} study.
However,  similar to us, they investigate whether the \ac{ME} bias naturally emerges from their machine learning network. 
Where we consider a speech-image model, they consider an image-text model trained on trials using a hinge loss: a trial contains $n$ number of text labels and images where each image contains a single isolated object.
These images and labels are not paired. 
By sampling negative trials, the hope is to learn which label fits to which image.
\Eg say we have trial \textit{A} with an image set \{\image{ball}, \image{lemon} and \image{zebra}\} and labels \{\textLabel{zebra}, \textLabel{ball} and \textLabel{lemon}\}, and trial \textit{B} with an  image set \{\image{cookie}, \image{elephant} and \image{zebra}\} and labels \{\textLabel{zebra}, \textLabel{cookie} and \textLabel{elephant}\}.
The two images that overlap in the two trails are \image{zebra}, and the two labels that overlap are \textLabel{zebra}.
As a result, the \image{zebra} images should belong to \textLabel{zebra} labels.
Before testing whether the model exhibits a \ac{ME} bias and finding that the bias did not hold over varying model parameters, they performed one gradient update on trails containing the novel classes.
This updating of the model weights on the novel classes results in the novel classes being somewhat known to the model.

Since \citet{gulordava_deep_2020} and \citeauthor{vong_cross-situational_2022}'s~(\citeyear{vong_cross-situational_2022}) models take written words that are inherently discrete as inputs, they have to learn a continuous embedding for each input class. 
If the word embeddings of novel classes are never updated, the embeddings remain randomly initialised.
\Ie the \ac{ME} test will compare learned vs random embeddings instead of novel vs familiar. 
Since updating the word embeddings for each class in models that take written input is necessary, the model has seen the novel classes before testing.
However, \ac{VGS} models can place an input it has never encountered in the representation space learned from the familiar classes.
Therefore, we consider whether the \ac{ME} bias naturally emerges from a \ac{VGS} model.

Another study that asks whether the \ac{ME} bias naturally emerges in a machine learning model, is \cite{gandhi_mutual_2020} which considered whether the \ac{ME} bias naturally develops over time through training.
Or is the \ac{ME} bias a built-in assumption?
To answer this, they used a neural network that takes one-hot representations for words and images as inputs. 
They showed that explicitly forcing the model to learn using a \ac{ME} objective could lead to learning new words.
Similarly, \citet{nematzadeh_competition_2020} incorporate the \ac{ME} bias -- the one-to-one mapping rule -- into its structure to improve word learning.
They implement a probabilistic model that attempts to learn an alignment between misaligned textual words and visual referents, similar to \citet{vong_cross-situational_2022}.
Both the studies of \citet{gandhi_mutual_2020} and \citet{nematzadeh_competition_2020} incorporate the \ac{ME} bias into their architecture to learn word categories. 
Again, we consider whether a \ac{VGS} model naturally exhibits a \ac{ME} bias.
 
\citet{ohmer_mutual_2022} considered whether an image-text model naturally exhibits the \ac{ME} bias.
They used the MNIST digit images paired with textual word labels to train a model and test whether it exhibits the \ac{ME} bias. 
Although we have the same overarching goal, we do this in a more realistic setting using natural images and spoken words instead of text word labels.
We are the first to use speech to computationally study the \ac{ME} bias.

\mysection{Publication: Research paper \ME}{Publication: Research Paper \ME}
\label{me_publication}

The publication in this section (\RQME) computationally studies whether a \ac{VGS} model naturally exhibits the \ac{ME} bias. 
The publication creates a speech-image setup to evaluate the \ac{ME} bias in \ac{VGS} models.
We train the \ac{VGS} model architecture of Chapter~\ref{chap:few-shot} on speech and image pairs from a set of familiar classes. 
After training, we apply the model to novel samples to establish whether it exhibits the \ac{ME} bias. 
The publication delves into various combinations of initialising the audio and vision branches of the model with prior knowledge similar to what a child will have. 
Additionally, the publication performs a finer-grained analysis of the model with the strongest \ac{ME} bias and considers various loss functions to evaluate how specific our findings are to our model design.

\mysubsection{Contribution declaration}{Contribution Declaration}

\begin{table}[h]
	\caption{A layout of the authors’ contributions to the publication: \bibentry{nortje_visually_2024-1}.}
	\label{tab:me_contributions}
	\centering
	\renewcommand{\arraystretch}{1.2}
	\begin{tabularx}{\linewidth}{@{}lL@{}}
		\toprule
		Author &  Contributions \\
		\midrule
		\leanneName & The layout of the research question, the model architecture and the data setup. The implementation and the generation of the numerical results and graphs. \\
		\addlinespace
		\danName &  The per keyword \ac{ME} analysis. \\  %
		\addlinespace
		\yevgenName & The execution of the significance tests. \\ 
		\addlinespace
		\hermanName & The layout of the research question and an editorial role.\\
		\bottomrule
	\end{tabularx}
\end{table} 

All work done in the publication was proposed and implemented by \leanneNameNoSpace, the author of this dissertation, except as stated otherwise in Table~\ref{tab:me_contributions}. 

\mysubsection{Paper}{Paper}

\begin{table}[tb]
	\caption{The notation we use in \RQMultilingualME\ and how it relates to the notation we use earlier in the dissertation.  }
	\label{tab:notation}
	\centering
	\renewcommand{\arraystretch}{1.2}
	\begin{tabularx}{0.7\linewidth}{@{}lCC@{}}
		\toprule
		 &  Notation used earlier in this dissertation & Notation used in the following publication \\
		\midrule
		Image & \image{image} & \textLabel{image}\\
		Word & \word{word} & \image{word}\\
		\bottomrule
	\end{tabularx}
\end{table} 

The following publication is \RQME.
In this publication, we break with the notation we use throughout this dissertation and the publications in Chapter~\ref{chap:few-shot} and Chapter~\ref{chap:me}.
This publication adopts the convention used by computational cognitive studies. 
Table~\ref{tab:notation} shows the differences in notation we used to refer to images and words previously and what we use in the next publication.
Usually, we would use the following notation to refer to an image, \eg of a \image{dog}. 
In this publication, we use the following notation to refer to an image: \textLabel{dog}.
Sticking with the example of a \textLabel{dog}, we usually refer to a spoken word as \word{dog}.
In the publication, we refer to the word as \image{dog}.

\begin{figure}[!t]
	\centering
	\includegraphics[width=0.6\linewidth]{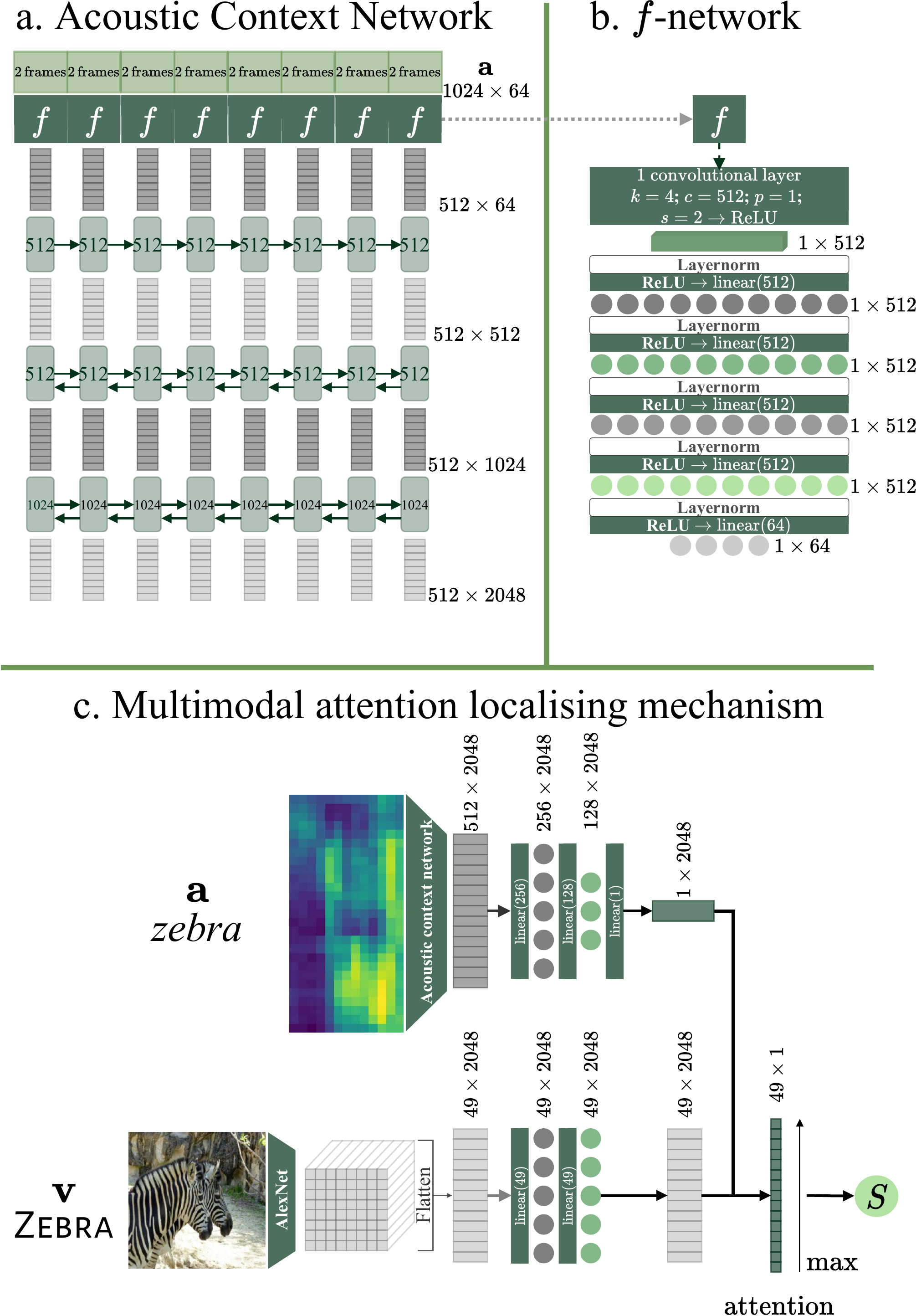}
	\caption{%
		(c) The \model\ architecture proposed in Chapter~\ref{chap:few-shot} consists of a vision network and an audio network (a+b) connected with a word-to-image attention mechanism. The mechanism outputs a similarity score for a speech and image input.%
	}
	\label{fig:me_model}
\end{figure}

We provide the full model architecture in Figure~\ref{fig:me_model} for more background when reading the publication. 
Specifically, we add the architecture of the audio branch in Figure~\ref{fig:me_model}a and Figure~\ref{fig:me_model}b. 
We now delve into the \model\ model trained on natural images and spoken words from familiar classes and investigate whether this model exhibits the \ac{ME} bias. 

\newpage~\newpage
\includepdf[pages={1-},scale=1.0]{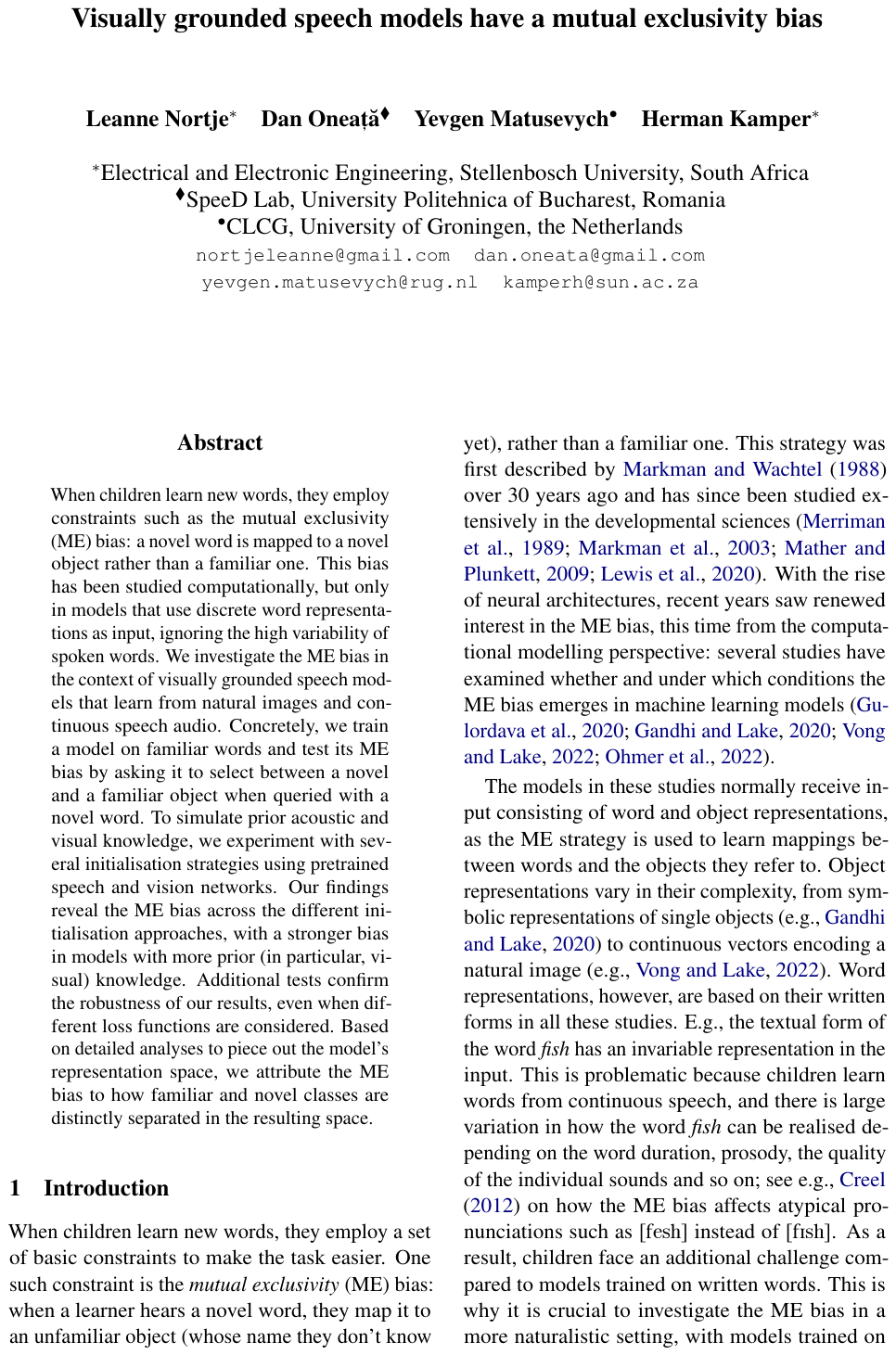} 

\mysection{Multimodal \ac{ME} dataset generation}{Multimodal \ac{ME} Dataset Generation}
\label{chap_me:dataset_generation}

In \RQME, we aim to investigate whether a \ac{VGS} model would naturally choose a novel object rather than a familiar one when prompted with a novel spoken word. 
\Ie test whether a \ac{VGS} model exhibits the \ac{ME} bias.
To do this, we constructed a dataset consisting of isolated spoken words and natural images. 
This dataset contains natural images of entire scenes for training and images of isolated natural objects for evaluating the inherent \ac{ME} bias of a model.
The training and testing sets also contain isolated spoken words.
To get a dataset that meets these requirements, we combined multiple image and speech datasets to get one large dataset of spoken word and image pairs for familiar and novel classes.  
We now give extra details regarding the dataset construction that was not given in the paper.

For the image part of the dataset, we use \mscoco~\citep{lin_microsoft_2014}, Caltech-101~\citep{fei-fei_one-shot_2006} and ImageNet~\citep{deng_imagenet_2009}, where the latter is mistakenly omitted from \RQME. 
We chose these datasets since they have object segmentations available.
\mscoco\ contains 328k natural images of 91 objects, and
Caltech-101 contains 9k natural Google images from 101 classes.
ImageNet contains 14M images organised according to WordNet~\citep{miller_wordnet_1995}.
We use entire natural images for training and images containing isolated natural objects for the development and test sets.
 We use the ground truth object segmentations to isolate the familiar and novel objects under consideration in the test and development images.

 For the audio part of the dataset, we use the Flickr8k Audio captions corpus~\citep{harwath_deep_2015}, the Buckeye corpus~\citep{pitt_mark_buckeye_2005} and the LibriSpeech corpus~\citep{panayotov_librispeech_2015}.
 The Flickr8k Audio corpus contains five parallel spoken English captions for each of the 8k images.
 We only use the audio utterances in this dataset and not the Flickr images since no ground truth object segmentations are available for the images.
 The Buckeye corpus of English conversational speech is recorded from 40 speakers.
 The LibriSpeech set is collected from the LibriVox project resulting in 982\,hours of audio from 1\,201 female and 1\,283 male speakers.
 We use forced alignments to isolate spoken words for the familiar and novel classes. 
 We use the Montreal forced aligner~\citep{mcauliffe_montreal_2017} to obtain forced alignments where alignments are not available.

\begin{table}[!b]
	\small
	\centering
	\renewcommand{\arraystretch}{1.2}
	\caption{The familiar and novel classes used in the ME setup.}
	\begin{tabularx}{\linewidth}{@{}lL@{}}
		\toprule
		Familiar classes& bear, bird, boat, car, cat, clock, cow, dog, elephant, horse, scissors, sheep, umbrella \\
		\addlinespace
		Novel classes & ball, barrel, bench, buck, bus, butterfly, cake, camera, canon, chair, cup, fan, fork, guitar, lamp, nautilus, piano, revolver, toilet, trumpet \\
		\bottomrule
	\end{tabularx}
	\label{tab:classes}
\end{table}

\begin{figure}[!t]
	\centering
	\includegraphics[width=0.9\linewidth]{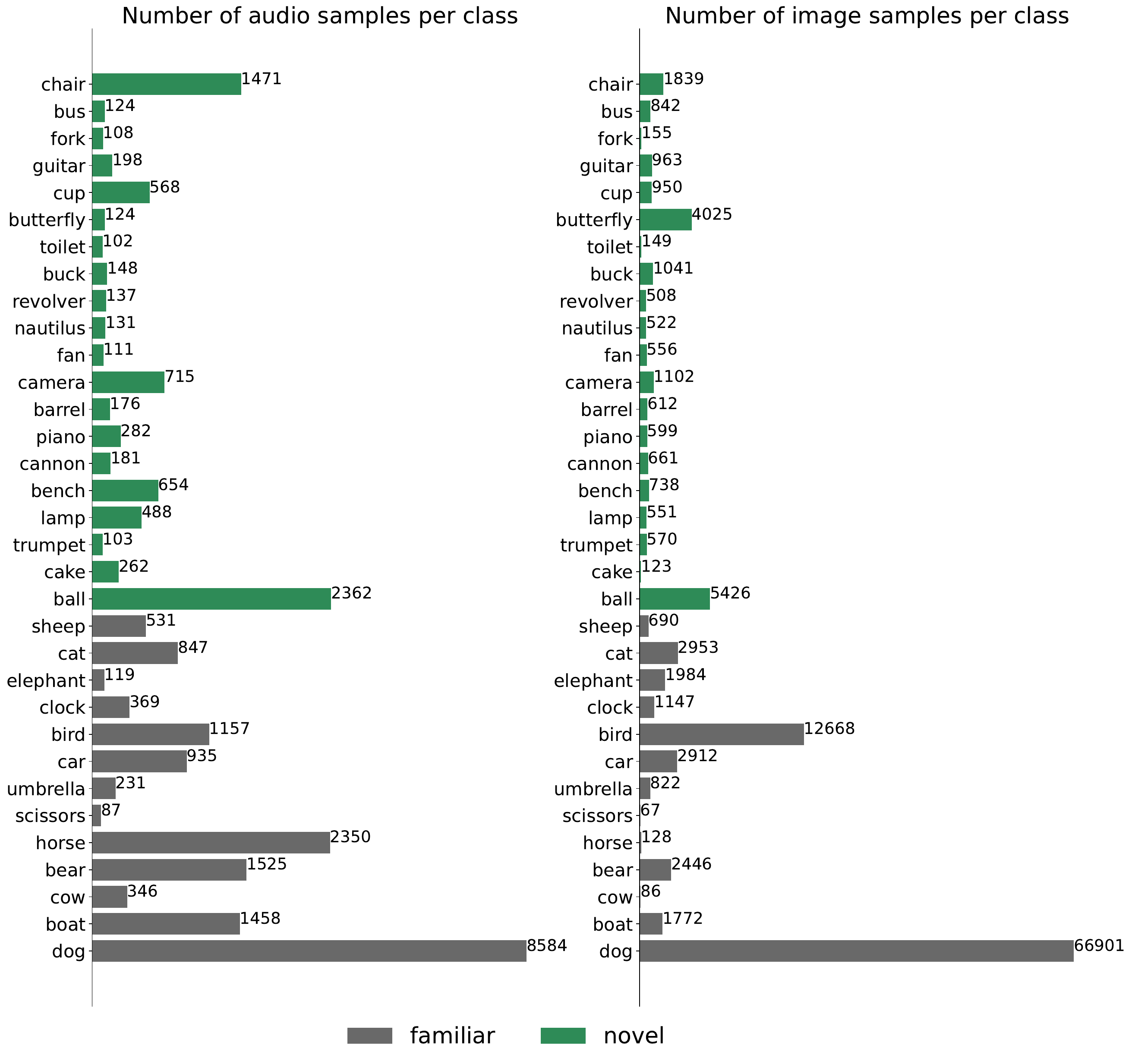}
	\caption{%
		The amount of spoken words and images for each novel and familiar class in the speech-image \ac{ME} dataset we construct. 
	}
	\label{fig:me_dataset_analysis}
\end{figure}

We divide this large multimodal dataset we just created into 13 familiar classes and 20 novel classes.
These classes are shown in Table~\ref{tab:classes} (repeated from Table~1 of \RQME).
Figure~\ref{fig:me_dataset_analysis} shows the amount of spoken words and images for each familiar and novel class in the generated dataset.
Next, we briefly set out the class selection process.

\begin{enumerate}[label=\textbf{Step \arabic*}:, leftmargin=*, parsep=0cm, itemsep=0em]
	\item We manually evaluate which classes of the image datasets are viable.
	I.e. which classes can be isolated and still be legible. 
	For example, the class \textit{curtains} are excluded since you cannot tell from the segmented images that \image{curtains} are depicted.
	\item We resolve any conflicting classes. 
	Examples of conflicting classes are \image{dog} and \image{dalmatian}, as well as \image{pigeon} and \image{bird}. 
	Therefore, images from subclasses like \image{dalmatian} and \image{husky} are added to the main \image{dog} class.
	\item In the next chapter, we extend this dataset to be multilingual by including Dutch and French spoken word translations of the familiar classes. 
	For the models in this chapter to be comparable to the multilingual models in the next chapter, we choose familiar classes available in English, Dutch and French speech datasets.
	For a given English class, if there are no word samples for the Dutch and French translations of the class available in the Dutch and French datasets we use, we exclude the class. 
	We also exclude a class if there is any overlap in the class's English, Dutch or French translations.
	\Eg the Dutch word \word{camera} and the English word \word{camera} for \image{camera} are the same. 
	
	If any English classes are excluded in this step and there are more than ten word and ten image samples, we use the class as a novel class since we only need English test samples.
	We leave a discussion of the Dutch and French datasets for the next chapter.
	\item For the familiar set of classes, we check that there are more or less 100 image and 100 audio samples in each language. 
	If there are not enough samples for a class, we add it to the novel set, given that there are more than ten word and image samples.
\end{enumerate}

In this chapter, we train \model\ using the training set containing spoken English words and natural images of entire scenes for the 13 familiar classes. 
To evaluate whether the model has an inherent ME bias, we use the test set containing spoken English words and images of isolated natural objects for the 20 novel and 13 familiar classes.
The development set contains ten spoken words, each paired with an image of an isolated natural object for each of the 13 familiar classes.
We use this set on the validation task for early stopping.
We ensure that the test, development and training samples do not overlap.

\graphicspath{{mutual_exclusivity/figures/}}
\mysection{Further Analysis}{Further Analysis}

In \RQME, we set out to establish whether the \ac{VGS} model, \model, proposed in Chapter~\ref{chap:few-shot}, exhibits a ME bias.
After we established that the model shows a robust \ac{ME} bias and that using prior unimodal vision and audio knowledge strengthens the \ac{ME} bias, we covered a thorough analysis of \model.
This included an analysis of the learned representation space and various experiments to ensure that we are indeed witnessing the \ac{ME} bias.
To paint a clearer picture of how the learned representation space is organised, we now give a visualisation of clustered speech and image embeddings generated by the model with the strongest \ac{ME} bias. 
We also investigate whether any biases are introduced by combining the various datasets (Section~\ref{chap_me:dataset_generation}). 

\mysubsection{Do dataset biases appear in the constructed \ac{ME} dataset?}{Do Dataset Biases Appear in the Constructed \ac{ME} Dataset?}
\label{chap_me:dataset_biases}

In \RQME, we proved that \model\ does not learn an explicit bias against novel items, \ie the model does not simply choose a novel object when it encounters one regardless of whether the spoken word query is from a novel or familiar class.
We also found that the model does not distinguish between the different novel classes but instead sees an object it does not know simply as `novel'.
However, we never analysed whether the model is biased towards samples from one of the datasets we use to generate the \ac{ME} dataset.
When creating the test setup, we ensured that the novel and familiar images the model has to choose between, after being prompted with a spoken query, are from the same image dataset to ensure no dataset biases within a comparison.
However, it may be that the model is better at distinguishing novel from familiar for one dataset than for another, thus resulting in a dataset bias.

\newcommand{\overall}{grey}
\newcommand{\buckeye}{purple}
\newcommand{\mscocoColour}{yellow}
\newcommand{\caltech}{blue}
\newcommand{\librispeech}{green}
\newcommand{\flickr}{orange}

\begin{figure*}[tb]
	\subfloat[The accuracies of \model\ without any warm starts.]{\includegraphics[width=0.5\linewidth]{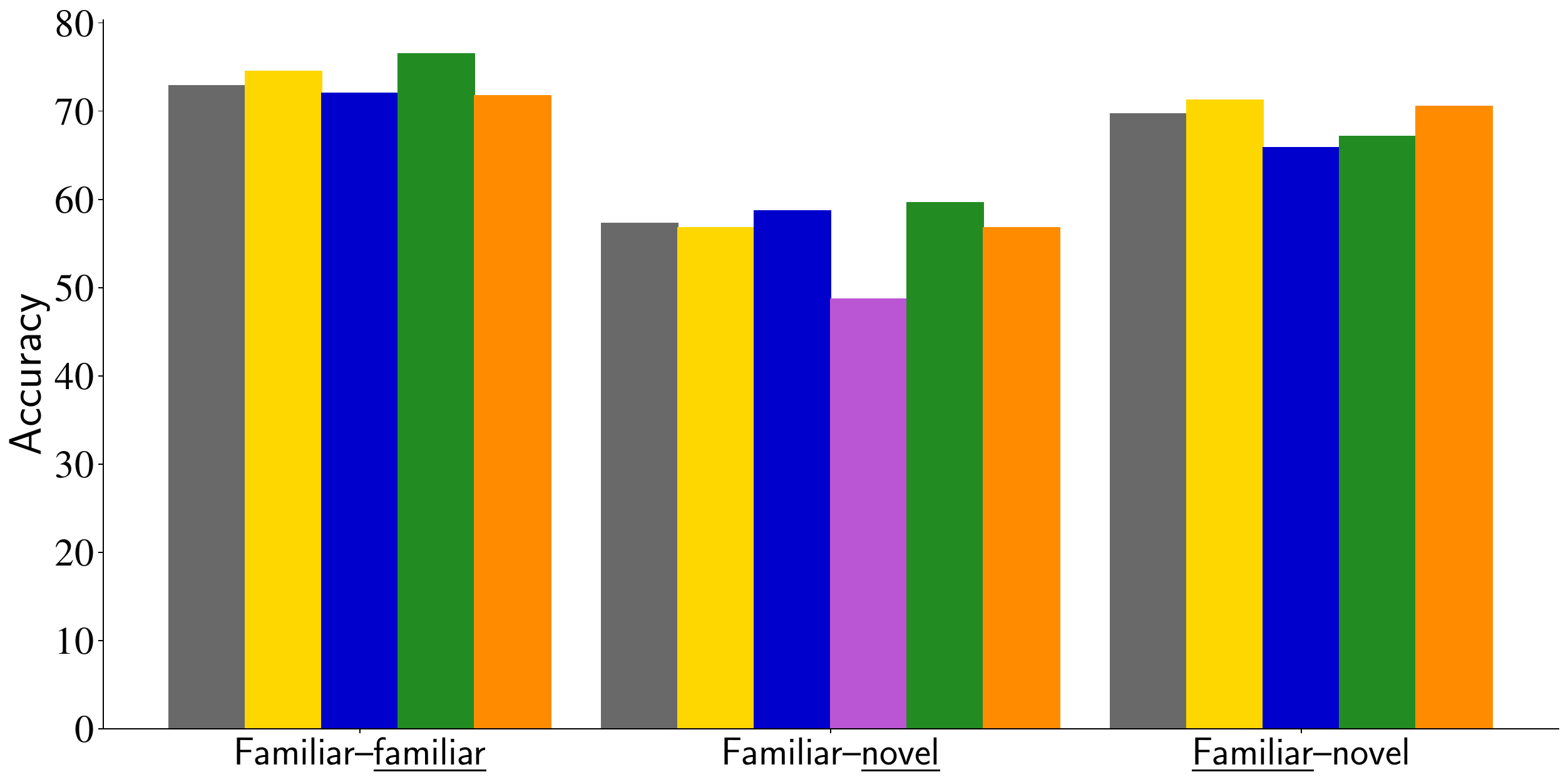}
		\label{fig:no_warm_start_image_discrepancy_test}}
	\hspace{0.1cm}
	\subfloat[The accuracies of \model\ with AlexNet warm start.]{\includegraphics[width=0.5\linewidth]{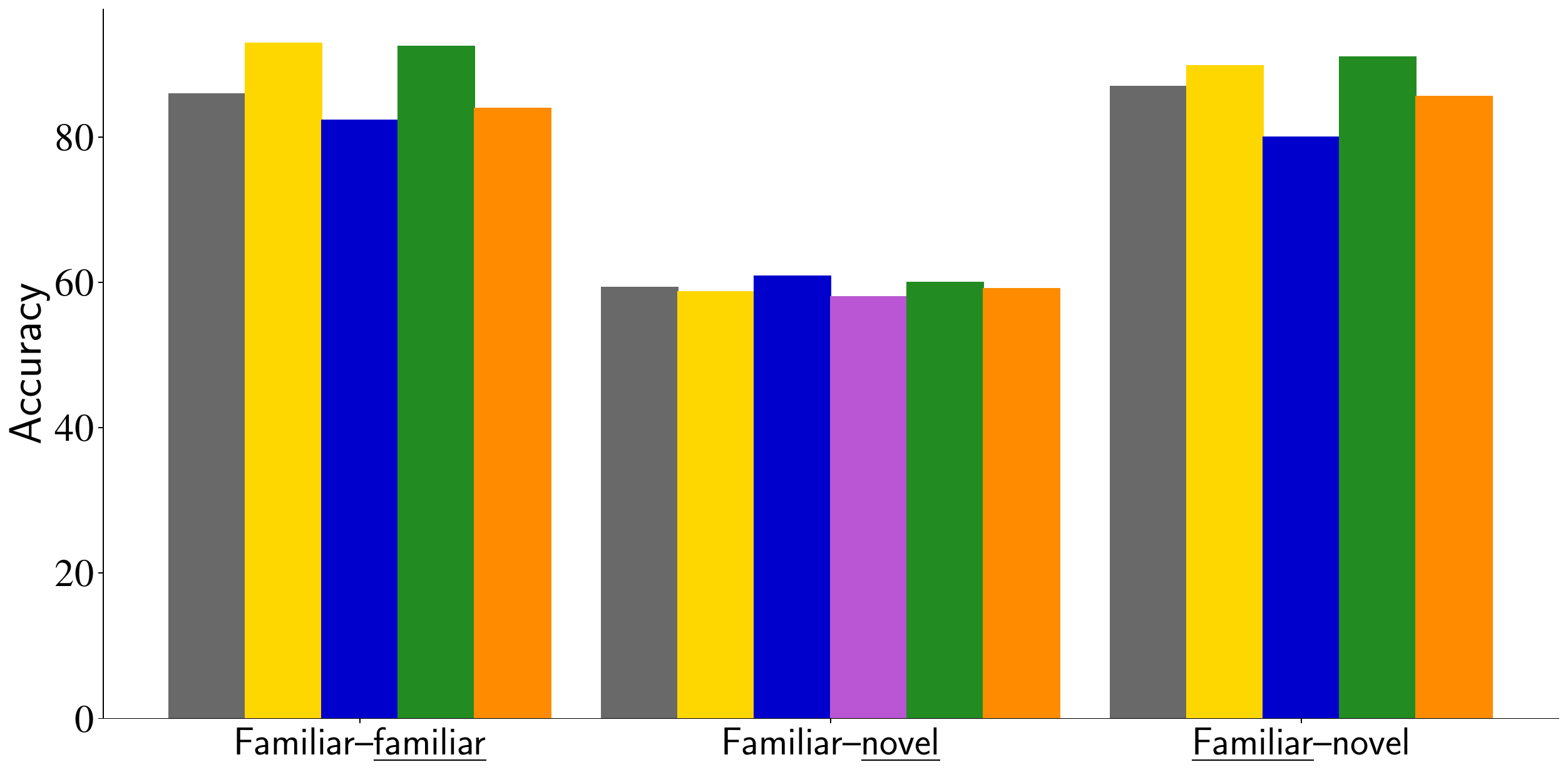}
		\label{fig:alexnet_warm_start_image_discrepancy_test}}
	
	\subfloat[The accuracies of \model\ with CPC warm start.]{\includegraphics[width=0.5\linewidth]{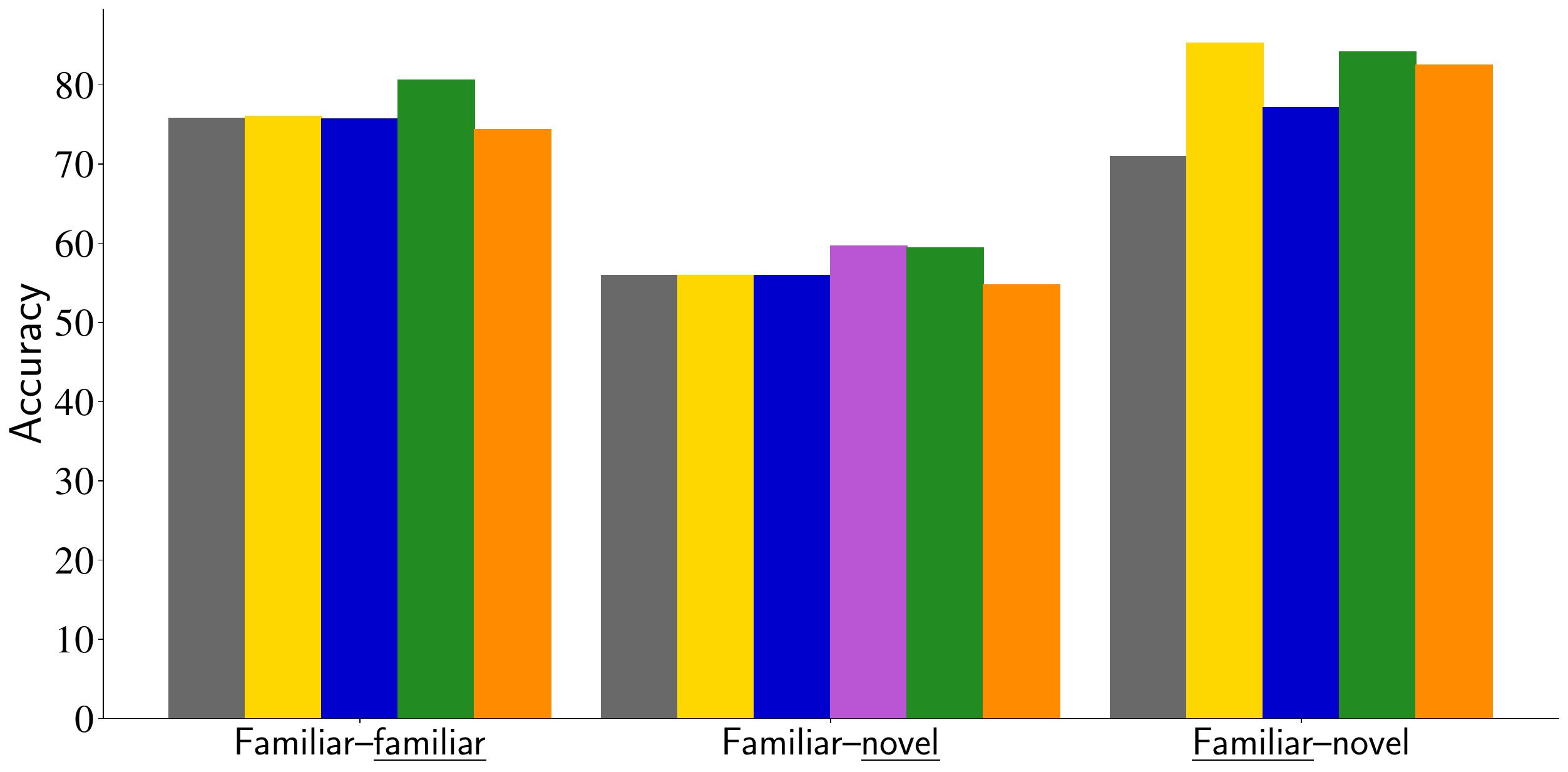}
		\label{fig:cpc_warm_start_image_discrepancy_test}}
	\hspace{0.1cm}
	\subfloat[The accuracies of \model\ with AlexNet and CPC warm starts.]{\includegraphics[width=0.5\linewidth]{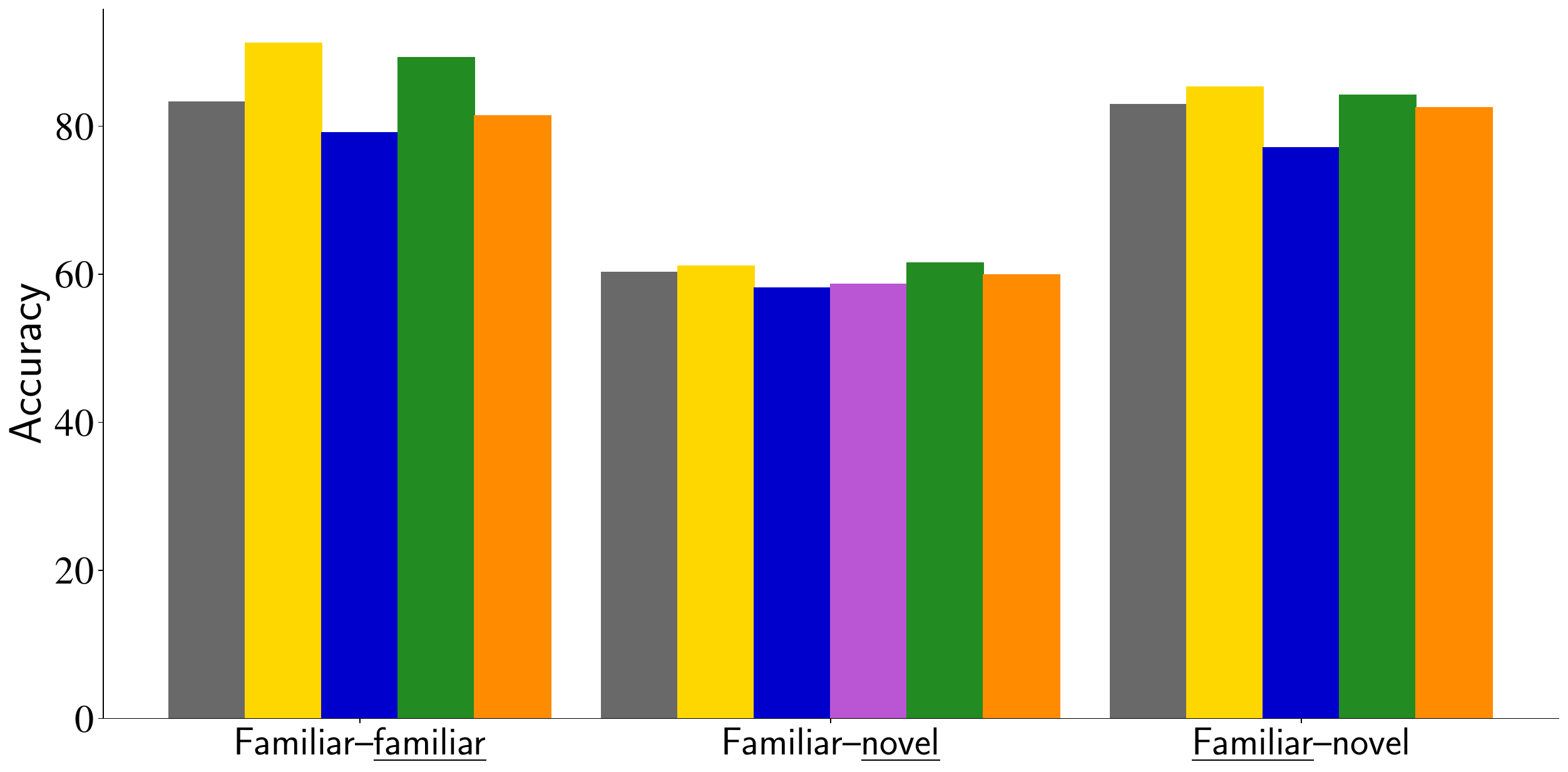}
		\label{fig:alexnet_cpc_warm_start_image_discrepancy_test}}
	
	{\hspace*{\fill} \includegraphics[width=0.4\linewidth]{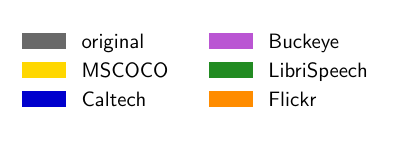} \hspace*{\fill}}
	\caption{
		The results of the disentangled \familiar, \me\ and \familiarWithNovel\ scores for each dataset. 
		The score for an image dataset considers the comparisons where the two images are from this image dataset and the spoken query is from any of the speech datasets. Similarly, the score for a speech dataset considers the comparisons where the spoken query is from this speech dataset and the two images are both from any one image dataset. The image datasets we consider are Caltech-101 and \mscoco, and the speech datasets are LibriSpeech, Buckeye or Flickr. The combination score is calculated on the full \ac{ME} episodes containing all the dataset instances. }
	\label{fig:data_discrepancy_tests}
\end{figure*}

In Figure~\ref{fig:data_discrepancy_tests}, we show the disentangled \familiar, \me\ and \familiarWithNovel\ scores. 
More specifically, we report the scores per dataset we use to construct our \ac{ME} dataset.
\Eg to get the \librispeech\ bars in Figure~\ref{fig:data_discrepancy_tests}, we prompt the model with every LibriSpeech word query in the test episodes and ask which one of two images each query belongs to. %
Both images in such a comparison are from the same image dataset but this image dataset can be any one of the three we use. 
The \overall\ bars are the scores on the original test episodes before disentangling them into the various datasets.
The \mscocoColour\ bars contain the test instances where the two images are from the \mscoco\ dataset, and the spoken word query is from any speech dataset.
We do the same for the \caltech\ bars, but the two images are from the Caltech-101 dataset. 
Although we also use ImageNet to construct our \ac{ME} dataset, we do not include an ImageNet bar in the plots since none of the image instances in the test episodes are from the ImageNet dataset.

In Figure~\ref{fig:no_warm_start_image_discrepancy_test}, we consider the \model\ variation without any initialisations.
We observe that the scores for the \familiar, \me\ or \familiarWithNovel\ setups stay relatively consistent for the Caltech-101, \mscoco\ and original tests.
This holds for the other \model\ variants using only the AlexNet initialisation (Figure~\ref{fig:alexnet_warm_start_image_discrepancy_test}), only the \ac{CPC} initialisation (Figure~\ref{fig:cpc_warm_start_image_discrepancy_test}) and both the AlexNet and \ac{CPC} initialisations (Figure~\ref{fig:alexnet_cpc_warm_start_image_discrepancy_test}).

Now, we do the same for the audio datasets: the \buckeye\ bars contain the test instances where the spoken word query is from the Buckeye dataset, and the two images are from the same image dataset but from any one of the image datasets.
We repeat this for the LibriSpeech queries in the \librispeech\ bars and the Flickr queries in the \flickr\ bars. 
It is important to note that for the \familiar\ setting, none of the familiar queries were sampled from the Buckeye dataset. 
Therefore, this bar is not in the \familiar\ and \familiarWithNovel\ tests.
Except for this, we see that the scores across all the model variants (\cref{fig:no_warm_start_image_discrepancy_test,fig:cpc_warm_start_image_discrepancy_test,fig:alexnet_warm_start_image_discrepancy_test,fig:alexnet_cpc_warm_start_image_discrepancy_test}) are relatively the same for the original (\overall), Buckeye (\buckeye), LibriSpeech (\librispeech) and Flickr (\flickr) tests.
We conclude that using different speech and image datasets does not result in dataset biases where the models are more accurate on instances from a particular dataset.
Concretely, the \model\ models exhibit a \ac{ME} bias due to how the novel and familiar classes are organised in its representation space.

\mysubsection{Further analysis of the learned representation space}{Further Analysis of the Learned Representation Space}
\label{chap_me:represetnation space}

 \begin{figure*}[!b]
 	 \centering
 	  \includegraphics[width=\textwidth]{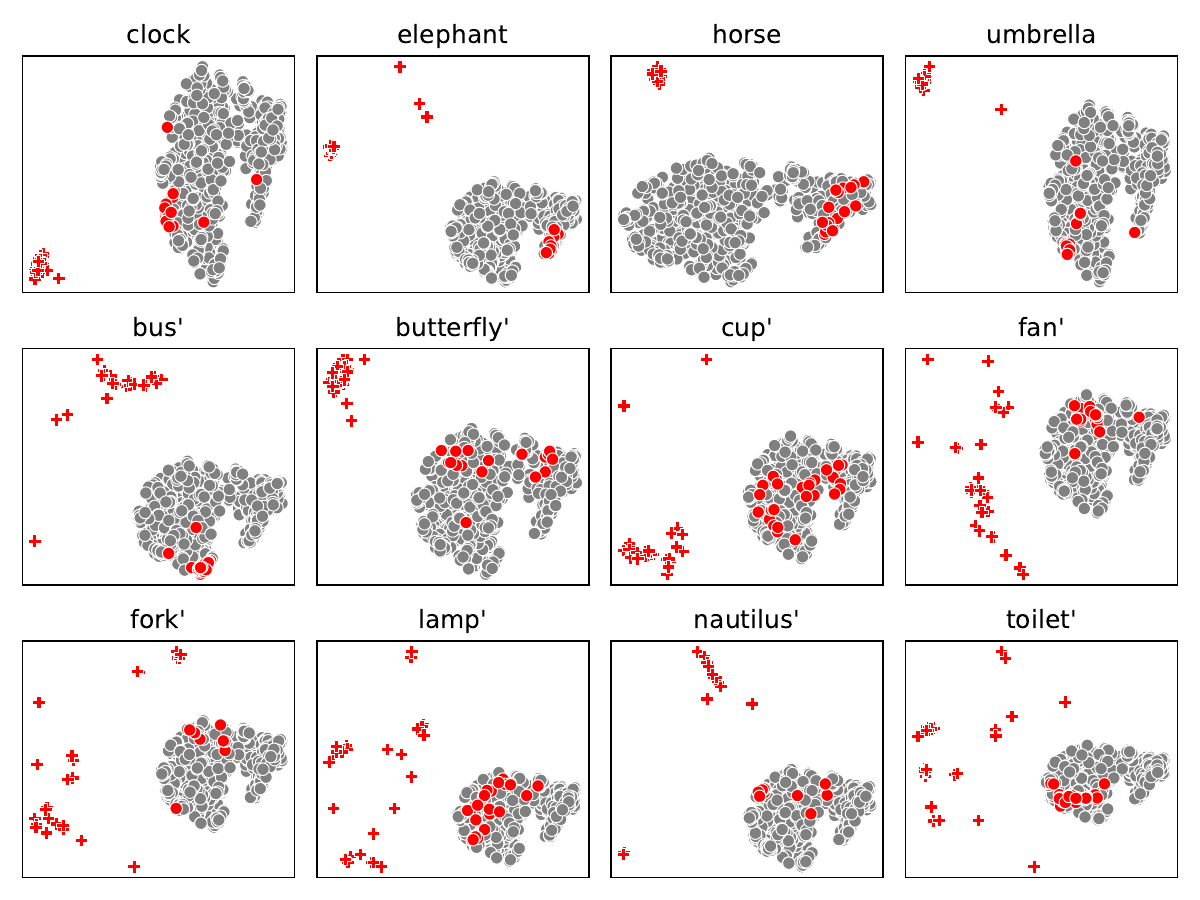}
 	   \caption{
 	   		Visualisations of the learned representation space of \model\ with self-supervised AlexNet and \ac{CPC} initialisations. 
 	   		The crosses indicate spoken word queries, and the circles indicate images. 
 	   		Red highlights the query class under consideration, and an apostrophe (’) indicates a novel class. 
 	   		For each plot, we only show the spoken words of the query class, but we plot the images from all the novel and familiar classes. 
    	}
 	   \label{fig:tsne-audio-and-image}
\end{figure*}

In Section~7.2 of \RQME, we found that the ME bias emerges due to the novel classes being closer to one another than to familiar classes. 
We also found that samples of a familiar class are grouped close together and the different familiar classes are placed far from one another.
However, there are exceptions to this, and we look into some classes that do not conform to this trend in Section~7.3 of \RQME.

In Figure~\ref{fig:tsne-audio-and-image}, we use \model\ with AlexNet and \ac{CPC} initialisations to plot the embeddings generated for word and image samples from some familiar and novel classes.
For each plot considering a specific class, we plot all the images of the familiar and novel classes but only the spoken words of the class under consideration.
The crosses indicate spoken words and the circles indicate images.
We highlight the word and image samples for the class under consideration in red and use an apostrophe to indicate a novel class.
To get these plots, we initially find that the learnt audio and vision modalities are isolated from one another. 
\cite{liang_mind_2022} found this phenomenon is common in \ac{VGS} models with a contrastive objective.
We align the two modalities by learning a linear transformation that minimises the Euclidean distance between speech-image embedding pairs, as proposed by \citet{udandarao_understanding_2022}.

From the plots in Figure~\ref{fig:tsne-audio-and-image}, we see that the word samples of the familiar classes are placed together, and the images of the familiar classes are also placed together. 
In some instances, images from other classes are closer to the spoken words of a familiar class than the class' corresponding images.
However, the spoken words of a familiar class are mostly closer to its corresponding images than those of other classes.
The novel spoken words and images are mostly scattered within the space.
When this model is prompted with a novel word query, most novel objects are closer to the novel word query than to a familiar object.
\Ie \model's representation space is mostly a visualisation of the \ac{ME} bias.
It is important to note that the strength of the \ac{ME} bias in the model and in children is not 100\%.
Therefore, we will see some instances where familiar images are closer to novel spoken words and even cases where familiar images are closer to spoken words of other familiar classes.

\mysection{Bug report}{Bug Report}

After the dissertation submission we found a small bug in the loading in of the \ac{CPC} weights, resulting in the weights not being loaded in successfully.
The result is that the audio branch was essentially randomly initialised.
However, this does not influence the main conclusion: our \ac{VGS} model shows a robust \ac{ME} bias.
The only conclusion it affects is the contribution of the pretrained \ac{CPC} weights to the strength of the \ac{ME} bias.

\mysection{Chapter summary}{Chapter Summary}

Our goal in this chapter was to establish whether a visually grounded word acquisition model, \model, exhibits a \ac{ME} bias.
We used this model since it is representative of most \ac{VGS} models in the literature.
We trained \model\ on spoken words and images from a set of familiar classes, after which we prompted the model with a novel spoken word query and asked whether the query belongs to an image of a familiar object or to an image of a novel object.

We considered \model\ with different initialisations that represent prior language and visual knowledge that children might have acquired when they start using the \ac{ME} bias.
All the \model\ versions exhibit the \ac{ME} bias, with the strongest bias occurring when both the visual AlexNet and acoustic \ac{CPC} initialisations are used.
We found that the visual AlexNet initialisation contributes the most to the strength of the bias regardless of whether we use a self-supervised or supervised AlexNet version.

With a deep dive into the performance of the \model\ variant with both AlexNet and \ac{CPC} initialisations, we found that the ME bias grows stronger earlier in model training and then stabilises over time. 
In additional tests, we showed that the ME bias could not be explained away by information leakage from the training data or biases towards some of the datasets we used to construct the large \ac{ME} dataset. 
By analysing the learned representation space, we found that novel classes and the various familiar classes are mostly distinguished from one another, resulting in the observed \ac{ME} bias.
We also considered how specific the results are to our model design and found that various losses all result in a \ac{ME} bias.
However, \model\ using an InfoNCE loss exhibited the strongest \ac{ME} bias.

\citet{byers-heinlein_monolingual_2009} and \citet{kalashnikova_effects_2015}, among others, took the original \ac{ME} studies on young children further by considering the effect multilingualism has on the bias. 
In the multilingual setting, words referring to the same object from different languages are acoustically different, thus violating the one-to-one mapping rule of the \ac{ME} bias.
These cognitive studies found that the \ac{ME} bias in bi- and trilingual children is weaker than in monolingual children.
In the next chapter, we consider the effect of multilingualism on the \ac{ME} bias of \model\ with the InfoNCE loss.

\mychapter{Mutual exclusivity in multilingual visually grounded speech models}{Mutual Exclusivity in Multilingual Visually Grounded Speech Models}{Research Question 4}
\label{chap:multilingual_me}

\noindent\begin{publicationinfo}{}
	\begin{center}
		{\large\ShortInTextTitle{Chapter~\ref{chap:multilingual_me} Specifics}}
	\end{center}
	
	\vspace{4pt}
	
	\researchqfourcentered
	
	\vspace{4pt}
	
	\researchpaper{Research Paper \MultilingualME}{\bibentry{nortje_using_2024}.}
	
	\vspace{4pt}
	
	\ShortInTextTitle{Project Website with Code Resources}: \href{https://sites.google.com/view/mutualexclusivityinvgs}{[Link]}
\end{publicationinfo}

In the previous chapter, we computationally studied the \acf{ME} bias in a setting that mimics the environment in which children use the bias.
This study was done in a monolingual English setting.
Our previous chapter, along with other computational  \ac{ME} studies~\citep[e.g.,][]{vong_cross-situational_2022} asks whether the ME bias observed in children~\citep[e.g.,][]{markman_childrens_1988}, can also be observed in machine learning models. 
Cognitive research has also explored how language experience, like multilingualism, influences the bias in children.
These studies considered children between the ages of 17 months to six years from various language backgrounds, \ie the number of languages they learn.

Various studies have compared the \ac{ME} bias in monolingual vs bilingual children.
\citet{kalashnikova_effects_2015}, \citet{davidson_monolingual_1997}, \citet{davidson_monolingual_2005} and \citet{houston-price_language_2010} found that monolingual children exhibited a stronger \ac{ME} bias than bilingual children.
Although this was the finding in most studies, \citet{rocha-hidalgo_patterns_2021} found no difference between \ac{ME} use in monolingual and bilingual children.

\citet{byers-heinlein_monolingual_2009} considered the \ac{ME} bias in children who learn more than two languages.
They tested whether 17- to 18-month-old children from a monolingual English background, and children from bilingual and trilingual backgrounds where English is one of their languages, use the ME bias. 
They found that monolingual children showed a strong use of the ME bias, while bilingual children showed minor usage, and trilingual children showed no usage.	

Several cognitive studies on the \ac{ME} phenomenon have explored how multilingualism affects children's bias. 
However, no computational study has considered this effect since some of the computational monolingual studies showed a ME bias, but the bias was not robust and consistent.
We are the first to computationally study the effect of multilingualism on the ME bias in \RQMultilingualME\ in this chapter.
To do this, we use the \model\ model that showed the strongest \ac{ME} bias in Chapter~\ref{chap:me}. 
We train \model\ in different multilingual settings and find that the monolingual model has a weaker \ac{ME} bias than its multilingual variants.

\mysection{Publication and contribution declaration: Research paper \MultilingualME}{Publication and Contribution Declaration: Research Paper \MultilingualME}
\label{multilingual_me_publication}

In \RQME, presented in Chapter~\ref{chap:me}, we trained a \ac{VGS} model, \model, on spoken English words and images from familiar classes.
We found that \model\ exhibited a \ac{ME} bias when prompted with a novel spoken English query and asked if the query matches an image of an unknown or familiar object.
In this chapter, we present \RQMultilingualME\ in which we train \model\ on different multilingual language combinations to test whether the \ac{ME} bias weakens as the model is trained to learn word representations from more languages.

\mysubsection{Contribution declaration}{Contribution Declaration}
\begin{table}[h]
	\caption{A layout of the authors’ contributions to the publication: \bibentry{nortje_using_2024}.}
	\label{tab:multilingual_me_contributions}
	\centering
	\renewcommand{\arraystretch}{1.2}
		\begin{tabularx}{\linewidth}{@{}lL@{}}
			\toprule
			Author &  Contributions \\
			\midrule
			\leanneName & The layout of the research question, the model architecture, the data setup, the implementation and the generation of the numerical results. \\
			\addlinespace
			\danName &   Assisted with model analysis. \\  %
			\addlinespace
			\yevgenName & The execution of the significance tests. \\ 
			\addlinespace
			\hermanName & The layout of the research question and an editorial role.\\
			\bottomrule
		\end{tabularx}
	\end{table} 

Table~\ref{tab:few-shot_contributions} lists the contributions of all the authors in \RQMultilingualME.
The majority of work was proposed and implemented by Leanne Nortje, the author of this dissertation, except as stated otherwise.

\mysubsection{Paper}{Paper}
\label{multilingual_me_paper}

\begin{figure}[!t]
	\centering
	\includegraphics[width=0.6\linewidth]{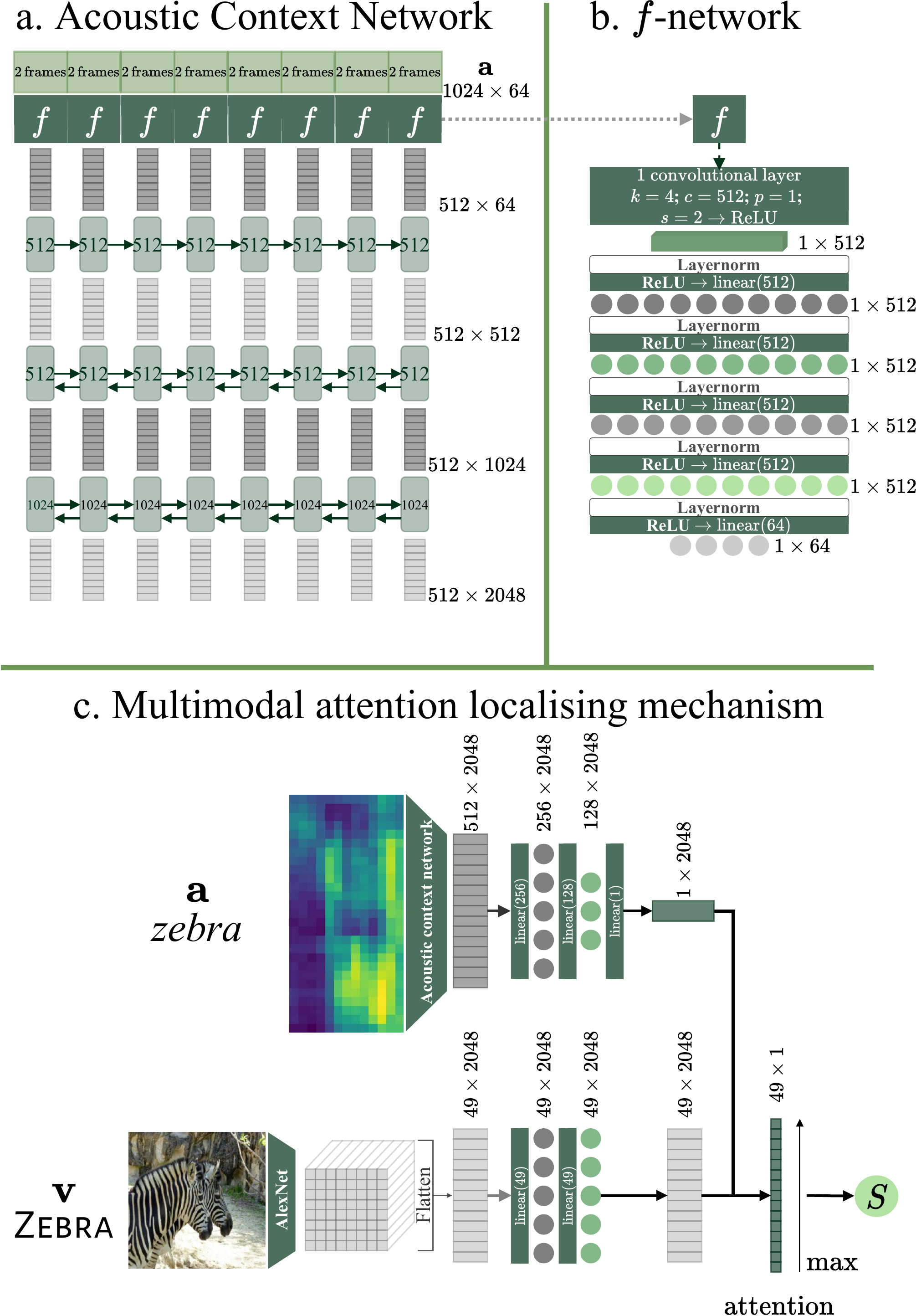}
	\caption{%
			(c) The \model\ model that exhibited the \ac{ME} bias in Chapter~\ref{chap:me}, consists of a vision network and an audio network (a+b) connected with a word-to-image attention mechanism. The mechanism outputs a similarity score for a speech and image input.%
	}
	\label{fig:me_multilingual_model}
\end{figure}

We present \RQMultilingualME\ in this section and use the same notation as \RQME\ in the previous chapter as shown in Table~\ref{tab:notation}, which is different than the notation used in the rest of this dissertation.
Figure~\ref{fig:me_multilingual_model} shows the entire architecture of the model we propose in the following publication since it provides more information on the architecture of the audio branch.

\newpage
\includepdf[pages={1-},scale=1.0]{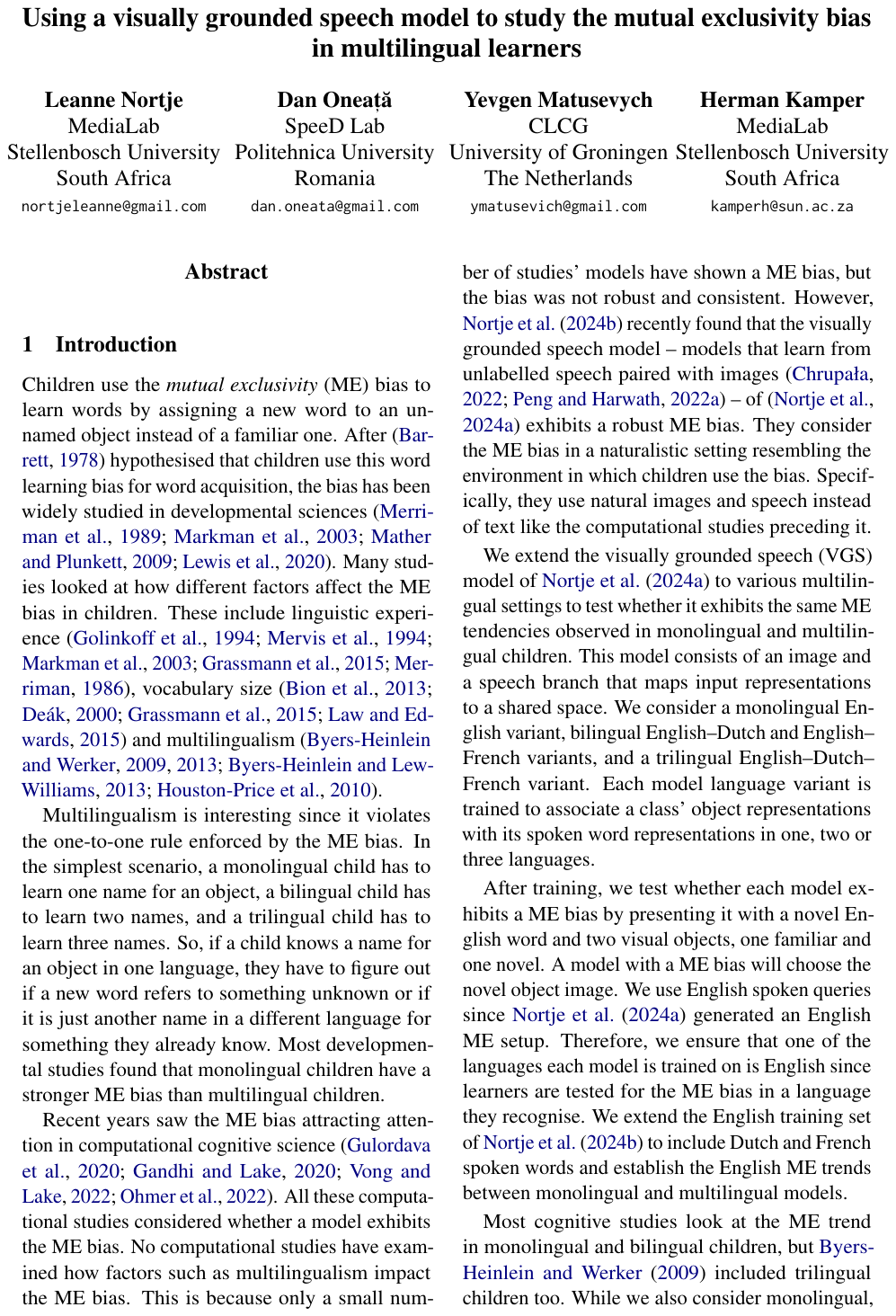} 

\graphicspath{{multilingual_mutual_exclusivity/figures/}}
\mysection{Multilingual mutual exclusivity dataset}{Multilingual Mutual Exclusivity Dataset}

\begin{table}[tb]
	\small
	\centering
	\renewcommand{\arraystretch}{1.2}
	\caption{The English, Dutch and French translations of the familiar and novel classes used in the multilingual ME test setup.}
	\begin{tabularx}{0.6\linewidth}{@{}ccccc@{}}
		\toprule
		\multicolumn{3}{c}{\textbf{Familiar classes}} &  & \multicolumn{1}{c}{\textbf{Novel classes}} \\
		\cline{1-3} \cline{5-5}
		English & Dutch & French &  & English\\
		\midrule
		bear & beer & ours &  & ball\\
		bird & vogel & oiseau &  & cake\\
		boat & boot & bateau &  & trumpet\\ 
		car & auto & voiture &  & lamp\\
		cat & kat & chat &  & bench\\
		clock & klok & horloge &  & cannon\\
		cow & koe & vache &  & piano\\
		dog & hond & chien &  & barrel\\ 
		elephant & olifant & éléphant &  & camera\\
		horse & paard & cheval &  & fan\\
		scissors & schaar & ciseaux &  & nautilus\\
		sheep & schaap & mouton &  & revolver\\
		umbrella & paraplu & parapluie &  & buck\\
		&  &  &  & toilet\\
		&  &  &  & butterfly\\
		&  &  &  & cup\\
		&  &  &  & guitar\\
		&  &  &  & fork\\
		&  &  &  & bus\\
		&  &  &  & chair\\
		\bottomrule
	\end{tabularx}
	\label{tab:multilingual_classes}
\end{table}

Our goal in \RQMultilingualME\ was to determine whether \model\ trained in monolingual and multilingual settings resemble the \ac{ME} trend observed in children.
To do this, we generated a multimodal dataset consisting of natural images paired with isolated spoken words from various languages. 
In Chapter~\ref{chap:me}, we combined multiple image and English speech datasets to get one large dataset.
The training set consists of spoken English words and natural images of entire scenes for a set of familiar classes. 
The test set consists of spoken English words and images containing isolated natural objects for the novel and familiar classes. 
The development set consists of spoken English words and isolated natural object images for only the familiar classes. 
We extend this dataset by adding Dutch and French isolated spoken words to the training English word-image pairs. 
This section gives more details on how we extended the English \ac{ME} dataset of Chapter~\ref{chap:me}.

Table~\ref{tab:multilingual_classes} shows the 13 English familiar classes from Chapter~\ref{chap:me} we also use in this chapter, and the Dutch and French translations for these classes.
To get spoken Dutch words for the familiar classes, we use the Dutch subset of multilingual LibriSpeech~\citep{pratap_mls_2020}, the \ac{CGN}~\citep{oostdijk_spoken_2000} and the Dutch subset of the Common Voice~\citep{ardila_common_2020} datasets.
The Dutch LibriSpeech set contains 2.25k hours of speech data from 206 LibriVox books and 91 speakers.
The \ac{CGN} dataset contains 900 hours of speech containing around 9 million words, whereas the Dutch Common Voice set contains 23 hours of speech from 502 speakers.

To get spoken French words for the familiar classes, we use the French subset of the multilingual LibriSpeech~\citep{pratap_mls_2020} corpus and the French subset of the Common Voice~\citep{ardila_common_2020} corpus.
The French LibriSpeech set contains 1.33k hours of speech data from 224 LibriVox books and 114 speakers, whereas the French Common Voice set contains 184 hours of speech from 3\,005 speakers.
All these Dutch and French speech datasets contain forced alignments, which we use to isolate spoken words.

\begin{figure}[tb]
	\centering
	\includegraphics[width=\linewidth]{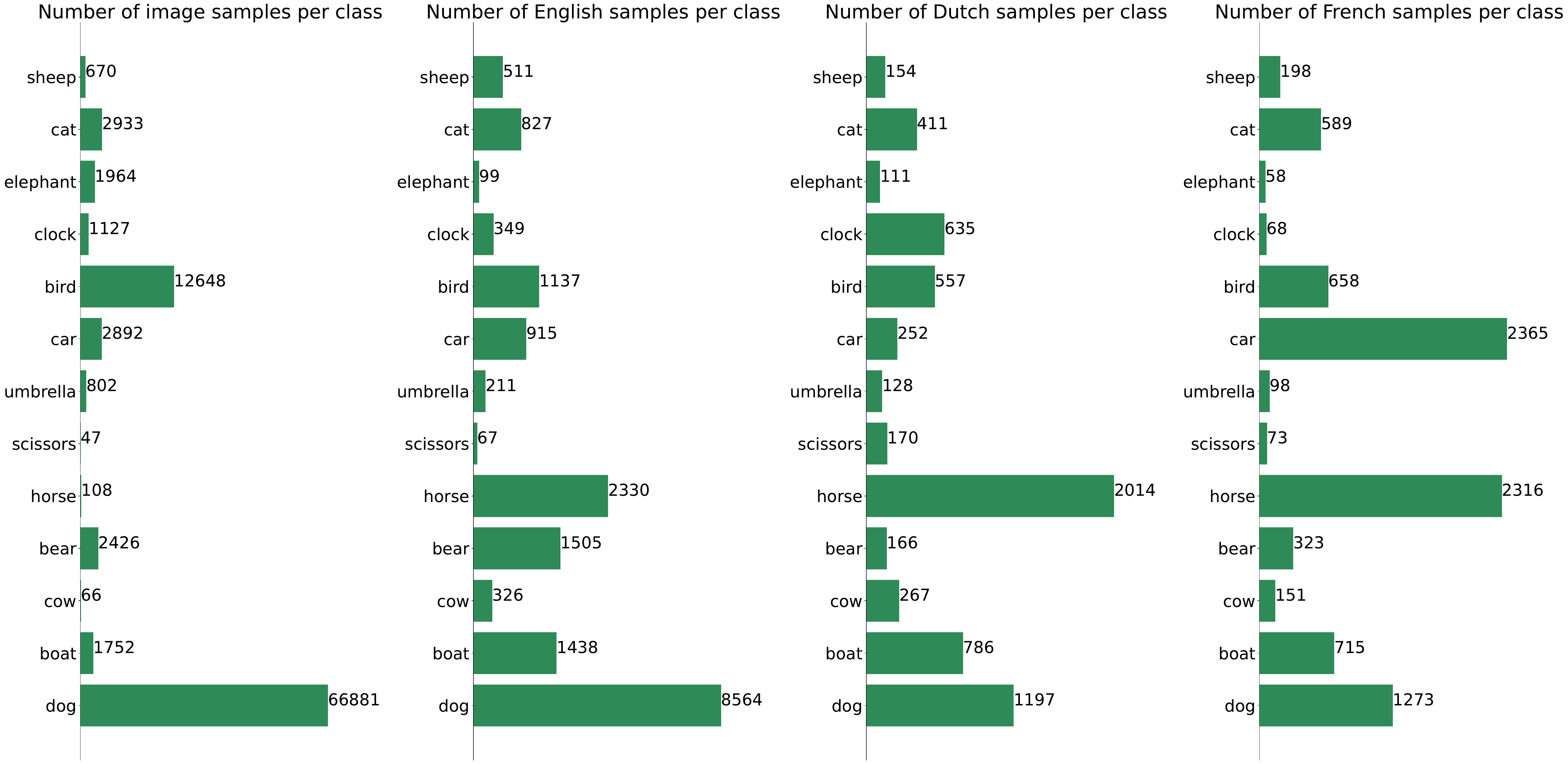}
	\caption{%
		The amount of English, Dutch and French spoken words and images for each familiar class in the speech-image \ac{ME} dataset we construct. 
	}
	\label{fig:multilingual_me_dataset_analysis}
\end{figure}

This results in a training set that contains natural images of entire scenes paired with spoken English, Dutch and French words for 13 familiar classes. 
Specifically, the training set contains 18\,279 unique spoken English word segments, 6\,848 unique spoken Dutch word segments, 8\,885 unique spoken French word segments and 94\,316 unique unsegmented natural images spanning the 13 familiar classes. 
Figure~\ref{fig:multilingual_me_dataset_analysis} shows the amount of English, Dutch, French and image samples per class.
To evaluate whether a model exhibits a ME bias, we use the test set containing English spoken words paired with isolated natural objects for both the 20 novel classes and the 13 familiar classes.
The test set comprises 8\,575 unique spoken English words with 22\,062 unique segmented natural object images.
The development set contains 10 unique spoken English words and 10 segmented natural object images for each of the 13 familiar classes.
The test, development and training samples for the familiar classes do not overlap.

\graphicspath{{multilingual_mutual_exclusivity/figures/}}
\mysection{Further Analysis}{Further Analysis}

In \RQMultilingualME, we considered the strength of the ME bias exhibited by various \model\ language variants.
We found that the monolingual and multilingual \model\ models show the opposite trend observed in monolingual and multilingual children. 
More specifically, the monolingual \model\ exhibited a weaker \ac{ME} bias than the multilingual \model\ models.
In contrast, monolingual children showed a stronger \ac{ME} bias than multilingual children.
We also performed various experiments to ensure that the ME trend we observe is stable.
The conclusion of \RQMultilingualME\ was that the cognitive plausibility of various design choices should be considered. 
This section considers one of these design choices we recommended for future work.

To recap, one cognitively plausible design choice we proposed is the impact of training data size. 
What will happen if we lessen the amount of data the trilingual model is trained on to the amount of data the monolingual English model sees? This relates to the question of whether trilingual children encounter more speech than monolingual children. Or do they encounter the same amount of speech?

What is the effect of the specific language combinations we choose to train our models on? 
If two languages have similar sounding words, like \word{elephant} in English and \word{éléphant} in French, then the one-to-one rule is still enforced. 
Then, using very distinct languages may lead to a lower ME bias than using more similar languages.
Moreover, is this also the case with children?
\Ie do children who learn various similar languages have a stronger \ac{ME} bias than children learning various distinct languages?

\newcommand{\ab}{\mathbf{a}_1}
\newcommand{\vb}{\mathbf{a}_2}
\newcommand{\el}{\mathbf{a}_{\textrm{English}}}
\newcommand{\dl}{\mathbf{a}_{\textrm{Dutch}}}
\newcommand{\fl}{\mathbf{a}_{\textrm{French}}}
Multilingual children are, in some instances, explicitly taught that the English word for a \image{boat} object is \word{boat} and the French word is \word{bateau}.
Therefore, do we consider this and add cross-lingual loss terms to our loss function?
As a preliminary study, we add cross-lingual terms to the loss function of the bilingual and trilingual models in Table~\ref{tab:me}. 
More specifically, we add a cross-lingual InfoNCE loss term: 
\begin{equation}
	\begin{aligned}
		\ell(\ab, \vb) = & \log \left(\frac{\exp S(\ab, \vb)}{\exp S(\ab, \vb) + \sum_{i=1}^{N_\textrm{neg}}{\exp S(\ab, \vb^{i-})}}\right) \\
		&+ \log \left(\frac{\exp S(\ab, \vb)}{\exp S(\ab, \vb) + \sum_{i=1}^{N_\textrm{neg}}{\exp S(\ab^{i-}, \vb)}}\right)
	\end{aligned}
	\label{eq:infonce}
\end{equation}
where $\ab$ are spoken words from one language and $\vb$ are the spoken words from another language.
For the trilingual model, we compare English and Dutch, Dutch and French, and French and English.
Therefore, for the trilingual model, we add the following terms to the loss in \RQMultilingualME: $\ell(\el, \dl)$, $\ell(\el, \fl)$ and $\ell(\fl, \dl)$. 
For the bilingual English--Dutch model, we add $\ell(\el, \dl)$, and for the English--French model, we add $\ell(\el, \fl)$.

The scores in Table~\ref{tab:me} show that the trend we previously observed in \RQMultilingualME\ changes.
The \ac{ME} or \me\ scores in Table~\ref{tab:me} show that the monolingual \model\ has a stronger ME bias than the bilingual \model\ models. 
However, the trilingual model still has the strongest \ac{ME} bias. 
The monolingual model is the only model not trained with language-language comparisons. 
Therefore, it might also be necessary to consider adding within-language loss terms. 

 \begin{table*}[tb]
	\small
	\caption{
		The scores for the monolingual, bilingual and trilingual \model\ models when language-language comparisons are added to the bilingual and trilingual models' loss functions. }
	\label{tab:me}
	\centering
	\renewcommand{\arraystretch}{1.2}
	\begin{tabularx}{\linewidth}{@{}lCCCCC@{}}
		\toprule
		&\multicolumn{4}{c}{Accuracy (in \%)}\\
		\cmidrule(lr){2-6}
		& \Familiar & \Me  & \MeOther & \FamiliarWithNovel & \Novel \\
		\midrule
		English & 93.16 & 63.91 & 65.48 & 92.34 & 49.21\\
		English-Dutch & 87.28 & 61.21 & 59.19 & 85.03 & 50.44\\
		English-French & 88.40 & 61.84 & 59.88 & 87.59 & 50.35\\
		English-Dutch-French & 88.00 & 67.69 & 68.97 & 85.36 & 49.07\\
		\bottomrule
	\end{tabularx}
\end{table*}

Nevertheless, we perform the \familiar, \meOther, \familiarWithNovel\ and \novel\ tests to validate our results.
To verify the \ac{ME} trend we observe in Table~\ref{tab:me}, we perform \meOther\ in which we prompt the model with a novel spoken English query and ask if the query corresponds to a familiar object or a novel object from a different novel class than the query.
We observe the same \ac{ME} trend in the \meOther\ scores: the monolingual model's bias is weaker than the trilingual model's and stronger than the bilingual models'.

We move onto the \novel\ test, where the two images a model has to choose from are from two different novel classes. 
The model should randomly pick an image when prompted with a novel query from the same class as one of the images.
The \novel\ scores in Table~\ref{tab:me} are around 50\%, which shows that the models randomly pick an image.
Since the models perform as expected on these two tests, we ensure that information leakage from the novel classes did not occur.  
Therefore, these tests could not explain why the \ac{ME} trend changed from the trend in \RQMultilingualME, and why the monolingual model has a weaker bias than the trilingual model but a stronger bias than the bilingual models.

Turning to the \familiar\ scores, we see that the bilingual English-Dutch model has a lower score than the trilingual English-Dutch-French model, and the bilingual English-French model has only a slightly higher score than the trilingual model.
We see a similar trend in the \familiarWithNovel\ scores, where the model is prompted to match a familiar spoken English word to its familiar object from two images: a novel image and the familiar image.
The trilingual model loss has three language-language comparisons vs the one language-language comparison in the bilingual models' loss. 
This could result in the trilingual model being forced to learn a representation space that better separates the familiar classes from one another and from the novel classes. 
If this was the case, the monolingual English model should have a weaker \ac{ME} bias than the bilingual models since it has no language-language comparisons in its loss. 
We conclude that the cognitive plausibility of various language-language comparisons in the model objective should be thoroughly investigated.
Additionally, the \familiar\ tests should also be conducted in Dutch for the English-Dutch and English-Dutch-French models, and in French for the English-French and English-Dutch-French models.

Future work should examine the effect of various language-language comparisons on the \ac{ME} bias and the resulting representation spaces. 
Nevertheless, the preliminary study into monolingual vs multilingual ME trends presented in this chapter, is a start to determine whether a \ac{VGS} model exhibit the same trends seen in children from various language backgrounds.

\mysection{Chapter summary}{Chapter Summary}

The \acf{ME} bias children use to learn new words states that a novel word belongs to an unknown object image rather than a familiar object. 
We aim to establish whether \model, proven to exhibit the \ac{ME} bias, can emulate the \ac{ME} bias trends seen in multilingual children~\citep{byers-heinlein_monolingual_2009}. 
We train \model\ variants on familiar class images paired with spoken words from English, or English and Dutch, or English and French, or English, Dutch and French. 
After training, we ask each model variant to which image a novel spoken English query belongs to. 

We concluded that, unlike children, the monolingual English \model\ shows a weaker \ac{ME} bias than the multilingual variants.
Furthermore, we concluded that various cognitive plausible design choices should be considered.
We performed preliminary tests to investigate the effect on the \ac{ME} bias if we add cross-lingual comparisons to the model objective.
However, the results were inconclusive.
Future work should consider the effect of training data size and the languages we use in our setup.
Moreover, the addition of cross-lingual and within-language comparisons should be investigated in more depth.

\acresetall
\graphicspath{{conclusion/fig/}}

\mychapter{Summary and conclusions}{Summary and Conclusions}{}
\label{chap:conclusion}

\Ac{VGS} are inspired by how children quickly and efficiently learn the word and object representations of new classes without transcriptions. 
As a result, \ac{VGS} models are suitable for developing speech applications for low-resource settings where textual transcriptions are limited or unavailable. 
Since \ac{VGS} models are inspired by how children learn words and their visual referents, these models are also ideal for computationally studying cognitive hypotheses about how children learn words.

In the first part of this dissertation, we investigated using \ac{VGS} models to solve two low-resource tasks: \acf{VPKL} and visually grounded few-shot learning of spoken words.
In the \ac{VPKL} task, a query keyword depicted by an image should be detected and localised in a spoken utterance.
The visually grounded few-shot word acquisition model should learn how a word class is represented in the speech and vision domains from only a few ground truth speech and image examples. 

In the second part of this dissertation, we used our few-shot model architecture to computationally study the \ac{ME} bias, where a novel word should belong to an unknown object rather than a familiar one.
Concretely, we trained the model on ground truth spoken word and image pairs from familiar classes. 
After we found that this model shows a robust \ac{ME} bias when applying it to novel samples, we investigated the effect of multilingualism on the model’s \ac{ME} bias.

In this chapter, we summarise the conclusions of each chapter and how these conclusions answer our research questions (Section~\ref{sec:conclusions_and_research_questions}).
Section~\ref{sec:conclusion_contribution} states our contributions to \ac{VGS} modelling and to the development of low-resource applications and cognitive modelling.
In Section~\ref{sec:future_work}, we discuss possible avenues for future work. 
We end off the chapter with Section~\ref{sec:dissertation_conclusion}, which concludes the dissertation and relays the key findings that the reader should take away. 

\mysection{Conclusions and the answers to our research questions}{ Conclusions and Answers to Our Research Questions}
\label{sec:conclusions_and_research_questions}

Throughout this dissertation, we used the same general \ac{VGS} model consisting of an acoustic and a vision network connected with a multimodal attention mechanism. 
This mechanism outputs a similarity score for a speech and image input.
Based on the research question we consider, we adapted the attention mechanism and the audio and vision networks to output either a single embedding or a sequence of embeddings.
This section summarises and discusses the conclusions we drew from each chapter.
First, we look at \rqone\ and \rqtwo\ relating to low-resource language systems.
Thereafter, we discuss \rqthree\ and \rqfour\ relating to computational modelling of the \ac{ME} bias.

\mysubsection{Visually grounded speech modelling for low-resource languages}{Visually Grounded Speech Modelling for Low-Resource Languages}

Our first two research questions asked whether images can be used to weakly supervise speech models for low-resource languages.
To answer \rqone\ in Chapter~\ref{chap:localisation}, we aimed to use \ac{VGS} models to do keyword localisation. 
For \rqtwo\ in Chapter~\ref{chap:few-shot}, we aimed to use \ac{VGS} models for few-shot word learning.

\vspace{4ex}
\noindent\begin{researchquestions}
	
	{\textbf \researchqone}
	
\end{researchquestions}

Various studies have shown that \ac{VGS} models can be used for syllable segmentation~\citep{peng_syllable_2023}, word segmentation~\citep{peng_word_2022}, few-shot word acquisition~\citep{miller_tyler_exploring_2022} and keyword localisation~\citep{olaleye_keyword_2022}.
Although \citet{olaleye_visually_2023} showed that a \ac{VGS} model can be used for keyword localisation in a low-resource language, their approach still requires written English keyword queries to search through low-resource speech datasets. 
More specifically, in \rqone, we proposed a new keyword localisation task called \acf{VPKL}: a given query keyword depicted by an image should be detected in spoken utterances. 
For each utterance where the keyword was detected, the keyword should be located within the utterance. 
Moreover, we set out to get a keyword localisation system for a low-resource language, \yoruba, that does not require textual English queries. %

To do this, we developed our approach by using English as an artificial low-resource language. 
Firstly, we proposed adapting our general \ac{VGS} model to output a sequence of embeddings for both the input image and speech utterance. 
These output sequences are then fed to an attention matchmap in which the similarity between each frame embedding and each pixel embedding is calculated.
We obtained an image and a speech context vector from this attention mechanism to calculate a similarity score between the input spoken word and image.
For training, we initially used ground truth transcriptions to obtain training pairs. 
Each pair contained an image and a spoken word corresponding to a keyword. 
Although this ground truth approach outperformed the visual \ac{BOW} model of \citet{olaleye_attention-based_2021} that localises textual English queries in speech, using unsupervised visual tagger targets on our approach performed poorly.

To get an approach that did not rely on ground truth transcriptions and could work in low-resource settings, we proposed using a simpler \ac{VGS} model and few-shot learning.
Concretely, we proposed another adaptation of our general VGS model that still outputs a sequence of embeddings for input speech and images. 
However, from here, the outputs were fed to a simpler attention mechanism that excludes the context vectors of the previous model; the similarity score for a given speech and image input was directly calculated from the attention matchmap.
Together with this, we also proposed a few-shot pair mining approach to get training pairs.
For this pair mining approach, we used a few spoken word examples for each keyword we wanted to localise, called a support set.
Using this set, we mined spoken utterance and image pairs for each keyword class by comparing a keyword’s word examples with the utterances in speech-image pairs from a large dataset.
This few-shot approach led to the best performing English \ac{VPKL} model.

Since the English \ac{VPKL} approach proved successful, we applied our few-shot method to a low-resource language, \yoruba.
The ground truth \yoruba\ model showed large performance gains over a \yoruba\ visual \ac{BOW} model~\citep{olaleye_yfacc_2023}, especially for localisation.
However, the precision of the \yoruba\ few-shot model is low since the mining scheme used an English model to do speech comparisons. 
This led us to believe that improving this mining approach could lead to a valuable \yoruba\ keyword localisation system.

Taking all this together to answer \rqone, we can conclude that \ac{VGS} modelling can be used to detect and locate visual keywords in speech of a low-resource language.

\vspace{4ex}
\noindent\begin{researchquestions}
	
	{\textbf \researchqtwo}
	
\end{researchquestions}

The few-shot mining methods we employed in the \ac{VPKL} approach came from the advances we made in answering \rqtwo: can we learn the visual and spoken representations of a word class from only a few isolated spoken words and image pairs?
To do this, we proposed adapting our general \ac{VGS} model so that the audio network outputs a single embedding for an input spoken word and a sequence of embeddings for an input image. 
We map an image into a sequence of embeddings since we want the attention mechanism to learn which part of the image the spoken word corresponds to. 
This model's attention mechanism calculated the similarity between the word embedding and each embedding in the sequence of embeddings generated for the image.

To mine training pairs, we are given a large image dataset, a large speech dataset and a set containing a few ground truth spoken words paired with natural images for each class. 
This set is called the support set. 
More specifically, for mining the spoken word pairs, we used a query-by-example speech system to predict in which utterances from the large unlabelled speech dataset the spoken word examples given in the support set occur. 
With this system’s predicted word boundaries, we isolated possible word matches. 
\Eg given a class \textLabel{zebra}, we match the spoken \word{zebra} words in the support set to unlabelled spoken utterances and isolate the speech segments predicted to match \word{zebra} words. 
For the image part of the pair mining scheme, we used pretrained embeddings to predict which images in the large unlabelled set matched the support set images of a few-shot class. 

Our proposed approach outperformed an existing approach when using fewer ground truth examples per class. 
As a result, we also applied our approach to a low-resource language, \yoruba, which outperformed an English variant and set a new competitive baseline. 
We found that for classes that occupy small parts of images, the model retains contextual information and associates the visual context of a class with the spoken word rather than the class’ object.

In summary, we found that the answer to \rqtwo\ was that a \ac{VGS} model can learn the spoken word and object representations of a class from a small number of examples. 

\mysubsection{Computational modelling of the mutual exclusivity bias}{Computational Modelling of the Mutual Exclusivity Bias}

The \ac{VGS} models considered in the first part of this dissertation and most \ac{VGS}  models, in general, are inspired by how children learn words and the objects they refer to. 
However, word learning constraints children use, such as the \ac{ME} bias, have not been computationally studied using \ac{VGS} models. 
In the last two research questions, we computationally studied the \ac{ME} bias using \ac{VGS}  models. 
This bias states that a novel word should belong to an unknown object rather than a familiar one since the familiar object already has a name.

\vspace{4ex}
\noindent\begin{researchquestions}
	
	{\textbf \researchqthree}
	
\end{researchquestions}

\rqthree\ asked whether a typical \ac{VGS} model can be used to computationally study a word learning constraint, the \ac{ME} bias, used by children.
 Before getting to a model application, we generated a speech-image \ac{ME} dataset using various existing speech and image datasets. 
 The resulting dataset consisted of spoken words paired with natural images. 
 The training set contains spoken words paired with natural images for familiar classes, whereas the development set contains spoken words paired with images of isolated natural objects for the same familiar classes. 
 The test set, however, contains spoken words paired with images of isolated natural objects for familiar and novel classes. 

We trained the model architecture we proposed in the few-shot study of \rqtwo\ on spoken words and images from the familiar classes. 
After training, we tested whether the model places a novel spoken word closer to an image depicting a novel object than to a familiar one. 
We found that all the models we considered exhibited a robust \ac{ME} bias, but the model with the strongest bias used visual AlexNet and audio \ac{CPC} initialisations. 
We also ensured that what we were observing was indeed the \ac{ME} bias by doing various experiments and analysing the learned representation space. 
In addition, we found that the vision AlexNet initialisation contributes the most to the strength of the \ac{ME} bias, regardless of whether the initialisation weights are obtained in a supervised or self-supervised manner. 
Lastly, we investigated how specific our model design is to the occurrence of the \ac{ME} bias and found that all losses we considered led to the ME bias. 
Using the InfoNCE loss resulted in the strongest ME bias.

The answer to \rqthree\ was that a \ac{VGS} model exhibits the \ac{ME} bias observed in children.

\vspace{4ex}
\noindent\begin{researchquestions}
	
	{\textbf \researchqfour}
	
\end{researchquestions}

In Chapter~\ref{chap:me}, we found that the \ac{VGS} model we proposed in Chapter~\ref{chap:few-shot} exhibited the \ac{ME} bias.
We also found that using the InfoNCE loss together with visual and audio initialisations, resulted in the strongest \ac{ME} bias. 
In \rqfour, we set out to establish the effect of multilingualism on the bias of this model. 
We specifically wanted to know whether the model is affected similarly to children~\citep{byers-heinlein_monolingual_2009}. 
To do this, we expanded the training set of the English \ac{ME} dataset we generated in Chapter~\ref{chap:me}, to also contain Dutch and French spoken words for the familiar classes. 

We trained various language variants of the model on the familiar classes’ images paired with spoken words from the languages under consideration. 
After training the model on this set of familiar classes, we tested all the models on English novel classes to measure the strength of each model’s \ac{ME} bias. 
We concluded that, unlike children, the monolingual English \model\ shows a weaker \ac{ME} bias than the multilingual variants. 
However, when adding language-language comparisons to the model objective, the bilingual models had the weakest bias, and the trilingual models had the strongest bias. 
Nevertheless, this preliminary study showed the importance of considering cognitively plausible design choices. 

The preliminary study we conducted in Chapter~\ref{chap:multilingual_me} partially answered \rqfour.
Specifically, we found that multilingualism does not affect the \ac{ME} bias in the specific \ac{VGS} model we considered, similarly than in children.

\mysection{Research contributions}{Research Contributions}
\label{sec:conclusion_contribution}
\renewcommand{\spacesize}{18pt}

In this dissertation, we asked four research questions regarding \ac{VGS} models. 
The first two questions related to using \ac{VGS} models for speech applications in low-resource languages.
The last two research questions related to using \ac{VGS} modelling to study the \ac{ME} bias observed in children and how a variable like multilingualism affects the bias.
In our attempt to answer these questions, this dissertation resulted in the following contributions.

\setcounter{contributionCount}{1} 
\vspace{\spacesize}
\noindent\researchContributionOne

\vspace{\spacesize}
\noindent\researchContributionTwo

\vspace{\spacesize}
\noindent\researchContributionThree

\vspace{\spacesize}
\noindent\researchContributionFour

\vspace{\spacesize}
\noindent\researchContributionFive

\vspace{\spacesize}
\noindent\researchContributionSix

\vspace{\spacesize}
\noindent\researchContributionSeven

\vspace{\spacesize}
\noindent\researchContributionEight

\vspace{\spacesize}
\noindent\researchContributionNine

\mysection{Future work}{Future Work}
\label{sec:future_work}

In this section, we discuss the possible avenues that can be pursued from the work we have done in this dissertation. 
We also briefly touch on more general avenues to pursue that are parallel to our work.

\mysubsection{Visually grounded speech modelling for low-resource languages}{Visually Grounded Speech Modelling for Low-Resource Languages}

In general, using \ac{VGS} modelling for low-resource speech applications should be extended to include more tasks. 
Such extensions could include spoken language modelling like \citet{peng_fast-slow_2022},  syllable segmentation like \citet{peng_syllable_2023}, word segmentation like \citet{peng_word_2022} or semantic speech retrieval like \citet{kamper_semantic_2019}.

\mysubsubsection{\Aca{VPKL}}{\Acl{VPKL}}

In order to improve our approach to obtain a smoothly functioning \ac{VPKL} system for low-resource languages, a few avenues can be pursued.
One avenue is to overcome the need for thresholds during detection, \eg a model output that can be rounded up to one if a keyword is present or down to zero if the keyword is not present. 
The final model we proposed could also be used to detect and localise spoken keywords.
In addition, this model could be used to detect and localise semantically relevant words to a query image. 
Overall, there is still much room for improvement on the \yoruba\ detection and localisation scores.

A more important limitation of our approach should be addressed first: we rely on a few-shot support set containing spoken word examples for the keywords we want to search for so that our approach can be applied to a low-resource setting.
However, this means that the vocabulary is constrained.
Future work should look at removing the support set, which would result in a model that can search for arbitrary words.

\mysubsubsection{Few-shot word acquisition}{Few-Shot Word Acquisition}

To get to a few-shot word learning approach that is feasible for low-resource languages, future work should look into extending this approach to more classes. 
Initial experiments showed that performance drops notably when extending the number of classes from five to 40.
However, the first step should be to improve the image matching step in the mining process since analysis on both English and \yoruba\ revealed that this step might limit performance. 
Since there are large amounts of image resources and vision is universal, this few-shot approach can take advantage of recent advances in computer vision.

\mysubsection{Computational cognitive modelling}{Computational Cognitive Modelling}

In this dissertation, we asked whether we can use \ac{VGS} modelling to study a word learning constraint called the \acf{ME} bias observed in children.
The \ac{ME} bias is a word learning constraint in which novel words are mapped to novel objects instead of familiar objects. 
Future work should consider using \ac{VGS} models to computationally study other learning constraints children use to learn words and the objects they refer to.
One such constraint is `the whole object assumption', which states that a novel word or term could refer to an entire object or a part of an object~\citep{markman_constraints_1990}. 
\Eg an object’s colour, shape or weight.
This constraint states that children initially constrain the meanings of new words to honour the \ac{ME} bias, whereafter they learn a novel word can also refer to an attribute of the object.
Next, we discuss future work that can be done on the \ac{ME} bias and the variables that might affect it.

\mysubsubsection{Mutual exclusivity}{Mutual Exclusivity}

We considered using \ac{VGS} modelling to computationally study the \ac{ME} bias in a realistic setting that resembles the environment in which children learn their native language. 
We trained a \ac{VGS} model on natural images and spoken words from a set of familiar classes.
However, future work should consider using entire utterances instead of isolated words paired with natural images. 
I.e. use standard \ac{VGS} training.
At test time, isolated words and objects should still be used. 
This would be an even more naturalistic version of our study and the study of \citet{vong_cross-situational_2022}. 
\citet{vong_cross-situational_2022} show that a model gets more confused when trained on unaligned data where it is unclear which word refers to which object in a scene.

\mysubsubsection{Effect of multilingualism on the mutual exclusivity bias}{Effects of Multilingualism on the Mutual Exclusivity Bias}

After proving that a \ac{VGS} word learning model exhibits a \ac{ME} bias similar to children, we investigated whether multilingualism has the same effect on the model’s bias as it has on children’s. 
A future avenue to pursue is to computationally study the effects of other variables on the ME bias and compare it to the effects observed in children. 
Such variables include linguistic experience ~\citep{golinkoff_early_1994, mervis_two-year-olds_1994, markman_use_2003, grassmann_childrens_2015, merriman_reasons_1986} like the amount of speech a child is exposed to, and vocabulary size~\citep{bion_fast_2013, deak_hunting_2000, grassmann_childrens_2015, law_effects_2015}.
However, before that, our preliminary study has to be extended to determine whether multilingualism affects the bias in VGS models and children similarly, along with assessing the cognitive plausibility of various \ac{VGS} modelling design choices.

\mysection{Dissertation conclusion}{Dissertation Conclusion}
\label{sec:dissertation_conclusion}

This dissertation aimed to study \ac{VGS} modelling with regard to developing speech applications for low-resource languages and computationally studying a word learning constraint, the \ac{ME} bias, observed in children. 
For low-resource language applications, we proposed a new keyword localisation task that is appropriate for low-resource languages. 
We proposed two new localisation models that use different attention mechanisms to do this. For the second model, we also proposed a new query-by-example based pair mining scheme using a few ground truth spoken words. 
We developed our approach using English as an artificial low-resource language and found that our best performing model outperforms a visual \ac{BOW} model that detects and localises written keywords. 
Finally, we applied our best approach to an actual low-resource language, \yoruba, and found that the mining approach required a query-by-example model more fine-tuned on \yoruba. Notably, the ground truth model showed promise that we can get a model for a low-resource language capable of doing \ac{VPKL} if we can improve the pair mining scheme.

Using a similar pair mining scheme with the same query-by-example system, we mined training pairs for our few-shot word acquisition model. 
Here, however, we use a support set containing a few ground truth spoken words paired with images and not just a set of speech examples. 
We also proposed a new multimodal attention model to learn the spoken word and visual representations of classes. 
After this approach outperformed an existing approach with fewer examples per class in English, we also applied the approach to a low-resource language, \yoruba. 
With this \yoruba\ model, we set a new and competitive baseline. Overall, we found that the model’s biggest flaw is that it uses contextual data when the visual depictions of a class are typically small.

On the cognitive side, we proposed a realistic setup to test whether \ac{VGS} models exhibit a \ac{ME} bias. 
The bias states that once and object has a name, it does not need another.
Therefore, novel words should belong to unknown objects.
We train a \ac{VGS} model to identify the spoken word and object representations of a set of familiar classes.
To evaluate the strength of the model’s ME bias, it is prompted with a novel spoken word and asked whether it belongs to a novel or familiar object.  
We found that a model using initialisations simulating the visual and audio knowledge that a child might have by the time they start using the bias, exhibited the strongest bias.
In addition, we performed extensive analysis to determine why the model exhibits the \ac{ME} bias and how specific our findings were to the model we chose.
We found that all models exhibited the \ac{ME} bias and the InfoNCE loss resulted in the strongest ME bias.

Lastly, we investigated the effect of multilingualism on the \ac{ME} bias exhibited by the InfoNCE model. 
We consider whether various language variants of the model exhibit the same \ac{ME} trends observed in children from different language backgrounds. 
We found that the monolingual model had a weaker \ac{ME} bias than the multilingual models, whereas monolingual children had a stronger \ac{ME} bias than multilingual children.
Additionally, we found that the \ac{ME} trends are affected by various design choices and concluded that the cognitive plausibility of design choices has to be considered carefully.

We believe this dissertation has given enough proof of how valuable VGS models can be and will encourage research in this field to build inclusive speech technology and contribute to understanding human language learning.

\chapter*{References}\markboth{}{\scshape References}
\addcontentsline{toc}{chapter}{References}
\renewcommand{\bibsection}{}
\bibliographystyle{plainnat}
\bibliography{mybib}
\appendix
\mychapter{Publication and contribution declaration: Research paper \fewshotInterspeech}{Publication and Contribution Declaration: Research paper \fewshotInterspeech}{}
\makeatletter\@mkboth{}{Appendix}\makeatother
\label{appen:interspeech}

In \RQfewshotInterspeech\ presented in this section, we propose our \ac{mattnet} to do multimodal few-shot learning using natural images and spoken words but with fewer word-image examples per class.
Before we get to \RQfewshotInterspeech\ in Section~\ref{interspeech_few-shot_publication}, we give a layout of each author's contribution to the paper in Section~\ref{interspeech_few-shot_contribution}.

\mysection{Contribution declaration}{Contribution Declaration}
\label{interspeech_few-shot_contribution}

\begin{table}[b]
	\caption{A layout of the authors’ contributions to the publication: \bibentry{nortje_visually_2023}.}
	\label{tab:few-shot_interspeech_contributions}
	\centering
	\renewcommand{\arraystretch}{1.2}
	\begin{tabularx}{\linewidth}{@{}lL@{}}
			\toprule
			Author &  Contributions \\
			\midrule
			\leanneName & The layout of the research question, the proposal of the model architecture, the pair mining algorithm, the implementation and the generation of the numerical results.\\
			\addlinespace
			\benjiName & The proposal and implementation of \qbert. \\
			\addlinespace
			\hermanName &   The layout of the research question and an editorial role.\\ 
			\bottomrule
	\end{tabularx}
\end{table} 
	
The first author, \leanneNameNoSpace, proposed and implemented all work in the following publication, except as stated otherwise in Table~\ref{tab:few-shot_interspeech_contributions}.	

\mysection{Paper}{Paper}
\label{interspeech_few-shot_publication}

We now present \RQfewshotInterspeech, which proposes \ac{mattnet} to do few-shot learning of natural images and spoken words with fewer word-image examples per class.

\newpage
\includepdf[pages={1-},scale=1.0]{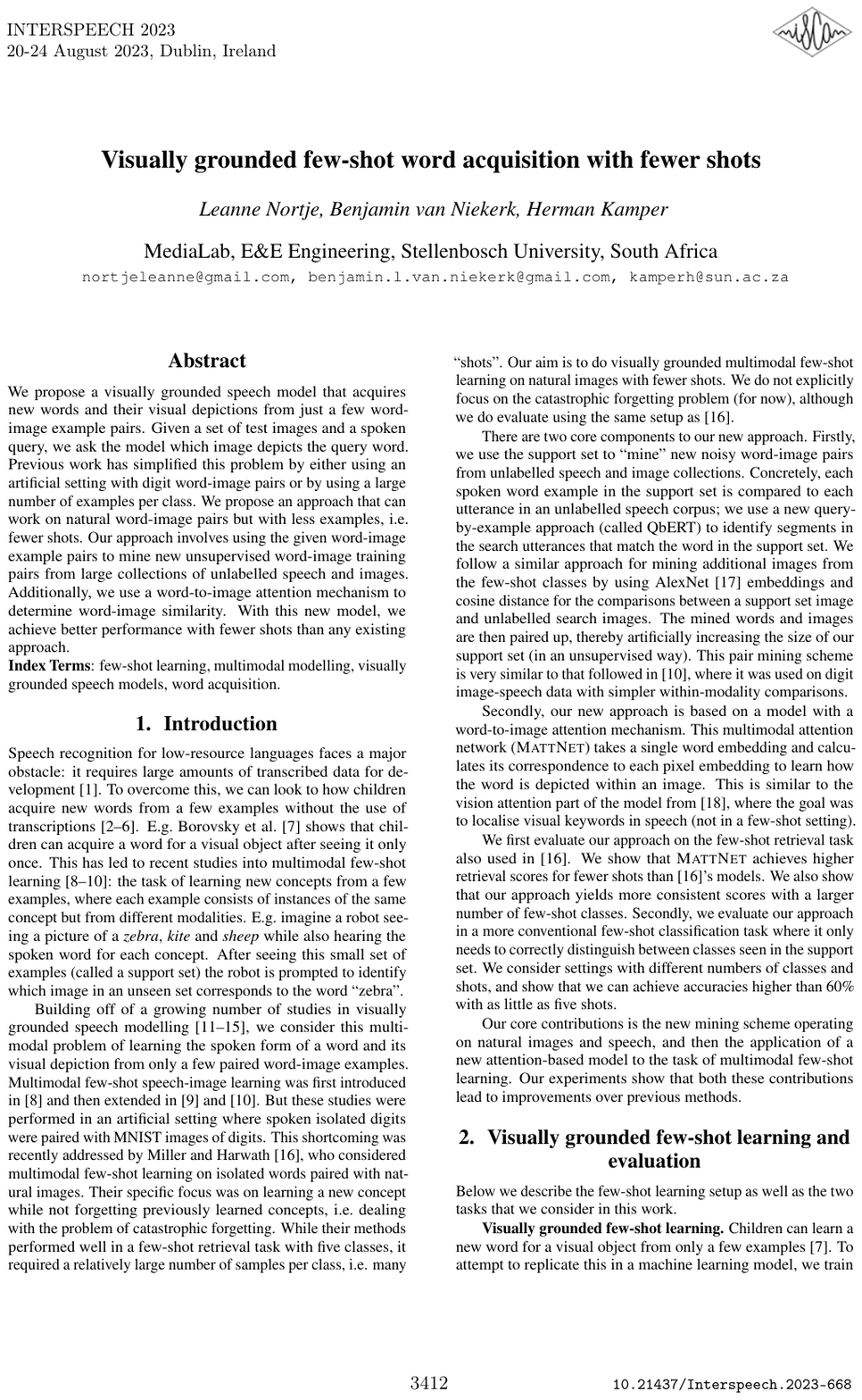}

\end{document}